\documentclass{report}
\ifdefined\directlua
  \usepackage{fontspec}
\else
  \usepackage[T1]{fontenc}
  \usepackage[nomath]{lmodern}
\fi
\usepackage[T1]{fontenc}
\usepackage[utf8]{inputenc}

\usepackage{longtable}

\usepackage[T1]{fontenc}
\usepackage{graphicx}
\usepackage[pagebackref=false,colorlinks,linkcolor=blue,citecolor=red]{hyperref}
\usepackage{caption}

\usepackage{amssymb}

\usepackage{amsthm}
\usepackage{amsmath}
\usepackage{mathtools}

\usepackage{csquotes}
\usepackage{bm}

\usepackage{xcolor}
\usepackage{algorithm,algorithmic}
\usepackage{caption}

\usepackage{eqparbox,array}

 \usepackage{parskip}
 \usepackage[a4paper, total={6in, 8in}]{geometry}

 \usepackage{enumitem}
\newlist{steps}{enumerate}{1}
\setlist[steps, 1]{label = Step \arabic*:}

\usepackage{float}

\usepackage{lipsum}
\usepackage{mwe}
\usepackage{graphicx}

\usepackage{titlesec}
\newcommand{\nocontentsline}[3]{}
\newcommand{\tocless}[2]{\bgroup\let\addcontentsline=\nocontentsline#1{#2}\egroup}

\usepackage{pdfpages}

\usepackage{fancyhdr}
\title{Control and implementation of fluid-driven soft gripper with dynamic uncertainty of object}

\newcommand{\thesisauthor}{Amirhosein Alian} 
\newcommand{\thesistype}{Master Thesis} 
\newcommand{\supervisor}{Dr. Mohammad Zareinejad, Dr. Heidar Ali Talebi}

\begin{document}

\newcommand\norm[1]{\left\lVert#1\right\rVert}
\newcommand\myeq{\stackrel{\mathclap{\normalfont\mbox{def}}}{=}}

\newcommand{\oldoptimal}[1]{{#1}^*}
\newcommand{\newoptimal}[1]{#1^*}


\makeatletter
\begin{titlepage}
    \begin{center}
         \begin{figure}[htbp]
             \centering
             \includegraphics[width=.4\linewidth, scale=1.5]{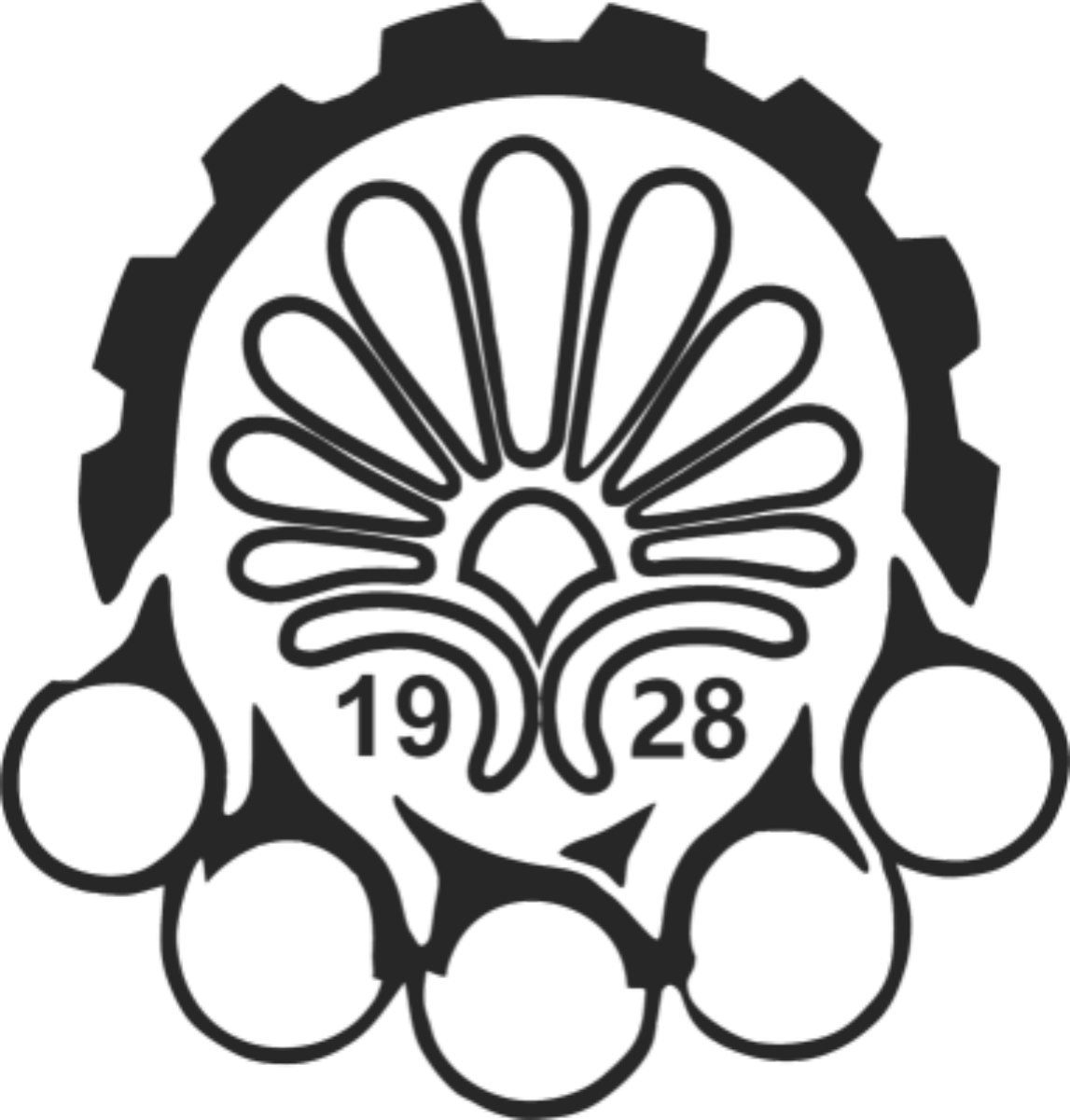}
        \end{figure}
        
        \huge
        Amirkabir University of Technology
        \\ 
        \smallskip
        
        \Large
        (Tehran Polytechnic) \\
        
        \smallskip
        \smallskip
        \Large
        Department of Mechanical Engineering\\
        
        \bigskip
        \bigskip
        
        \Large
        \thesistype{}
        
        \Large
        Field: Mechatronics Engineering \\
        
        \bigskip
        \bigskip
        \bigskip
        \bigskip
        
        \LARGE
        Title\\
        \@title

        \bigskip
        \bigskip
        
        \large
        Author \\
        \Large{}
        \thesisauthor{} \\ 
        
        \bigskip
        \bigskip
        
        \large
        Supervisor \\ 
        \Large
        \supervisor{}\\

        \bigskip
        \bigskip
        \bigskip
        \bigskip
        
        \Large
        September, 2020

    \end{center}
\end{titlepage}
\makeatother



\clearpage

\pagenumbering{gobble}
\clearpage
\thispagestyle{plain}
\clearpage
\clearpage
\thispagestyle{empty}
\section*{Disclaimer}
\label{sec:SOOA}


I hereby declare that Master Thesis of mine has been  written in Persian originally, and the following is a summary of my Thesis. 

\vspace{1cm}

\vspace{3cm}
\begin{center}

\end{center}
\vspace{0.5cm}

\clearpage
\thispagestyle{plain}
\pagenumbering{Roman}

\medskip
\medskip
\medskip
\begin{center}
\Large
\textbf{Abstract}
\end{center}
\vspace{1cm}
Soft grippers, for stable grasping of objects, with high compliance could be considered a suitable candidate for replacement of conventional rigid grippers, and in recent years, they have been emerging exponentially in industries. Not only are these highly adaptable grippers capable of static grasping of an object, but also they can be utilized for performing object manipulation tasks. \\
Plenty of contemporary studies have been emphasizing on static grasping ability of soft grippers. However, in this thesis, planar in-hand object manipulation in a soft gripper which comprises a pair of soft finger is investigated. Each soft finger is created by connection of two pneumatic bending actuators in series. Thus, a suitable mold is designed so as to cast soft fingers in the first place.  In order to derive the dynamic model of soft finger, the behavior of its rigid counterpart is modeled, which by determining an appropriate mapping is related to that of soft finger. After identification of dynamic parameters, in presence of gravity and external force exerted on the end effector, the dynamic model is validated.\\
 Considering the uncertainty in identification of parameters, bending curvature of each segment of the soft finger is controlled in experiments using sliding-adaptive controller. In addition, due to the effects of kinematic uncertainties caused by external forces, the trajectory of soft finger in Cartesian space is controlled by means of an adaptive controller which takes the kinematic uncertainties into account, and its results are compared to those of PID controller.\\
Finally, considering the dynamic model of soft fingers and their kinematic uncertainties, a control algorithm for in-hand manipulation of a rigid object, with predefined geometry, based on shaped and locked system is devised. Indeed, the dynamic of object would not be involved and its pose is controlled by the position of fingers' position in vertical plane.\\
\vfill

\textit{Keywords}: Soft Gripper, Piecewise Constant Curvature, Sliding-Adaptive Controller, Object In-Hand Manipulation, Dynamic and Kinematic Uncertainty

\clearpage
\tableofcontents
\clearpage
\listoffigures
\clearpage
\listoftables
\clearpage
\newpage

\pagenumbering{arabic}

\clearpage
\chapter{Introduction}
\clearpage

\section{Soft Robotics}

\begin{figure}[H]
	\centering
\captionsetup{justification=centering}		\includegraphics[width=95.71mm,height=56.82mm]{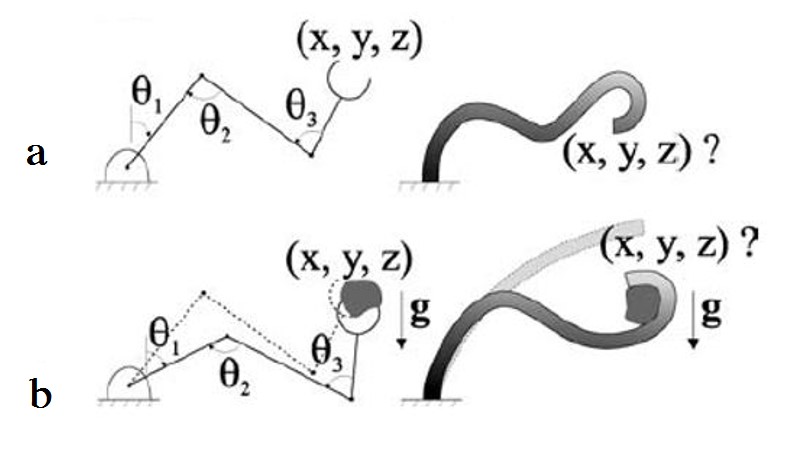}
\caption{ Soft robotics' challenges in comparison to conventional robotics \cite{13} a) infinite degree of freedom b) greater impacts of external forces }
\label{fig:1}
\end{figure}

Without the shadow of doubt, living organisms have been always a source of inspiration for engineers to come up with new ideas about their innovative robots and machineries \cite{23}. Having high compliance and secure interaction with environment, these creatures prompted scientists to create a new field of robotics referred to as soft robotics. Soft robots are defined as a special group of robots which are made of highly flexible materials like elastomers \cite{24} and their attributes are identical to those of living organisms. Compliance and resilience of soft robots are of significant importance due to establishment of safe interaction especially with human tissues and prevention of concentrated forces between contact surfaces.  \\
Beside its universal applications, soft robotics has always challenged scholars. Owing to their infinite degree of freedom, Soft robots' motion cannot be defined as accurately as conventional robots. In addition, their compliance turns soft robots to under-actuated systems, which makes the controlling of them more demanding. Moreover, as depicted in figure \ref{fig:1}, the impacts of gravitational force and external ones are more disturbing on soft robots because of their pliable structure \cite{13}.

\subsection{Soft Grippers} \label{subsec:timeconcept}

One of the most prominent applications of soft robotics is introduced in soft grippers. In fact, the capability of picking up an object and stably holding it against external disturbances is defined as grasping \cite{27}, and a system equipped with grasping is called a gripper, namely anthropomorphic hands. Rigid grippers, however, struggle with challenges like complexity in dynamical design and hardships in devising a functional control system in order to imitate the exact behavior of human's hand. Thus, in order to tackle these challenges, soft materials are implemented in grippers so as to elevate their compliance and dexterity to the point that soft grippers incorporate actuators which are devoid of any rigid components. Usage of soft grippers reduces the chance of damaging grasped objects, and due to their compliance, soft grippers increase the range of objects that can be handled. In other words, there would be no need to design computationally inefficient control algorithms to grasp objects with diverse geometry.\\

\begin{figure}[H]
	\centering
\captionsetup{justification=centering}		\includegraphics[width=75.32mm,height=79.08mm]{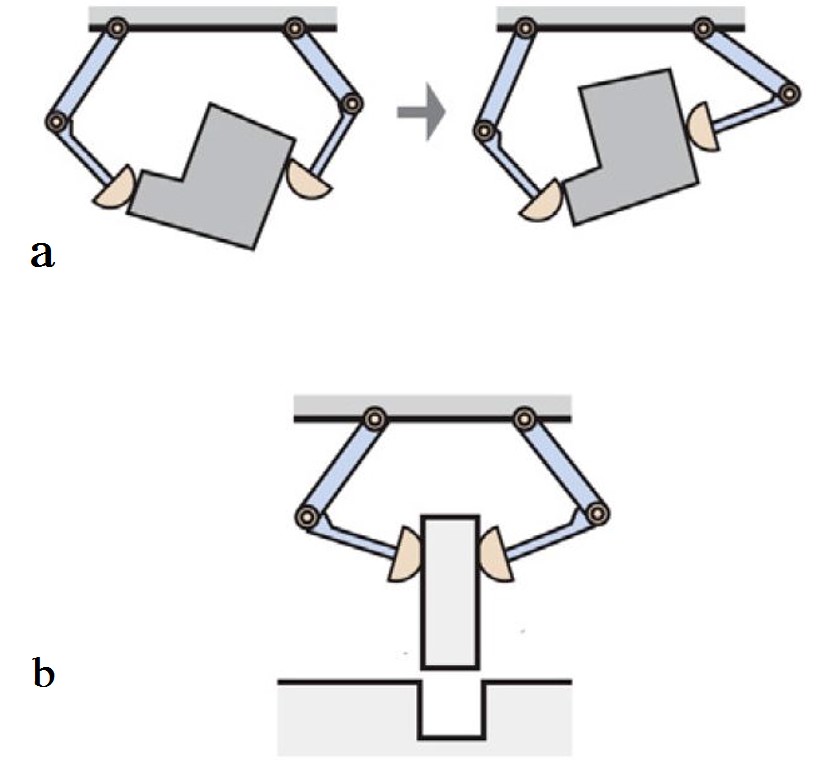}
\caption{ Performing in-hand object manipulation  \cite{59} . Changing the a) orientation b) position of the object }
\label{fig:2}
\end{figure} 
Soft grippers could be utilized for grasping an object statically or, inspired by human hand, changing its pose dynamically, which is referred to as in-hand object manipulation (figure \ref{fig:2}). For the purpose of illustration, considering our hand, it consists of arm and fingers.  Obviously, restraining the movement of arm does not necessarily prevent fingers from changing the position and orientation of an object within the palm \cite{59}. Therefore, a great deal of robotic manipulators are in demand for increasing their dexterity by performing in-hand object manipulation, which can be responded to by means of a gripper capable of in-hand manipulation. In addition to dexterity, manipulators could profit from in-hand manipulation skill in several aspects \cite{19}:
\begin{itemize}
\item{They can change the pose of an object even in confined spaces and in the presence of obstacles.}
\item{Energy efficiency: A vast amount of energy can be conserved since compared to movement of gripper solely, motion of the manipulator entirely dissipate more energy.}
\item{Manipulators can alter the pose of the grasped object even in their singularities.}
\end{itemize}

Beyond all the aforementioned benefits of in-hand manipulation, it can make a breakthrough in soft wearable prosthetic hands to the point that these artificial hands have abilities comparable to those of human hand. As a result, many people will have a more tangible and realistic experience interacting with their surroundings. \\
It is worth mentioning that there exist two approaches for object manipulation in terms of control system: 1) considering dynamic of the object, 2) also known as blind grasp, controlling the gripper regardless of object's dynamic. In the first approach, the dynamic of grasped object is related to dynamic equations of the whole system by grasp matrix which determines the forces and moments exerted on the object caused by the forces and moments applied by fingers of the gripper \cite{60}. Moreover, grasp matrix depends on the characteristics of contact surface, namely soft contact and frictionless contact. Obviously, this approach requires the real-time estimation of object's states \cite{21}. Conversely, blind grasp do not take the object's states into account, which obviates the need for real-time observation of the object \cite{1}. In this approach, it is postulated that no slippage occurs at the contact surfaces between object and fingers \cite{61}.\\
Considering the merits and different approaches of in-hand object manipulation, in this project we aim to lay the foundations for fabrication and implementation of a soft gripper with the capability of in-hand manipulation which takes advantage of blind grasping.\\
 
\section{Thesis outline}
\begin{figure}[H]
	\centering
\captionsetup{justification=centering}		\includegraphics[width=116.77mm,height=86.57mm]{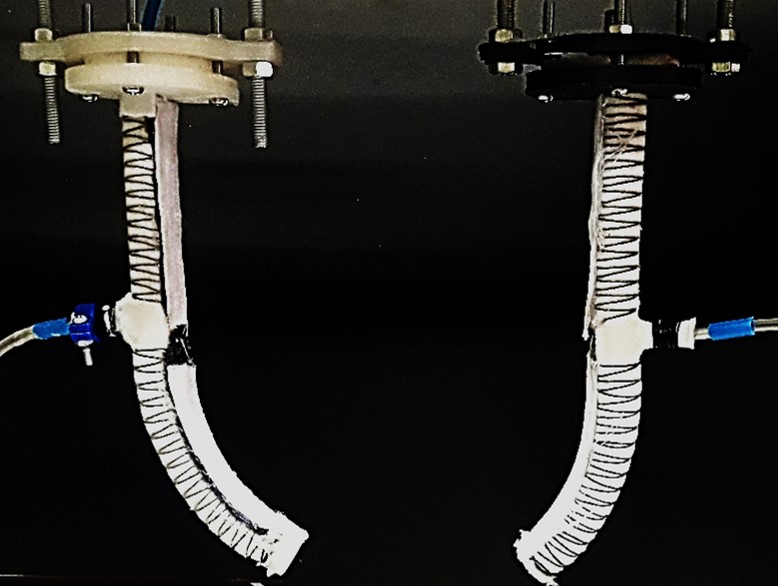}
\caption{ Proposed two-fingered gripper }
\label{fig:3}
\end{figure}

Regarding all the aforementioned explanations, the ultimate goal of this thesis is to propose a two-fingered soft gripper which has the benefit of in-hand manipulation and is shown in figure \ref{fig:3}. Thereby, in the following chapter, prior works are introduced first which chiefly underscore modeling and control of soft bending actuators. Then, in chapter 2, the fabrication of soft fingers are elaborated on, which incorporates the design of 3D-printed mold and sensors' embedding. In chapter 4, using piecewise constant curvature assumption, the dynamic modeling of fabricated soft finger will be investigated which is followed by identification of its parameters and validation of proposed model. Chapter 5 presents a sliding-adaptive controller and experiments for evaluation of designed control system that are conducted in both free space and interaction with the environment. The influence of external force on soft fingers will be explored in chapter 6, and an adaptive controller in Cartesian space is offered so as to neutralize the position error of end tip. Finally, we employ shaped and locked control system in order to enable the gripper to perform blind grasp in-hand manipulation in chapter 7. It should be noted that chapters 6 and 7 will not include experimental tests, and they solely provide simulation results.

\clearpage

\clearpage
\chapter{Related Works} \label{chap2}
\clearpage
A glimpse into the robotics researches published in the preceding years reveals that a tremendous amount of these articles are focusing on soft robotics. Apart from developing innovative procedures for fabrication of soft actuators, kinematic and dynamic modeling of their behavior along with designing a control algorithm have drawn researchers' attention. In the following paragraph, some of the most salient studies regarding the modeling and controlling of soft bending actuators are noted.\\

Basically, modeling of soft bending actuators falls into four distinct categories, the first of which is obtaining a relation between attributes of soft actuators and output force at their tip. In connection with this case, introducing a novel usage of mckibben muscles, Al-fahaam \cite{30} elevated the force exertion capacity of these pneumatic artificial muscles by redirecting the elongation of them to bending motion. Furthermore, he derived an analytical model for output force as a function of length and inlet pressure of actuators exploiting the theory of conservation of energy. In his study, dissipation of energy due to the radial expansion of actuator was compensated as well. Following his previous research on soft fiber reinforced bending actuators, Wang \cite{34} investigated the contribution of fiber pitch to the behavior of the actuator and proposed a novel quasi-static model between tip force and pneumatic pressure of the bending actuator. Thereafter, his model was validated by experiment and finite element method (FEM). Zhou \cite{39} inspected the correlation of tip force with the bending curvature of soft fingers involved in a gripper. Considering three different planar mechanism for gripping, Zhou utilized Euler's theory for the elastica to illustrate the relation between tip force and curvature.\\
The second category accommodates data-driven models for soft actuators. Elgeneidy \cite{32} implemented regression analysis and neural network to derive the empirical model between the voltage of flex sensor and curvature of actuator using image processing. Subsequently, he controlled the bending curvature with a simple PID controller. Likewise, Zheng \cite{53} extracted the relationship between inlet pressure and curvature of soft bending actuator, based on which he designed robust controller afterwards. This approach, however, required a vast set of data, which makes it computationally inefficient. Moreover, Tan \cite{58} developed a model-free controller for motion of a continuum robotic manipulator by data analysis of sensory output and input of the robot. In spite of controlling the kinematics, the algorithm was capable of compensation of diverse external disturbances.\\
Due to the significant importance of model-based controller, noticeable amount of studies have been dedicated to analytical dynamic modelling of soft actuators. In accordance with his prior research concerning a three-fingered gripper, Wang \cite{36} liken each soft finger to a serial manipulator composed of rigid links with point masses and viscoelastic revolute joints. Using Lagrangian equation, he attained dynamic model of soft fingers and subsequently, identified their parameters with an optimization-based method. Studying on Festo's soft manipulator, Falkenhahn \cite{42} exploited piecewise constant curvature hypothesis and derived the robot's dynamic model. In fact, he considered parallel dampers and springs for each section of the manipulator to illustrate their viscoelastic characteristics. Taking into account the work done by the input pressure, Falknhahn described the dynamic behavior of soft manipulator thoroughly by Lagrangian equation. Wang and Zhang in \cite{37} achieved a computationally efficient dynamic model of a single soft bending actuator with the help of constant curvature assumption and Euler-Lagrange equation. Instead of point mass, they surmised that the overall mass of actuator is distributed uniformly. Furthermore, they reduced the order of dynamic equation through Taylor series expansion. Later on, in their succeeding research \cite{38}, Wang and Zhang divided the dynamic of the soft actuator into motion dynamic and air dynamic with independent control systems. Thus, they assigned a second-order transfer and nonlinear function to the motion dynamic and air dynamic respectively and identified their parameters. Unexpectedly, they observed that identified parameters for motion dynamic varied with operating frequency. Furthermore, using backstepping method, they designed a nonlinear robust controller for trajectory tracking of the soft actuator.\\

In line with the researches on the subject of dynamic modeling of soft robots, Della Santina \cite{40} developed a dynamic model for a six-segmented soft arm with the help of constant curvature assumption. As a matter of fact the model was based on the association of the dynamic of virtually identical rigid serial manipulator to that of soft manipulator by a specific mapping. Assuming linear coefficients for damping and stiffness of the soft arm, he designed curvature and impedance controllers for the soft manipulator. Following a unique approach, Renda \cite{49} deployed constant curvature assumption and continuous Cosserat model to introduce discrete Cosserat approach and describe a soft manipulator's state by a set of constant strains. To enhance the accuracy of his model, Renda considered shear and torsional deformation of the cross section of its manipulator as well.\\
Eventually, the fourth and the last class of studies covers kinematic modeling of soft actuators. Regarding this case, Wang and Chen \cite{45} applied a Jacobian-based method for visual servo control of a cable-driven continuum robot. Finding the kinematic behavior of soft manipulator using constant curvature hypothesis, they linked translational and rotational speed of the actuator tip to the rate at which the curvature of soft arm was changing by Jacobian matrix. As a result, Wang and Chen were capable of implementing an error-based controller with gravity compensation. Shapiro \cite{50} developed a quasi-static model for kinematic of a soft bending actuator using Euler-Bernoulli beam theory for frequencies under 2 Hz. In addition, he enhanced the accuracy of his model by predicting material hysteresis.  Shapiro's model, however, did not take longitudinal and torsional deformation of cross section into account. The majority of studies based on the Euler-Bernoulli beam theory speculate the perpendicularity of cross section of the bean to its neutral axis, so neglect the impact of shear deformation and external load on behavior of the system. In contrast, using Timoshenko beam theory, Lindenroth \cite{52} investigated the influence of external force on actuator's structure with a stiffness-based modelling for forward kinematics. Besides, his model tackled the challenges associated with the usage of Timoshenko beam theory, namely overlooking elongation of the beam. Chen in \cite{57} scrutinized the kinematic behavior of a hyperelastic actuator through estimating its deflection after coming into contact with an obstacle in the environment as a function of pneumatic pressure and obstacle's whereabouts. In addition, employing moment balance and conservation of energy equations, he presented an approximate calculation of contact force and workspace of the actuator.

\clearpage

\clearpage
\chapter{Soft Finger Fabrication}
\clearpage
\label{subsec:fabr}

In order to perform in-hand object manipulation within vertical plane for this project, two criteria for designing the soft finger were set:

\begin{itemize}
\item{Its workspace has to include an area rather than a curve, which leads to the fact that each finger has to possess at least two degree of freedom.}
\item{Since not more than four pneumatic valves were available for experiments, we were bound to utilize unidirectional soft actuators in our study}
\end{itemize}
Thereby, we decided to take advantage of two fiber-reinforced bending actuators which were connected in series. The fabrication of the soft finger incorporates three main steps \cite{galloway2013}: 1) designing the mold in order to cast the finger. 2) Constraining the radial expansion of actuator by fiber windings along the actuator's length in a double helix pattern. 3) Impeding elongation of actuators by means of an inextensible layer which is usually a piece of cloth.\\

\section{Mold Design}

\begin{figure}[H]
	\centering
\captionsetup{justification=centering}		\includegraphics[width=137.63mm,height=73.45mm]{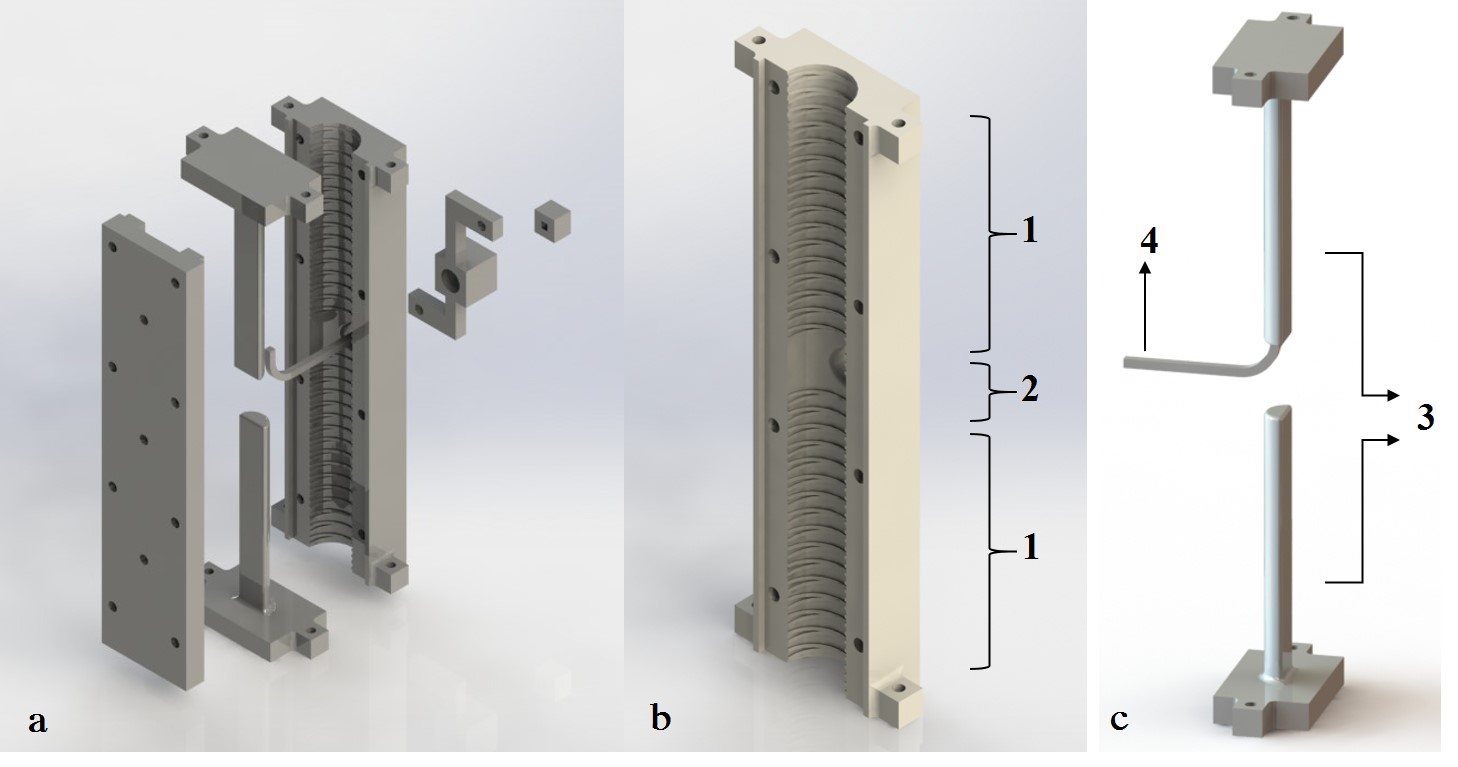}
\caption{ Proposed mold for fabrication of soft fingers. a) its components and how they are assembled. b) Staple component of the mold. c) half-round rods to define interior hallow portions of the actuators. }
\label{fig:4}
\end{figure}

Considering the serial connection of two bending actuators and the fact that the soft fingers are supposed to be exposed to external forces, it seems impractical to mold each one of the actuators separately and then connect them since discontinuity could occur. In other words, molding each actuator in an asynchronous manner and linking them afterwards might result in weak molecular bond at the junction, which has adverse impacts on the structure of the soft finger. As a result, the mold should be designed in a way that provides simultaneous fabrication of two bending actuators and their connection. Furthermore, since each of the actuators should be driven independently, separate pressure inlets should be devised in the design of the proposed mold. Considering all the aforementioned explanations, we concluded that the mold depicted in Figure \ref{fig:4}.a can meet our expectations. In addition, more detailed information is offered in Figure \ref{fig:4}.b and  \ref{fig:4}.c.\\

As illustrated in Figure \ref{fig:4}.b, the staple component of the mold comprises two main portions. Portion 1 indicates the location in which the bending actuators lie, and portion 2 is responsible for the formation of lower bending actuator's pressure inlet. Shown in Figure \ref{fig:4}.c, the rods of mold prevent the actuator from getting filled with silicon entirely. Portion 3 generates a vacant semi-cylinder volume in bending actuators, and portion 4 defines a path for pneumatic pressure. Moreover, geometric properties of the proposed mold are listed in table \ref{tab.1}.

\begin{table}[ht]
\caption{geometric properties of the mold} 
\centering 
\begin{tabular}{c c} 

\hline\hline 
Properties & Size (mm)  \\ [0.5ex] 
\hline 
Length of the mold & 175  \\ 
Inner diameter of cross section & 12.7  \\
outer diameter of cross section & 20.7  \\
Length of the junction & 15  \\
Fiber pitch & 5  \\
Width of fiber path & 1  \\ [1ex] 
\hline 
\end{tabular}
\label{tab.1} 
\end{table}

\begin{figure}[H]
	\centering
\captionsetup{justification=centering}		\includegraphics[width=102.63mm,height=93.45mm]{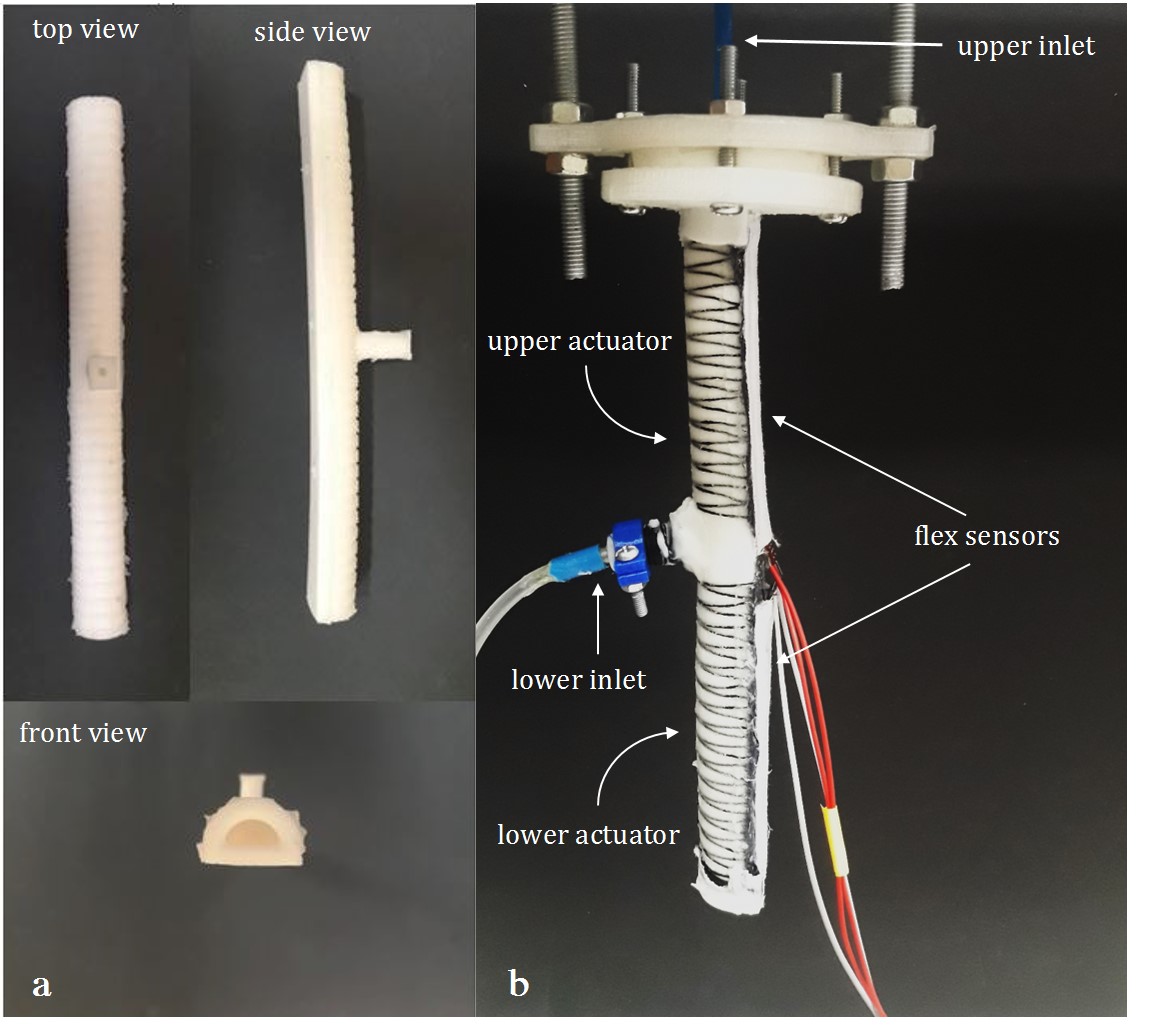}
\caption{ Fabricated soft finger. a) Three main views of molded rubber body. b) Different parts of the soft finger  }
\label{fig:5}
\end{figure}
After 3-D printing of the mold, we made use of silicon rubber $RTV2-325$ as the main material for casting the soft finger. Below in  Figure \ref{fig:5}.a are three main views of molded rubber body of the soft finger.\\

At the next step, the reinforced fibers were woven along the actuators, and the inextensible layers were fixed and positioned on the flat side of the two actuators. Finally, the process of fabrication was brought to an end with two ends of the finger being capped. For the purpose of measuring the bending curvature of actuators, 2.2 inch flex sensor of Spectra Symbol Corporation was employed. Depicted in Figure \ref{fig:5}.b, flex sensors were embedded on the inextensible layers so as to observe bending curvatures as accurately as possible. Furthermore, $LM324N$ was utilized to amplify the voltage output of the flex sensors.

\vfill

\clearpage
\chapter{Modelling of Soft Finger} \label{sec:nnmf}
\clearpage
%
Since developing a model-based controller requires determination of the dynamic behavior of the system, in this chapter, modelling of the soft finger is explored. Among the studies concerning the modelling of soft robots, the frame work proposed in \cite{40} was utilized for our case.\\

\section{ Kinematic Model }
\begin{figure}[H]
\centering
\captionsetup{justification=centering}		\includegraphics[width=139mm,height=65mm]{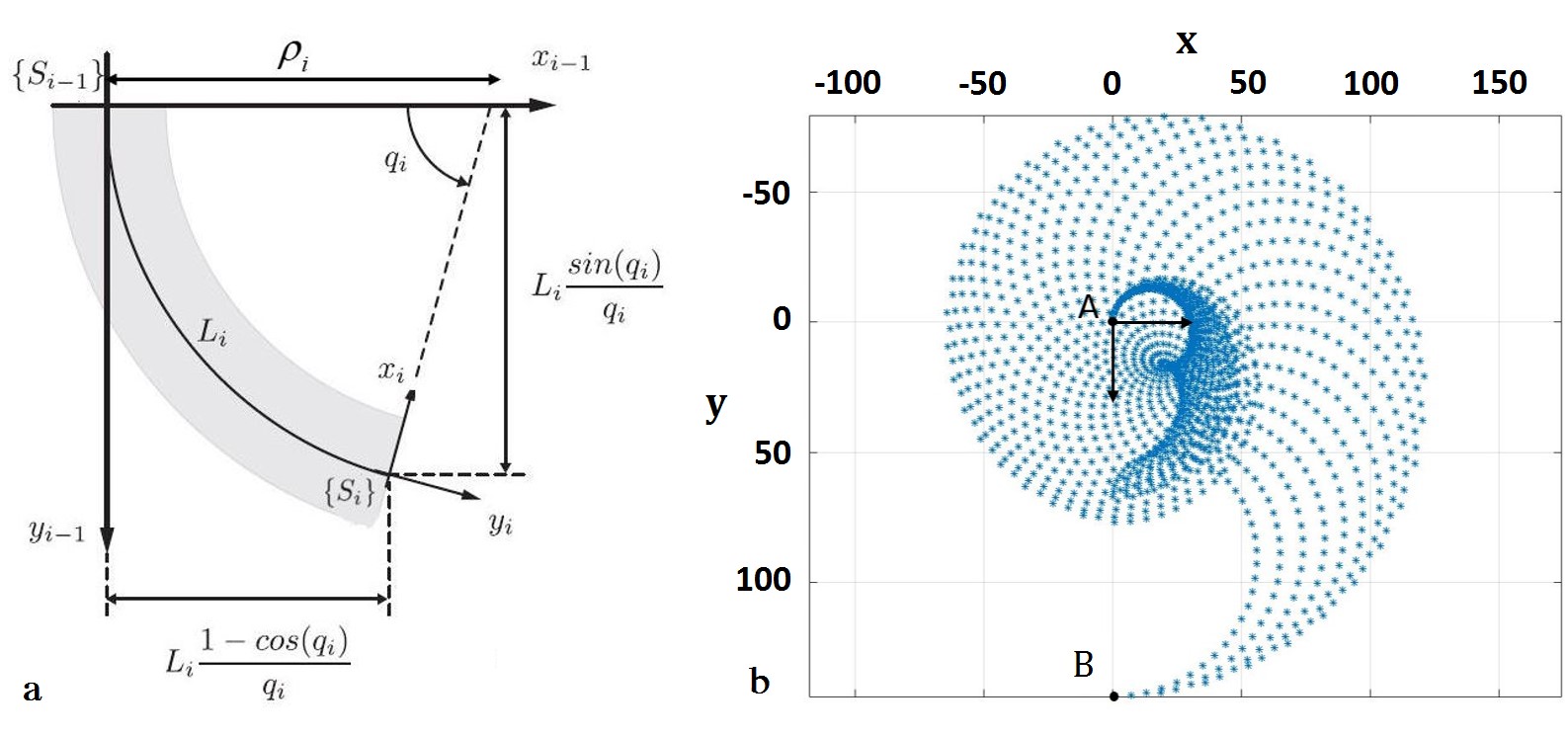}
\caption{ a) Single bending actuator planar bending \cite{20}. b) Work space of the two-segmented soft finger for 360 degree bending of each actuator. }
\label{fig:6}
\end{figure}

As illustrated in Figure \ref{fig:6}.a, the bending curvature of each actuators is defined by $q_i \in \mathbb{R}^{2}$ , so the transformation matrix relating the tip of the actuator to its base frame could be obtained by \eqref{eq:3-3} through applying the constant curvature hypothesis. Therefore, the work space of the proposed soft finger containing two bending actuators could be shown as in Figure \ref{fig:6}.b. It should be noted that points A and B in Figure \ref{fig:6}.b indicate the base and the tip of the soft finger respectively.

\begin{equation} \label{eq:3-3}
T_{i-1}^i (q_i) = 
\begin{bmatrix}
cos(q_i) & -sin(q_i) & L_i \frac{1 - cos(q_i)}{q_i} \\
sin(q_i) & cos(q_i) & L_i \frac{sin(q_i)}{q_i} \\
0 & 0 & 1 
\end{bmatrix}
\end{equation}

\section{ Dynamic Model }
\begin{figure}[H]
\centering
\captionsetup{justification=centering}		\includegraphics[width=65mm,height=57mm]{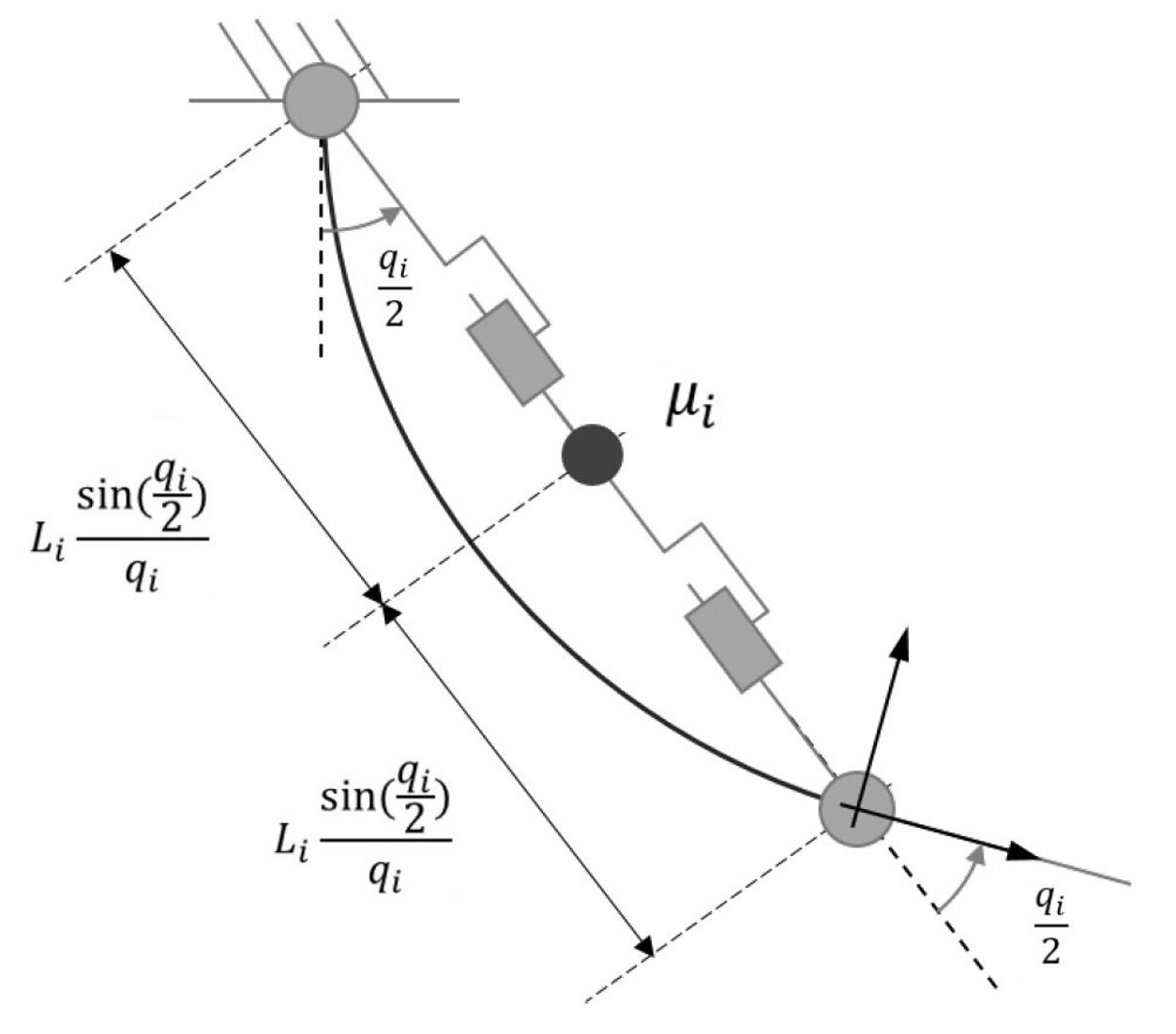}
\caption{ Single bending actuator and proposed equivalent rigid manipulator \cite{40}}
\label{fig:7}
\end{figure}

Having described the kinematic behavior of the soft finger by multiplying transformation matrices, we could head for modelling of its dynamic behavior. In order to do so, the dynamic behavior of equivalent rigid manipulator is modeled, followed by determining an appropriate mapping which relates joint variables of rigid manipulator to bending angles of soft finger. It should be taken into account that both soft finger and its rigid counterpart have to exhibit identical kinematic and dynamic behavior. Accordingly, the tip of proposed rigid manipulator and its center of mass should coincide with those of soft finger. Considering the uniform mass distribution along soft bending actuator, Figure \ref{fig:7} presents a prospective candidate for rigid manipulator with its center of mass which is assigned as a point mass to its second link. The proposed rigid robot has two revolute joints at its both ends and two prismatic joints at the middle (RPPR).

Since the soft finger in this project consists two bending actuator, the equivalent rigid manipulator, which is obtained by connecting two RPPR manipulators in series, would end up having eight degree of freedom. As previously mentioned, the mapping $m(q)$ is devised so as to link dynamic behavior of two robots such that $m: \mathbb{R}^2 \to \mathbb{R}^{2 \times 4}$. Assigning $\zeta \in \mathbb{R}^{2 \times 4}$ to joint variables of rigid manipulator, we can compute the mapping through $\zeta = m(q)$. Hence, calculation of DH parameters results in deriving the elements of $m(q)$ \cite{40}. for the configuration shown in Figure \ref{fig:7} $m(q)$ is computed as equation \eqref{eq:3-8} which is based on the DH parameters of Table \ref{tab.3.1}.

\begin{table}[ht]
\caption{ DH parameters for equivalent rigid manipulator of a single bending actuator} 
\centering 
\begin{tabular}{c c c c c} 

\hline\hline 

		Link number &  $\theta$  & $d$   &  $a$	& $\alpha$	
		\\[0.5ex]
		\hline 
		$1$		       &  $\frac{q_1}{2}$			  		&$0$		 		 &  $0$			     & 		$-\pi$    \\[1ex]
		$2$		       &  $0$			  		&$L_1 \frac{sin(\tfrac{q_1}{2})}{q_1}$	  	 		 &  $0$			& 	  $\frac{\pi}{2}$       \\[1ex]
		$3$		       &  $0$			  		&$L_1 \frac{sin(\tfrac{q_1}{2})}{q_1}$	     		 &  $0$		& 	   $0$ 	 \\[1ex]
		$4$		       &  $-\frac{q_1}{2}$		 	   &$0$		  		  &  $0$   &		$\frac{\pi}{2}$		 \\[1ex]
		\hline 
	\end{tabular}
	\label{tab.3.1}
\end{table}

\begin{equation} \label{eq:3-8}
m(q)={{\left[ \begin{matrix}
   {{m}_{1}}{{({{q}_{1}})}^{T}} & {{m}_{2}}{{({{q}_{2}})}^{T}}  \\
\end{matrix} \right]}^{T}}
\end{equation}

After obtaining the dynamic equation of equivalent rigid manipulator as equation \eqref{eq:3-10}, substitution of joint variables and their derivatives with equations \eqref{eq:3-11} to \eqref{eq:3-13} generates equation \eqref{eq:3-17}, the matrices of which are defined according to equation \eqref{eq:3-18}.

\begin{equation} \label{eq:3-10}
{{M}_{\zeta }}(\zeta )\ddot{\zeta }+{{C}_{\zeta }}(\zeta ,\dot{\zeta })\dot{\zeta }+{{G}_{\zeta }}(\zeta )=J_{\zeta }^{T}(\zeta ){{f}_{ext}}
\end{equation}

\begin{equation} \label{eq:3-11}
\zeta =m(q)
\end{equation}
\begin{equation} \label{eq:3-12}
\dot{\zeta }={{J}_{m}}(q)\dot{q}
\end{equation}
\begin{equation} \label{eq:3-13}
\ddot{\zeta }={{\dot{J}}_{m}}(q,\dot{q})\dot{q}+{{J}_{m}}(q)\ddot{q}
\end{equation}	

\begin{equation} \label{eq:3-17}
M(q)\ddot{q}+C(q,\dot{q})\dot{q}+G(q)={{J}^{T}}(q){{f}_{ext}}
\end{equation}

\begin{equation} \label{eq:3-18}
\begin{cases}
M(q) = J_m^T (q) M_{\zeta} (m(q)) J_m (q) \\

C(q,\dot{q})  = J_m^T (q)  M_{\zeta} (m(q)) \dot{J}_m (q,\dot{q}) + J_m^T (q) C_{\zeta} \big(m(q), J_m (q) \dot{q}\big) J_m (q) \\

G(q) = J_m^T (q) G_{\zeta} (m(q))\\

J^T (q) = J^T (q) = J_m^T (q) J_{\zeta}^T
\end{cases}
\end{equation}

Moreover, since the viscoelastic property of soft finger's material plays a prominent role in its dynamic behavior, constant diagonal matrices of $K$ and $D$, which have yet to be identified, are assigned to elastic and dissipative traits of soft finger respectively. In addition, considering external wrenches and control torques caused by pneumatic pressure, equation \eqref{eq:3-19} describes the dynamic behavior of our soft fingers. It should be noted that the control torques are associated to pneumatic pressure through linear functions, slopes of which are identified and indicated by $\alpha_i$.

\begin{equation} \label{eq:3-19}
M(q)\ddot{q}+(C(q,\dot{q})+D)\,\dot{q}+G(q)+K\,q=\tau +{{J}^{T}}(q){{f}_{ext}}
\end{equation}

\subsection{Identification}

Needless to say, some of the parameters of equation \eqref{eq:3-19}, namely mass, actuators' length, stiffness and damping coefficients should be identified. Mass and length of each bending actuator are determined with direct measurement while identification procedure for stiffness and damping coefficients as well as slopes of torque-pressure function follows an indirect method \cite{santina2020}. After exerting a ramp pressure signal from 0.3 bar to 1.8 bar with slope of 0.1 bar per second on each one of the actuators, based on the running time of experiment and sampling time, a host of numerical dynamic equations are acquired. Rearrangement of these equations as an equilibrium between known terms and the product of unknown parameters and their regressor, as in equation \eqref{eq:3-21}, could lead to identification of unknown dynamic parameters.

\begin{equation} \label{eq:3-21}
\begin{cases}
M(q_{{sens}_1}) \ddot{q}_{{sens}_1} + \big(C (q_{{sens}_1}, \dot{q}_{{sens}_1}) + D \big) \dot{q}_{{sens}_1} + G(q_{{sens}_1}) + Kq_{{sens}_1} = \alpha P_1 \\

\vdots

\hspace{12.2cm}\implies Ax = Y
\\

M(q_{{sens}_n}) \ddot{q}_{{sens}_n} + \big(C (q_{{sens}_n}, \dot{q}_{{sens}_n}) + D \big) \dot{q}_{{sens}_n} + G(q_{{sens}_n}) + Kq_{{sens}_n} = \alpha P_n \\

\end{cases}
\end{equation}

In equation \eqref{eq:3-21}, matrices of $A$, $x$ and $Y$ are defined as in equations \eqref{eq:3-22} to \eqref{eq:3-24}. Therefore, $x$ is obtained by pre multiplying of both sides of equation \eqref{eq:3-21} by inverse of $A$.

\begin{equation} \label{eq:3-22}
A = 
\begin{bmatrix}
diag(q_{{sens}_1}) & diag(\dot{q}_{{sens}_1}) & -diag(P_1)\\
\vdots & \vdots & \vdots \\
diag(q_{{sens}_n}) & diag(\dot{q}_{{sens}_n}) & -diag(P_n)
\end{bmatrix}
\end{equation}
\begin{equation} \label{eq:3-23}
x=\left[ \begin{matrix}
   {{K}_{1}} & {{K}_{2}} & {{D}_{1}} & {{D}_{2}} & {{\alpha }_{1}} & {{\alpha }_{2}}  \\
\end{matrix} \right]
\end{equation}				
\begin{equation} \label{eq:3-24}
Y=-M({{q}_{sens}}){{\ddot{q}}_{sens}}-C({{q}_{sens}},{{\dot{q}}_{sens}}){{\dot{q}}_{sens}}-G({{q}_{sens}})
\end{equation}

Eventually, following the aforementioned procedures, we can identify the parameters as in Table \ref{tab.3.2}. The most noticeable parameters among the rest are damping coefficients which are considerably low; however, based on the actuators' material, it was anticipated due to the instantaneous response of the actuators to change in the pressure.

\begin{table}[h]
\caption{Identified parameters for each actuator} 
\centering 
\begin{tabular}{c c c c} 
\hline\hline 
Parameter &Unit & bending actuator & quantity
\\ [0.5ex]
\hline 
& &Upper &20  \\[-1ex]
\raisebox{1.5ex}{$\mu$} & \raisebox{1.5ex}{$gr$}&Lower
& 25.1  \\[1ex]
& &Upper & 67 \\[-1ex]
\raisebox{1.5ex}{$L$} & \raisebox{1.5ex}{$mm$}& Lower
&77  \\[1ex]
& &Upper & 0.068  \\[-1ex]
\raisebox{1.5ex}{$K$} & \raisebox{1.5ex}{$\frac{N m}{rad}$}& Lower
&0.07 \\[1ex]
& &Upper & 0.0029  \\[-1ex]
\raisebox{1.5ex}{$D$} & \raisebox{1.5ex}{$\frac{N m s}{rad}$}& Lower
&0.0029 \\[1ex]
& &Upper & 0.076  \\[-1ex]
\raisebox{1.5ex}{$\alpha$} & \raisebox{1.5ex}{$\frac{N m}{bar}$}& Lower
&0.062  \\[1ex]

\hline 
\end{tabular}
\label{tab.3.2}
\end{table} 

\begin{figure}[H]
	\centering
	\captionsetup{justification=centering}		\includegraphics[width=156mm,height=73mm]{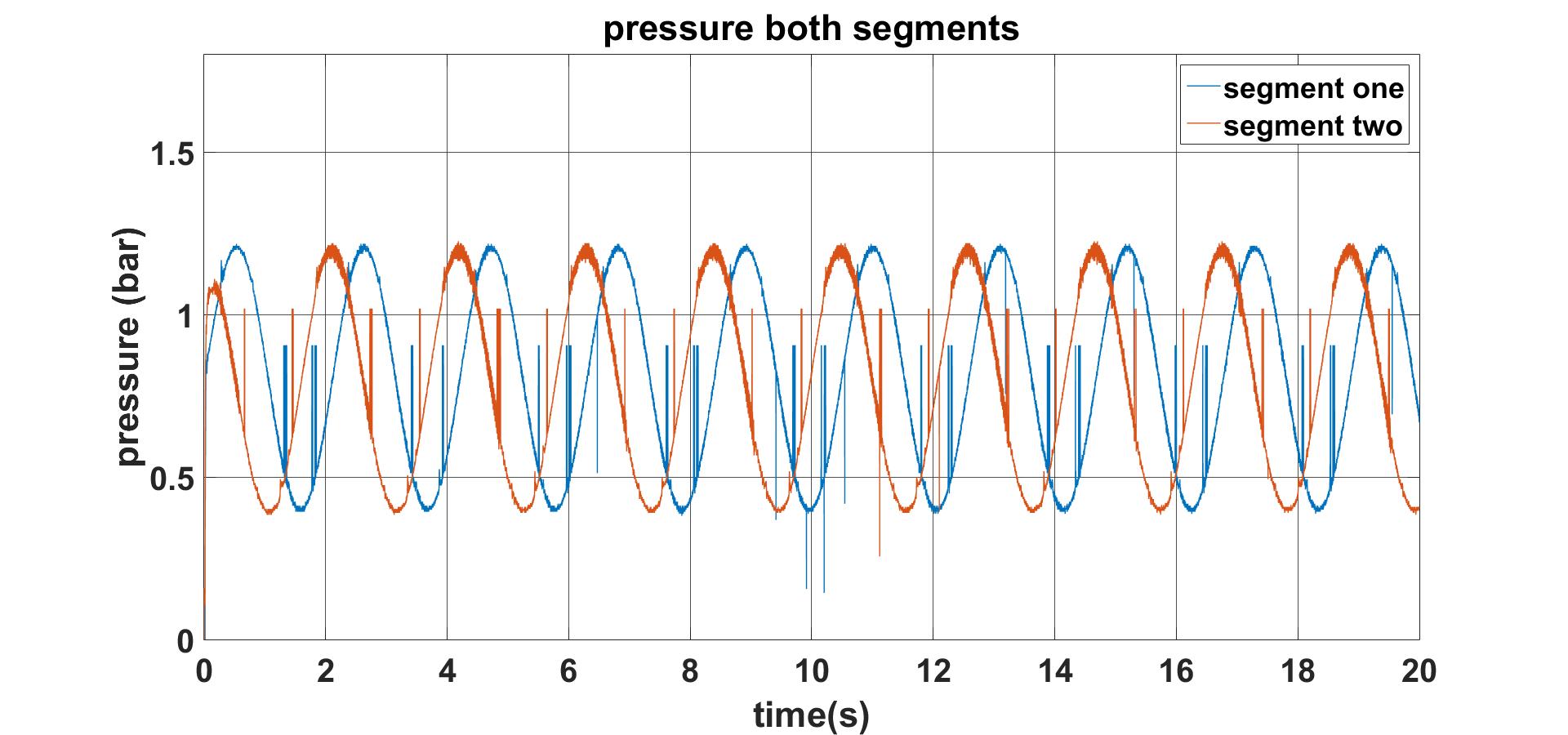}
\caption{ Input pressure for dynamic validation}
\label{fig:3-10}
\end{figure}

\begin{figure}[H]
	\centering
	\captionsetup{justification=centering}		\includegraphics[width=156mm,height=73mm]{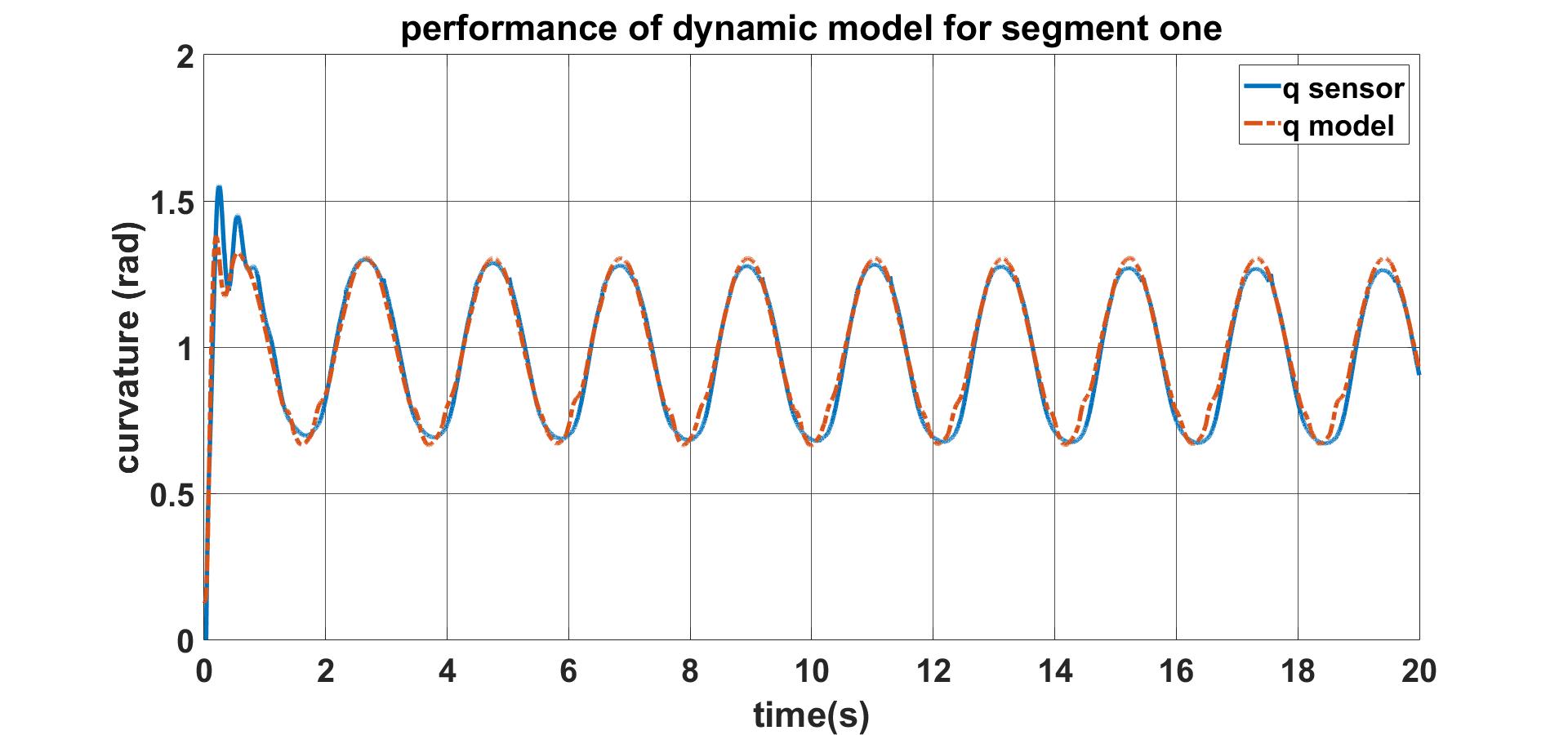}
\caption{ Accuracy of dynamic model for upper bending actuator as a response to pressure $0.8 + 0.4 sin(3t)$}
\label{fig:3-11}
\end{figure}

\begin{figure}[H]
	\centering
	\captionsetup{justification=centering}		\includegraphics[width=156mm,height=73mm]{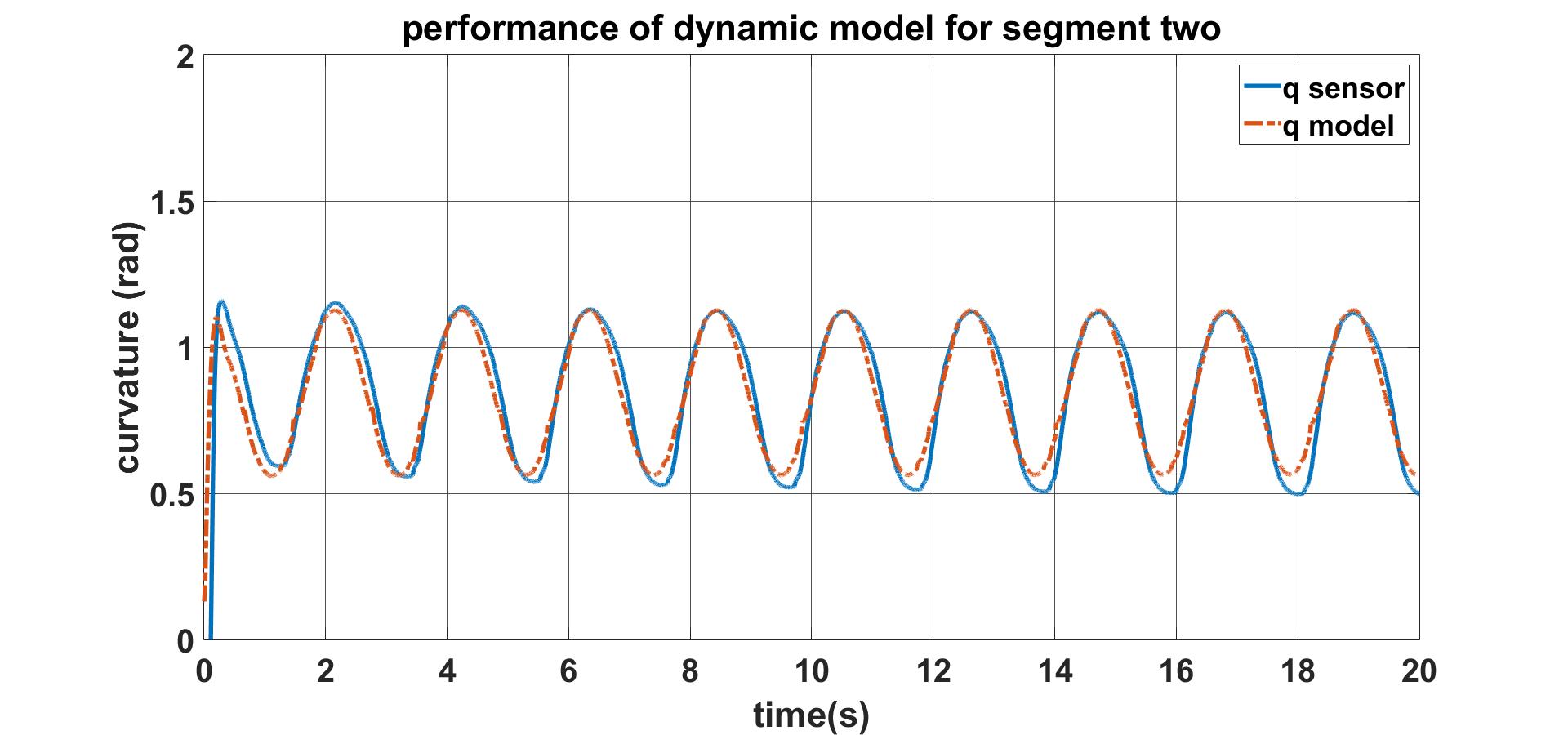}
\caption{ Accuracy of dynamic model for lower bending actuator as a response to pressure $0.8 + 0.4 sin(3t+1.57)$}
\label{fig:3-12}
\end{figure}

\begin{figure}[H]
	\centering
	\captionsetup{justification=centering}		\includegraphics[width=156mm,height=73mm]{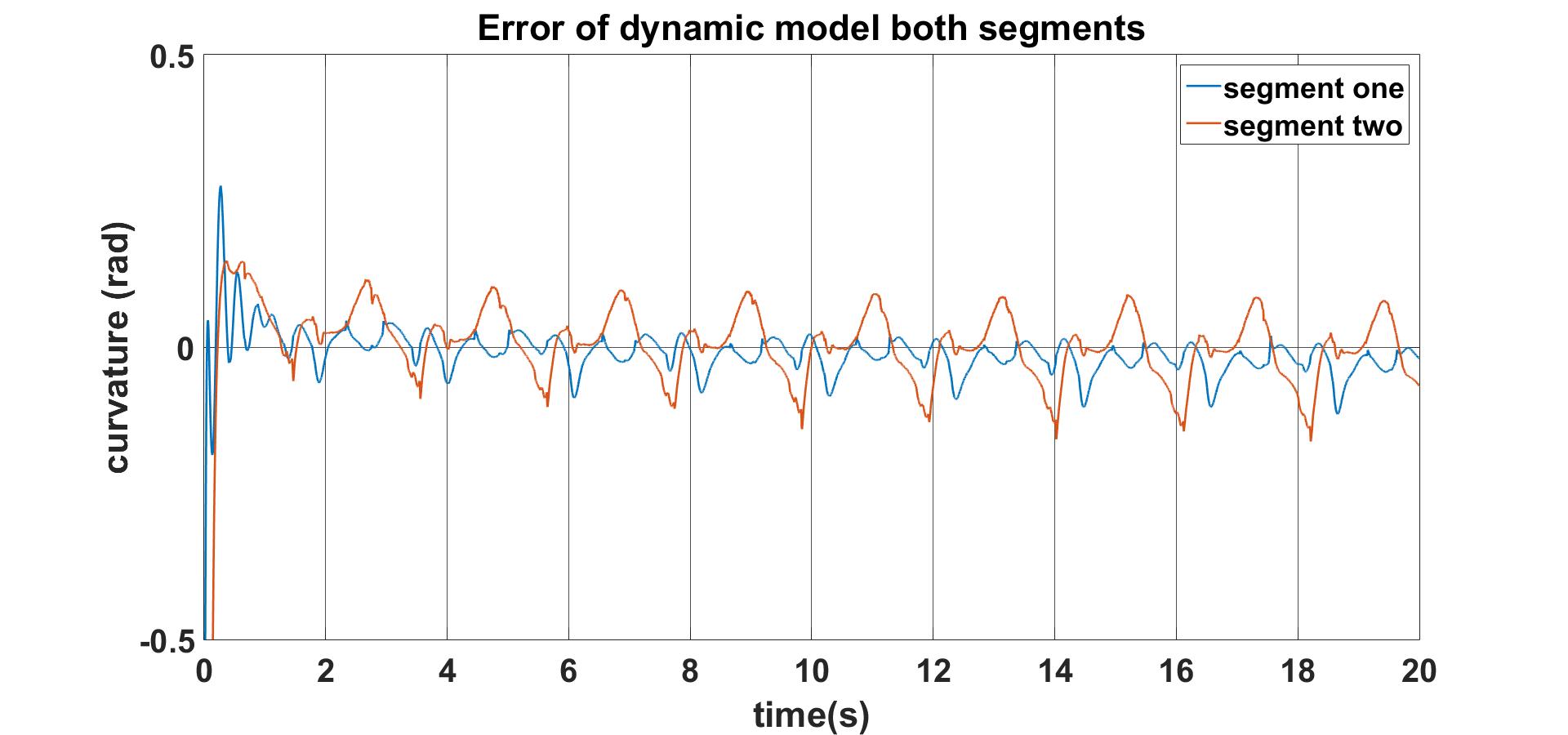}
\caption{ Error of dynamic modelling for both actuators}
\label{fig:3-13}
\end{figure}

In order to validate identified parameters and dynamic equations, two harmonic pressure input are injected to bending actuators both in experiment and simulation. Data acquired from flex sensors in experimental testing, then, are compared to bending curvature prediction of derived dynamic equation through simulations conducted in Simulink MATLAB. Therefore, as depicted in Figure \ref{fig:3-10} pressure inputs of $0.8+0.4 sin (3t)$ and $0.8+0.4 sin (3t+1.57)$ bar were exerted on segment one (upper bending actuator) and segment two (lower bending actuator) respectively. The results of experimental testing and simulation for each segment are shown in Figures \ref{fig:3-11} to \ref{fig:3-13}.

It should be noted that:
\begin{itemize}
\item{Since the output voltage of flex sensors are accompanied with high-frequency noise, we utilized a low-pass filter with frequency of 30 Hz so as to collect more smooth data.}
\item{Pneumatic pressure is applied by Rexroth EP-1262A-01 valves and measured by Festo SDE1 sensors. In fact, the output of pressure sensors are used as pressure inputs for simulations.}
\end{itemize}

Apparently, the proposed dynamic model could predict the bending angle of the upper bending actuator more precisely according to Figure \ref{fig:3-13} since the error between simulation results and flex sensors' output seems noticeably less for segment one. This could be accounted for by the fact that the segment one is shorter in length, so its hollow portion get filled more swiftly, which results in more immediate response to pressure variation. In order to have a brief overview on the accuracy of proposed dynamic model and parameters identification, Table \ref{tab.3.4} indicates root mean square error $RMSE$ between dynamic model and flex sensor measurements for each segment individually.

\begin{table}[ht]
\caption{$RMSE$ for dynamic modeling of the bending actuators} 
\centering 
\begin{tabular}{c c} 

\hline\hline 
 $RMSE$ segment one $rad$ ($i=1$) &  $RMSE$ segment one $rad$ ($i=2$) \\ [0.5ex] 
\hline 
 0.0474 & 0.1278  \\ 

\hline 
\end{tabular}
\label{tab.3.4} 
\end{table}
   

\clearpage
\chapter{Bending Angle Control of Soft Finger} \label{sec:nnmf}
\clearpage
%
Having derived dynamic model of soft finger, we planned to design a controller which guaranteed the convergence of the bending angles to reference inputs. Thus, in this chapter we elaborate on a proposed control system and its validation through experimental testing. In addition, experiments are conducted for both free motion and interaction with environment. Furthermore, the results of proposed controller are compared to those of conventional PID controller.\\
Exploring the studies concerning modeling of soft robots, one soon realizes that properties of a soft robot can vary over the course of time, or they can have diverse values depending on the operating frequency \cite{38}. Besides, in this study we postulated that stiffness and damping of our soft finger exhibit linear behaviors, which is at odds with the fact that dynamic parameters might be time variant. Consequently, we introduce an adaptive-sliding controller based on study \cite{66} with compensating term for external disturbances. \\

\section{ Adaptive-Sliding Controller }

Rewriting the dynamic equation of soft finger presented in equation \eqref{eq:3-19} in regressor form, we could obtain equation \eqref{eq:4-1} in which the gravity and Coriolis matrices incorporate elastic and dissipative terms as well. $\theta_d$ defines dynamic parameters including mass, stiffness and damping coefficients which would be estimated by adaptive laws. 

\begin{equation} \label{eq:4-1}
M(q)\ddot{q}+C(q,\dot{q})\dot{q}+G(q)={{Y}_{d}}(q,\dot{q},\ddot{q}){{\theta }_{d}}
\end{equation}

Sliding vector and its derivative are specified by equations \eqref{eq:4-2} and \eqref{eq:4-4} in which $\dot{q}_r$ is computed as in equation \eqref{eq:4-3}. In this equation $\Lambda$  is a positive diagonal matrix, and $q_d$  is the vector of desired bending angles.

\begin{equation} \label{eq:4-2}
s = \dot{q} - \dot{q}_r
\end{equation}

\begin{equation} \label{eq:4-4}
\dot{s} = \ddot{q} - \ddot{q}_r
\end{equation}

\begin{equation} \label{eq:4-3}
\dot{q}_r = \dot{q}_d - \Lambda (q-q_d)
\end{equation}

Equation \eqref{eq:3-19} can be rearranged by replacing the derivatives of $q$ from equations \eqref{eq:4-2} and \eqref{eq:4-4}.

\begin{equation} \label{eq:4-5}
M(q)\dot{s}+C(q,\dot{q})s+M(q){{\ddot{q}}_{r}}+C(q,\dot{q}){{\dot{q}}_{r}}+G(q)=\tau +{{J}^{T}}(q){{F}_{ext}}
\end{equation}	

Using equation \eqref{eq:4-1}, equation \eqref{eq:4-5} can be replaced by equation \eqref{eq:4-6}.

\begin{equation} \label{eq:4-6}
M(q)\dot{s}+C(q,\dot{q})s+{{Y}_{d}}(q,\dot{q},{{\dot{q}}_{r}},{{\ddot{q}}_{r}}){{\theta }_{d}}=\tau +{{J}^{T}}(q){{F}_{ext}}
\end{equation}	

Thereby, the control signal of proposed algorithm can be determined by equation \eqref{eq:4-7} in which $\Delta q=q-{{q}_{d}}$, $k_p$ and $k_d$ are positive diagonal matrices, $\hat{F}_{ext}$  and $\hat{\theta}_{d}$  are estimated values for external disturbances and dynamic parameters respectively. These parameters are adapted with control laws of equations \eqref{eq:4-8} and \eqref{eq:4-9} in which $P$ and $L_d$ are positive diagonal matrices.

\begin{equation} \label{eq:4-7}
\tau ={{Y}_{d}}(q,\dot{q},{{\dot{q}}_{r}},{{\ddot{q}}_{r}}){{\hat{\theta }}_{d}}-{{K}_{v}}\Delta \dot{q}-{{K}_{p}}\Delta q-{{J}^{T}}(q){{\hat{F}}_{ext}}
\end{equation}

\begin{equation} \label{eq:4-8}
{{\dot{\hat{F}}}_{ext}}={{P}^{-T}}J(q)s
\end{equation}

\begin{equation} \label{eq:4-9}
{{\dot{\hat{\theta }}}_{d}}=-{{L}_{d}}Y_{d}^{T}(q,\dot{q},{{\dot{q}}_{r}},{{\ddot{q}}_{r}})s
\end{equation}

Finally, the closed loop dynamics in equation \eqref{eq:4-10} could be attained by substitution of control signal in equation \eqref{eq:4-6}.

\begin{equation} \label{eq:4-10}
M(q)\dot{s}+C(q,\dot{q})s+{{Y}_{d}}(q,\dot{q},{{\dot{q}}_{r}},{{\ddot{q}}_{r}})\Delta {{\theta }_{d}}+{{K}_{v}}\Delta \dot{q}+{{K}_{p}}\Delta q-{{J}^{T}}(q)\Delta {{F}_{ext}}=0
\end{equation}

In succeeding section, stability of the closed loop dynamics of the system would be investigated.

\subsection{ Stability Analysis of Proposed Adaptive-Sliding Controller}

\newtheorem{theorem}{Theorem}[section]
\begin{theorem}

Adaptive-sliding controller with control signal of equation \eqref{eq:4-7} and adaptive laws in equations \eqref{eq:4-8}  and \eqref{eq:4-9} ensures the stability and convergence of the bending angles and their derivative to origin.
\end{theorem}

\begin{proof}
Lyapunov function of equation \eqref{eq:4-11} is proposed \cite{67} which is a positive definite function, and Its derivative is defined by equation \eqref{eq:4-12}.
\begin{equation} \label{eq:4-11}
V=\frac{1}{2}{{s}^{T}}M(q)s+\frac{1}{2}\Delta \theta _{d}^{T}L_{d}^{-1}\Delta {{\theta }_{d}}+\frac{1}{2}\Delta {{q}^{T}}({{K}_{p}}+\Lambda {{K}_{v}})\Delta q+\frac{1}{2}\Delta F_{ext}^{T}P\,\Delta {{F}_{ext}}
\end{equation}

\begin{equation} \label{eq:4-12}
\dot{V}={{s}^{T}}M(q)\dot{s}+\frac{1}{2}{{s}^{T}}\dot{M}(q)s-\Delta \theta _{d}^{T}L_{d}^{-1}{{\dot{\hat{\theta }}}_{d}}+\Delta {{q}^{T}}({{K}_{p}}+\Lambda {{K}_{v}})\Delta \dot{q}+\Delta \dot{F}_{ext}^{T}P\,\Delta {{F}_{ext}}
\end{equation}

Since $\dot{M}(q)-2C(q,\dot{q})$ is skew-symmetric, substitution of closed loop dynamics and adaptive laws would results in equation \eqref{eq:4-13} for derivative of Lyapunov function.

\begin{equation} \label{eq:4-13}
\dot{V}=-{{s}^{T}}({{K}_{v}}\Delta \dot{q}+{{K}_{p}}\Delta q)+\Delta {{q}^{T}}({{K}_{p}}+\Lambda {{K}_{v}})\Delta \dot{q}
\end{equation}

Rewriting equation \eqref{eq:4-2} with the help of equation \eqref{eq:4-3}, we could obtain a different form for sliding vector as in equation \eqref{eq:4-14}. Substituting \eqref{eq:4-14} into \eqref{eq:4-13} gives us the final derivative of Lyapunov function.

\begin{equation} \label{eq:4-14}
s = \Delta \dot{q} + \Lambda \Delta_q 
\end{equation}

\begin{equation} \label{eq:4-15}
\dot{V}=-\Delta {{\dot{q}}^{T}}{{K}_{v}}\Delta \dot{q}-\Lambda \Delta {{q}^{T}}{{K}_{p}}\Delta q\le 0 
\end{equation}

Due to the fact that inertia matrix is positive definite, proposed Lyapunov function is positive definite as well, and according to \eqref{eq:4-15}, $\dot{V}$ is negative semi-definite, which means $V$ is bounded. Boundedness of Lyapunov function leads to boundedness of its variables including $s$, $\Delta{\theta}_d$, $\Delta{q}$, $\Delta{F}_{ext}$. Therefore, since $q_d$, $\theta_d$ and $F_{ext}$ are bounded, $\hat{\theta}_d$, $q$ and $\hat{F}_{ext}$  remain bounded too. Considering \eqref{eq:4-3}, $\dot{q}_r$ is bounded, which according to \eqref{eq:4-2}, it is implied that $\dot{q}$ is bounded. Differentiation of \eqref{eq:4-3}  with respect to time brings about the boundedness of $\ddot{q}_r$. In addition, corresponding to closed loop dynamics of \eqref{eq:4-10}, $\dot{s}$ is determined to be bounded, which based on \eqref{eq:4-4}, results in boundedness of $\ddot{q}$.\\
Taking advantage of Barbalats's lemma, we have to prove the uniform continuity of $\dot{V}$ through the boundedness of second order derivative of Lyapunov function.

\begin{equation} \label{eq:4-16}
\ddot{V}=-2\Delta {{\dot{q}}^{T}}{{K}_{v}}\Delta \ddot{q}-2\Lambda \Delta {{q}^{T}}{{K}_{p}}\Delta \dot{q}
\end{equation}

With respect to boundedness of all its variables, \eqref{eq:4-16} is proven to be bounded, which culminates in uniform continuity of $\dot{V}$. As a result, the adaptive-sliding tracking control law \eqref{eq:4-7} and parameter adaptive laws \eqref{eq:4-8} and \eqref{eq:4-9} guarantee the stability of closed loop system and convergence of its position and velocity tracking error to the origin.\\
\end{proof}

In following section, results of implementation of adaptive-sliding controller in experimental testing will be presented.

\section{Performance Evaluation of Adaptive Controller by Experiments} 

Throughout this section, the performance of proposed controller is assessed in free motion and interaction with the environment. Moreover, the responses of PID controller are reported as benchmark. \\
\subsection{Free Motion}

For free motion condition, the objective of adaptive-sliding controller is curvature trajectory tracking of a reference input $q_d=0.8+0.4 sin (2t)$ for both segments. Coefficients of control signal \eqref{eq:4-7} are selected as in Table \ref{tab.4.1}.
\begin{table}[ht]
\caption{Adaptive controller gains for experiments} 
\centering 
\begin{tabular}{c c} 

\hline\hline 
			
				 Gains for controller& Magnitude \\[0.5ex] 
				 \hline 
			$K_V$	  & $0.002$
				 \\
			
				$K_P$  & $0.015$
				 \\
			
			$L_d$	  & $0.001$ \\
				
			$\Lambda$	  & 10
				 \\[1ex]
				\hline 
			\end{tabular} 
		\label{tab.4.1}
	
\end{table}

\begin{figure}[H]
	\centering
\captionsetup{justification=centering}		\includegraphics[width=163mm,height=77mm]{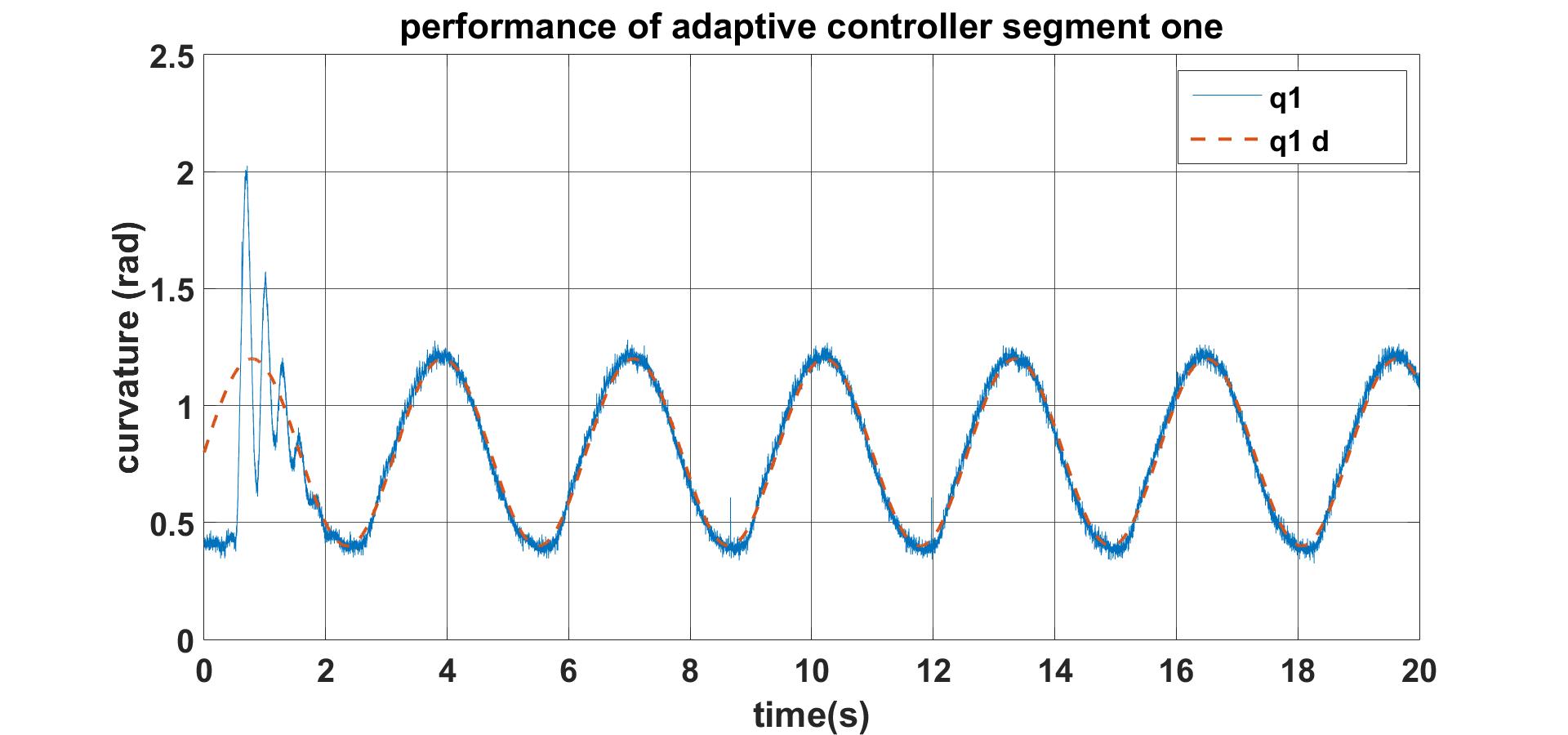}
\caption{ Tracking the reference trajectory of upper bending actuator by adaptive controller }
\label{fig:4-1}
\end{figure}

\begin{figure}[H]
	\centering
\captionsetup{justification=centering}		\includegraphics[width=163mm,height=80mm]{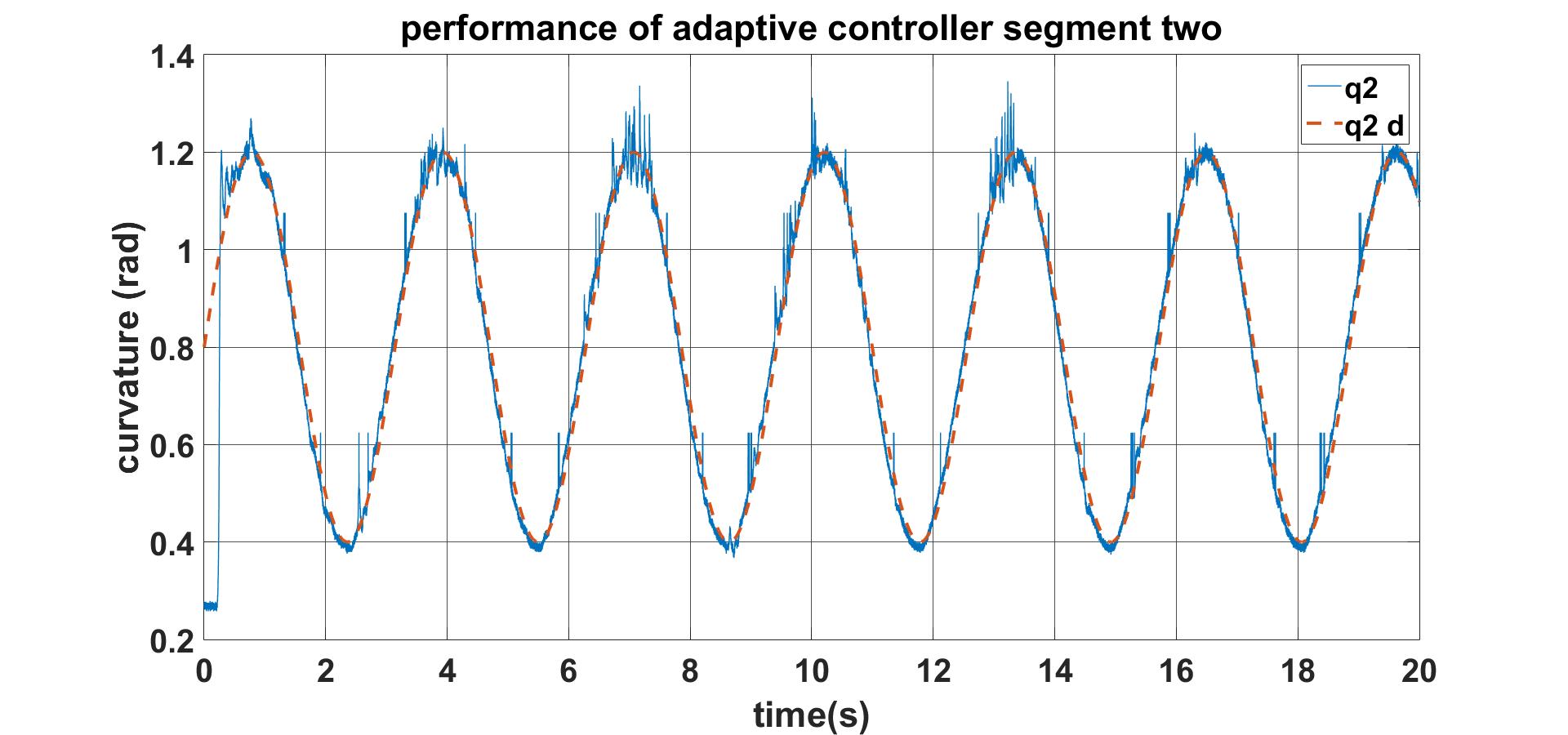}
\caption{Tracking the reference trajectory by lower bending actuator through adaptive controller}
\label{fig:4-2}
\end{figure}

\begin{figure}[H]
	\centering
\captionsetup{justification=centering}		\includegraphics[width=163mm,height=80mm]{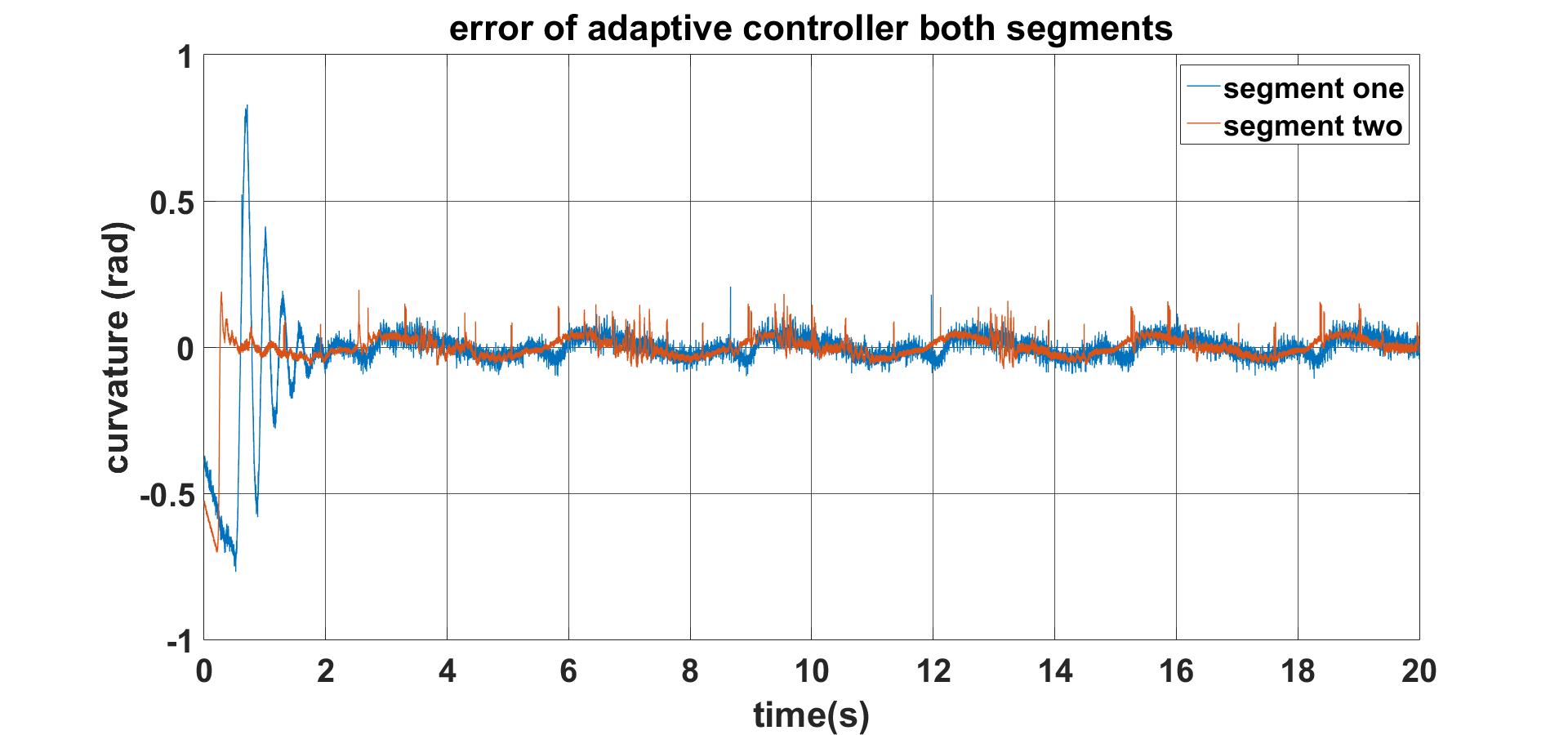}
\caption{ Error of tracking the reference trajectory through adaptive controller for both actuators}
\label{fig:4-3}
\end{figure}

\begin{figure}[H]
	\centering
\captionsetup{justification=centering}		\includegraphics[width=163mm,height=80mm]{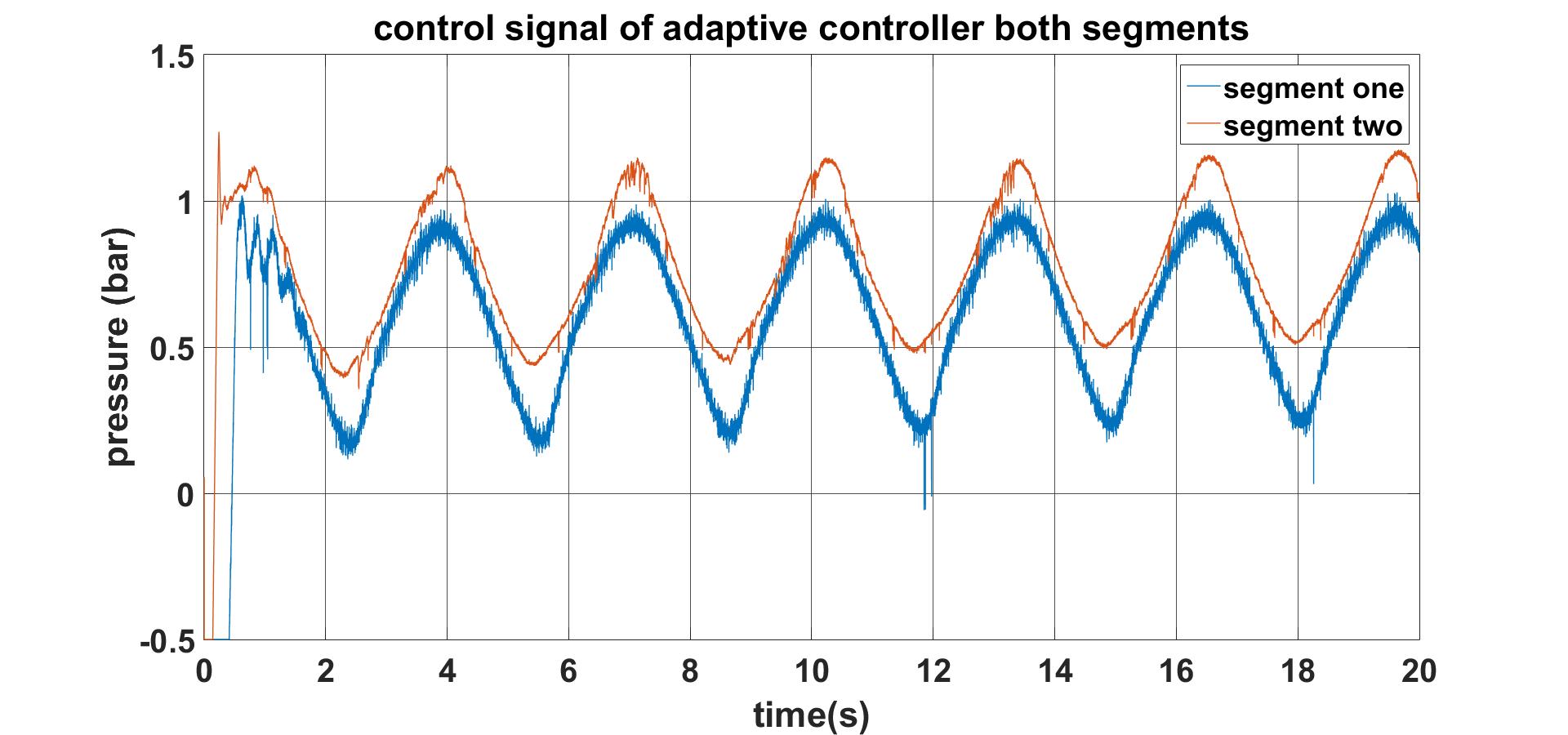}
\caption{ Control effort for both actuators in tracking the desired trajectory by adaptive controller}
\label{fig:4-4}
\end{figure}

Tracking performance for both segments, their bending angle errors and control efforts are depicted in Figures \ref{fig:4-1} to \ref{fig:4-4}. Accordingly, the proposed controller meets the expectations although its transient response might seem unsatisfactory. This can be explicated by low quantity for damping coefficients of bending actuators. Besides, despite using low-pass filter, the output voltage of flex sensors, as mentioned previously, is yet followed by high frequency noise, which can drastically affect the performance of the controller.

Furthermore, overshoot percentage of segment one is more considerable in comparison to that of segment two. This issue could be accounted for by the fact that segment one carries an excessive weight which is due to lower bending actuator's mass. Hence, this extra load on segment one is capable of yielding great momentum in response to a change in pneumatic pressure.\\

\subsection{Interaction With The Environment}
\begin{figure}[H]
	\centering
\captionsetup{justification=centering}		\includegraphics[width=70mm,height=70mm]{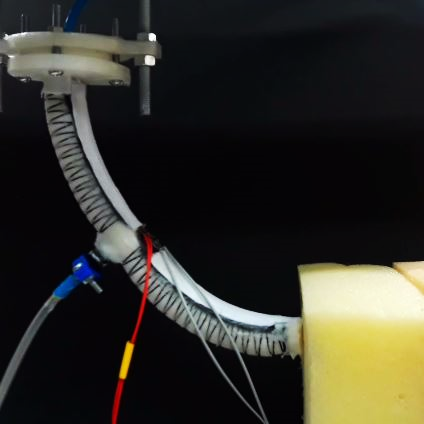}
\caption{ Using a resilient object for interaction with the environment experiments}
\label{fig:4-9}
\end{figure}

As shown in figure \ref{fig:4-9}, in order to provide the soft finger with an opportunity to experience interaction with the environment, we used a resilient obstacle that intersects the trajectory of the finger's tip. The finger, however, is allowed to penetrate into the obstacle. The objective of adaptive-sliding controller, therefore, is to compensate the resisting wrenches exerted on the finger. This can be executed through the adaptive law \eqref{eq:4-8} for external disturbances.

\begin{figure}[H]
	\centering
\captionsetup{justification=centering}		\includegraphics[width=160mm,height=76mm]{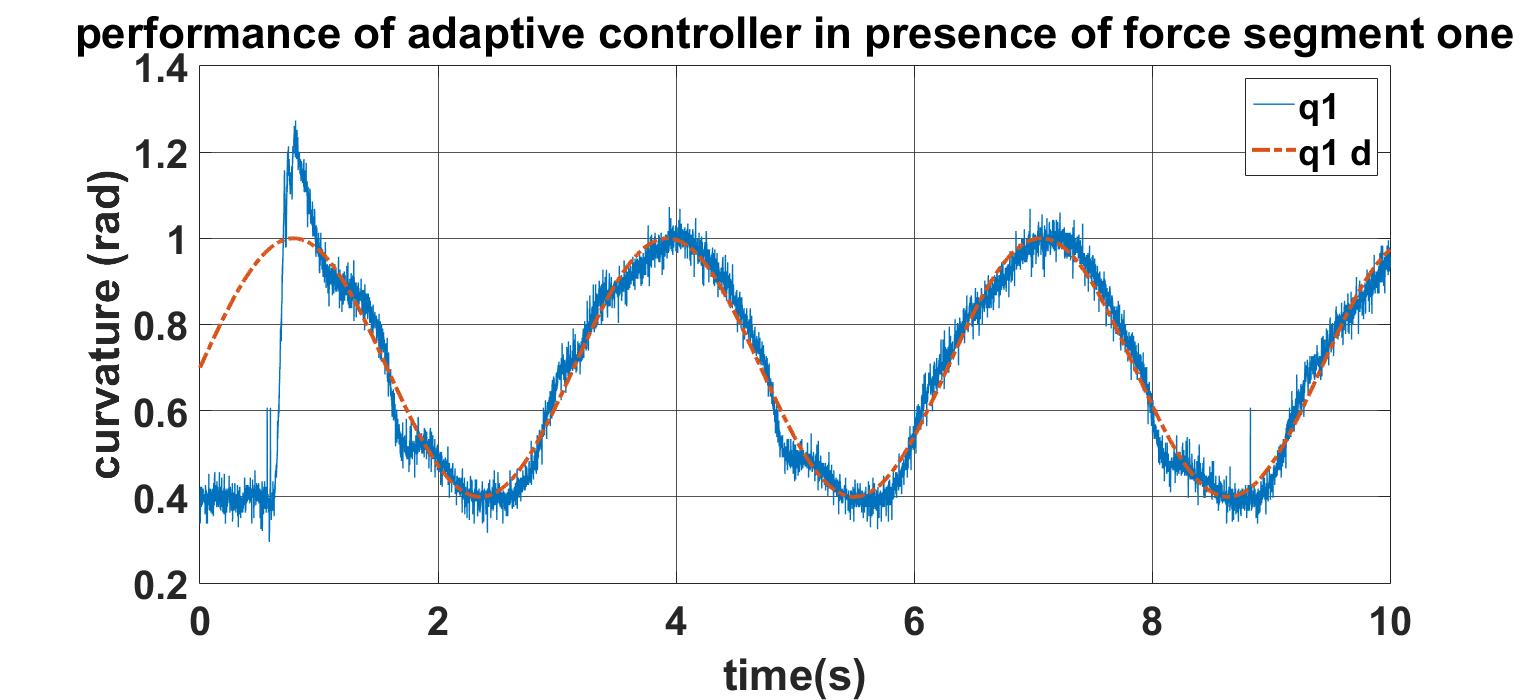}
\caption{ Adaptive controller tracking for upper actuator in interaction with the environment }
\label{fig:4-10}
\end{figure}

\begin{figure}[H]
	\centering
\captionsetup{justification=centering}		\includegraphics[width=163mm,height=80mm]{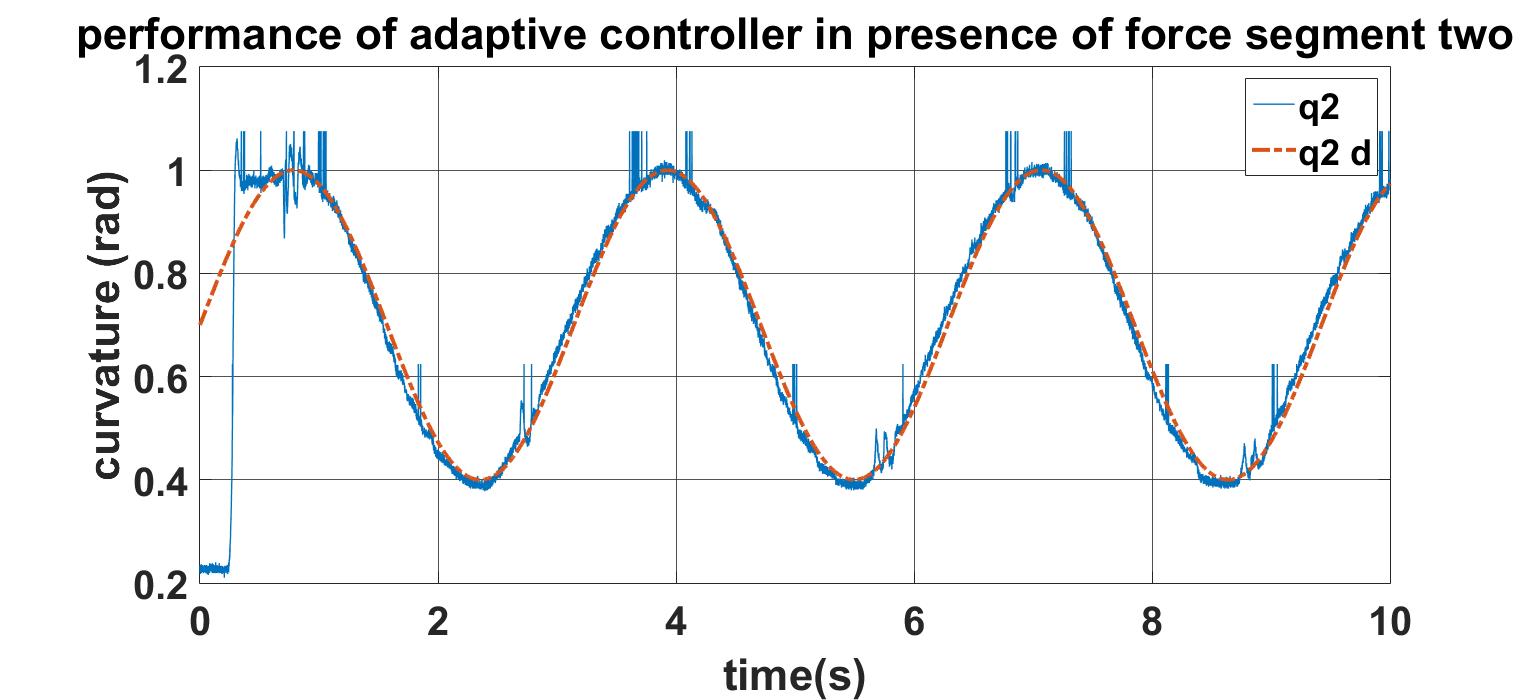}
\caption{Adaptive controller tracking for lower actuator in interaction with the environment}
\label{fig:4-11}
\end{figure}

\begin{figure}[H]
	\centering
\captionsetup{justification=centering}		\includegraphics[width=163mm,height=80mm]{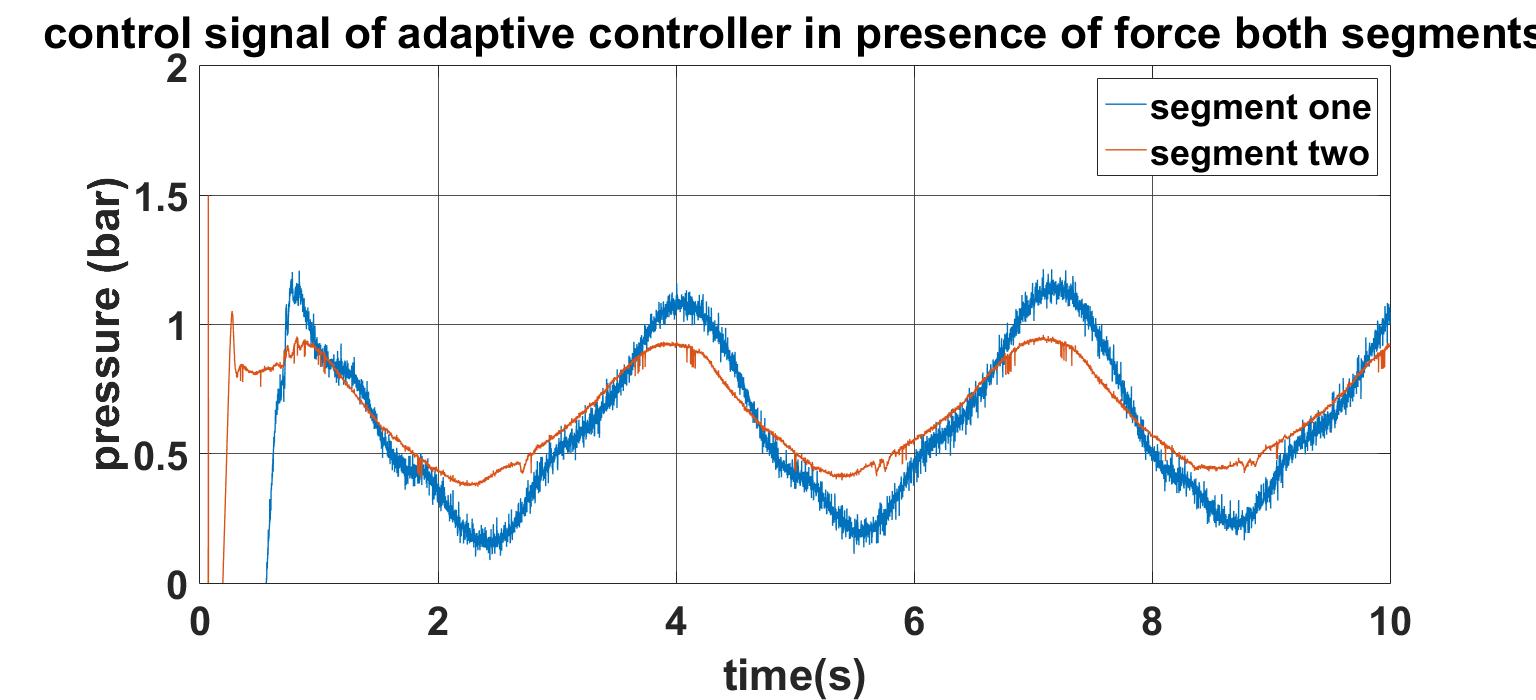}
\caption{Control effort for both actuators in tracking the desired trajectory by adaptive controller in interaction with the environment }
\label{fig:4-12}
\end{figure}

\begin{figure}[H]
	\centering
\captionsetup{justification=centering}		\includegraphics[width=163mm,height=80mm]{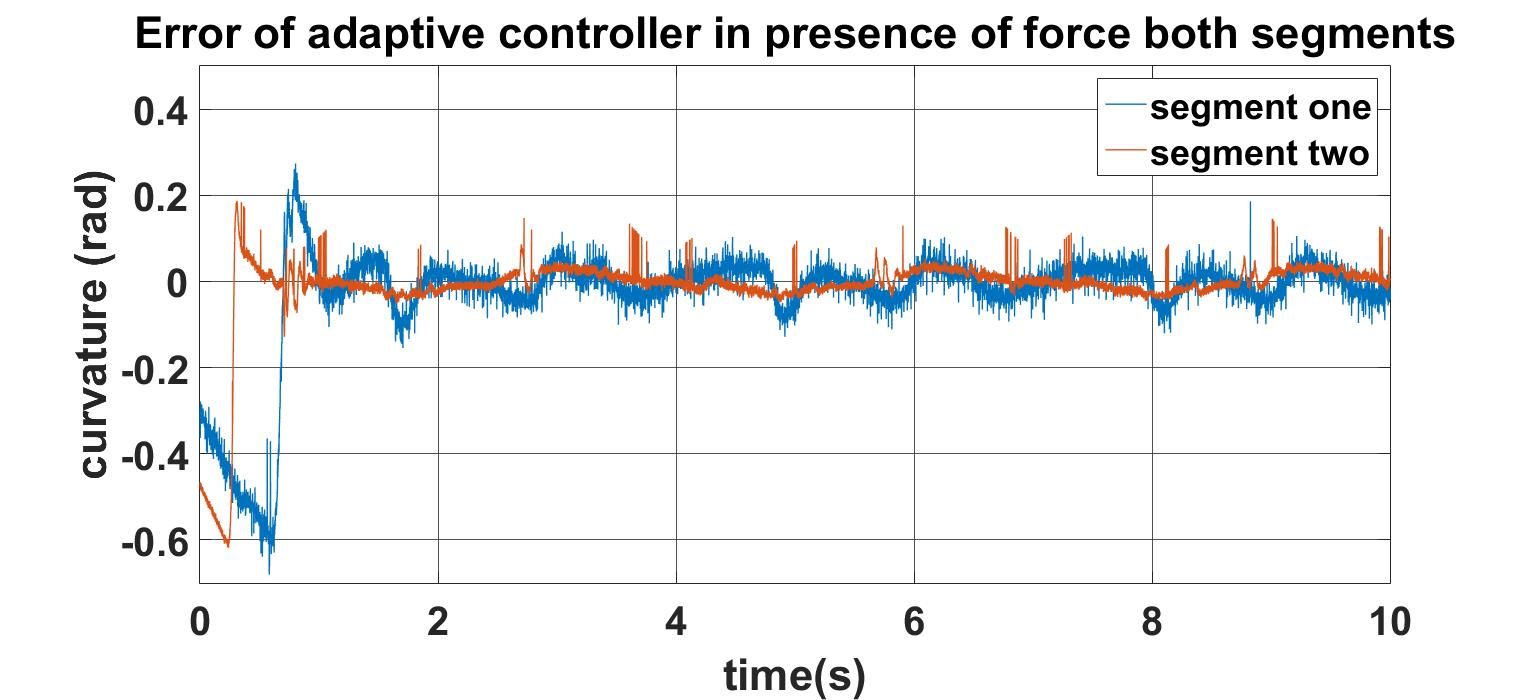}
\caption{ Error of tracking the reference trajectory through adaptive controller for both actuators in interaction with the environment}
\label{fig:4-13}
\end{figure}

Tracking performance of reference signal of $q_d=0.7+0.3sin(2t)$ for both segments, their bending angle errors and control efforts under interaction with the environment are depicted in Figures \ref{fig:4-10} to \ref{fig:4-13}.


\section{Performance Evaluation of PID Controller}

As stated earlier, PID controller is implemented for controlling the bending curvature of the soft finger as well so as to compare its results with those of proposed adaptive-sliding controller. Likewise, both free and constrained motion of the system are investigated.\\

\subsection{Free Motion}
\begin{figure}[H]
	\centering
\captionsetup{justification=centering}		\includegraphics[width=163mm,height=80mm]{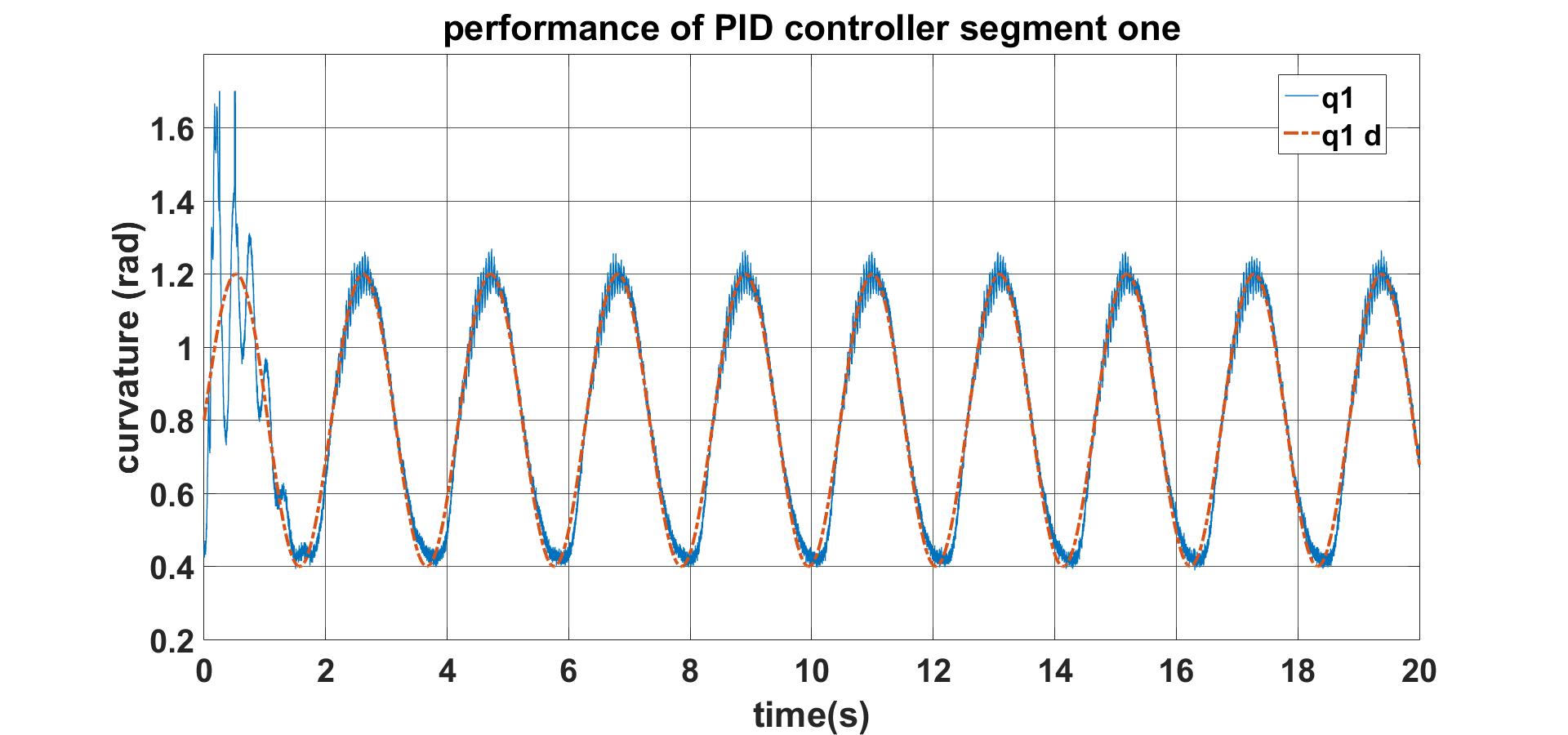}
\caption{Tracking the reference trajectory of upper bending actuator by PID controller}
\label{fig:4-5}
\end{figure}

\begin{figure}[H]
	\centering
\captionsetup{justification=centering}		\includegraphics[width=163mm,height=80mm]{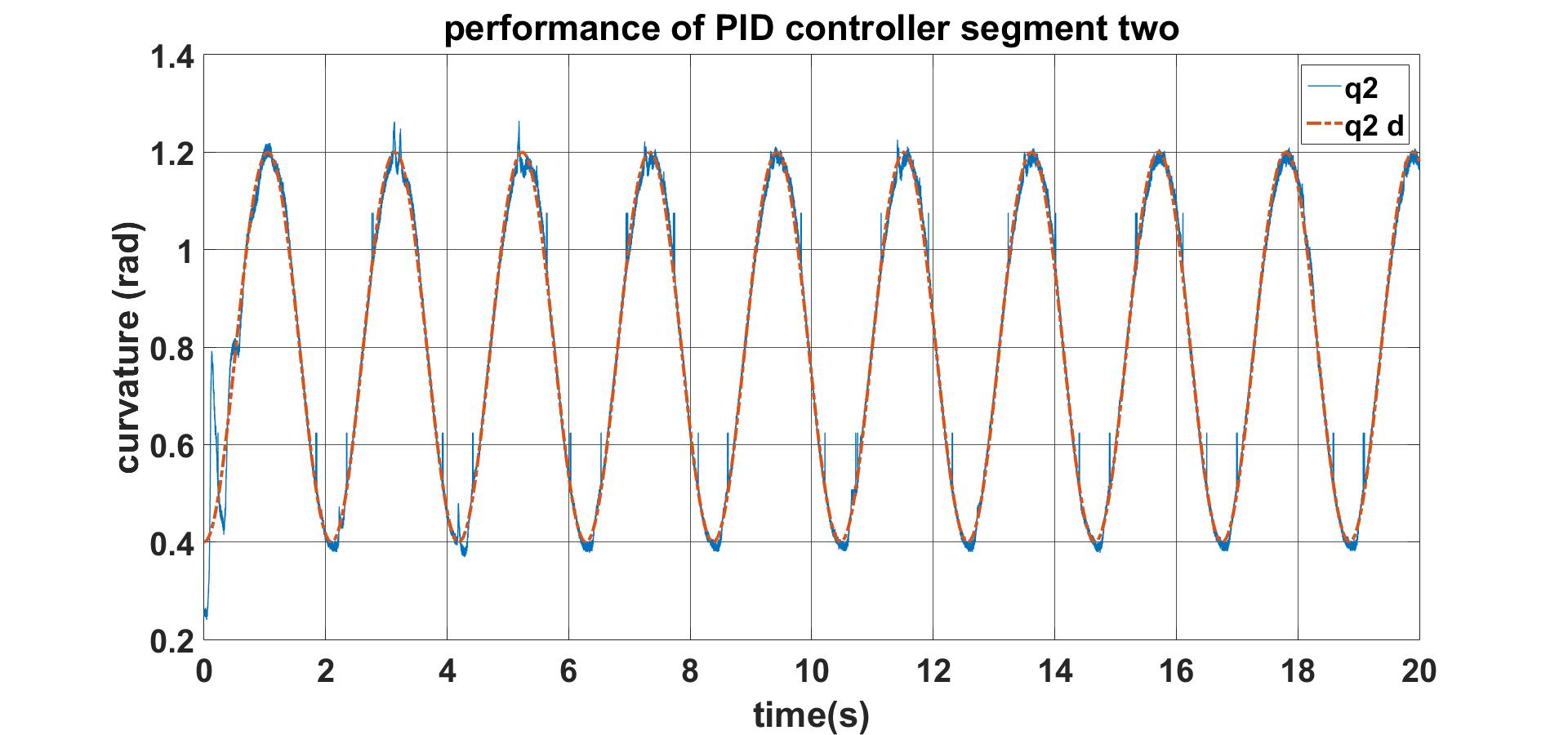}
\caption{Tracking the reference trajectory of lower bending actuator by PID controller}
\label{fig:4-6}
\end{figure}

\begin{figure}[H]
	\centering
\captionsetup{justification=centering}		\includegraphics[width=163mm,height=80mm]{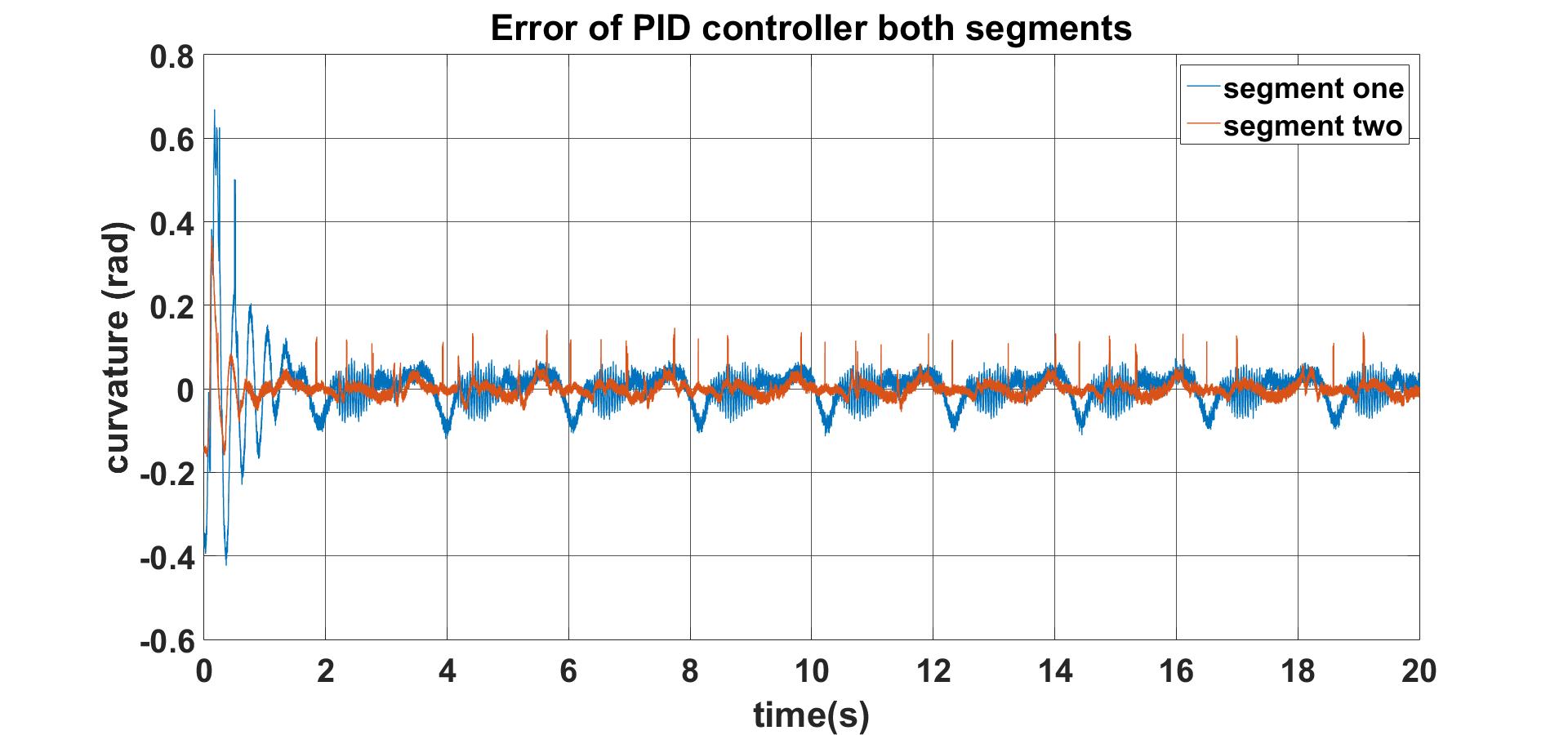}
\caption{Error of tracking the reference trajectory through PID controller for both actuators}
\label{fig:4-7}
\end{figure}

\begin{figure}[H]
	\centering
\captionsetup{justification=centering}		\includegraphics[width=153mm,height=72mm]{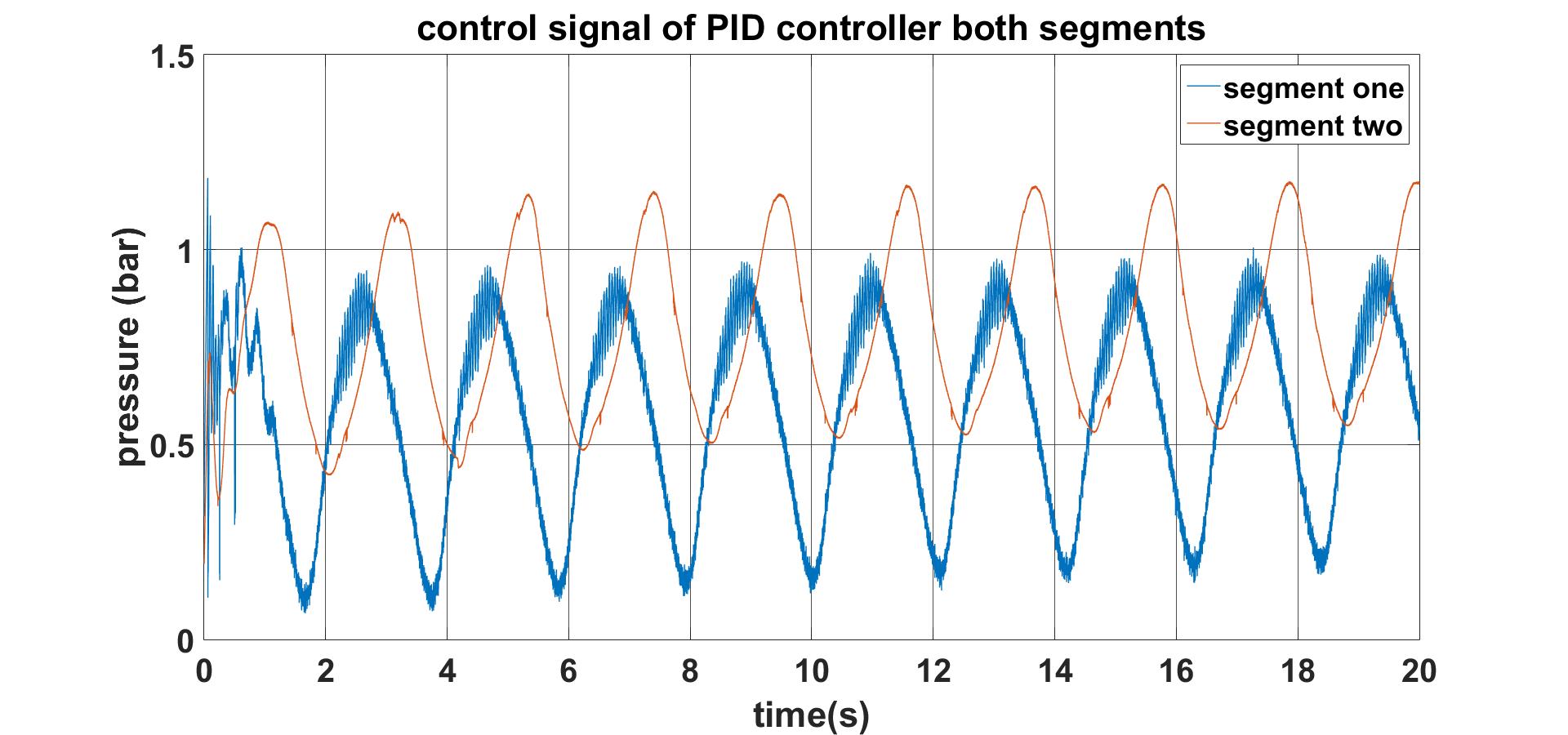}
\caption{Control effort for both actuators in tracking the desired trajectory by PID controller}
\label{fig:4-8}
\end{figure}

The objective of PID controller is tracking the reference inputs $q_d=0.8+0.4sin(3t)$ and $q_d=0.8+0.4sin(3t+1.57)$ by upper and lower bending actuators respectively. Coefficients of PID controller are selected as in Table \ref{tab.4.2} and its performance is depicted through Figures \ref{fig:4-5} to \ref{fig:4-8}.

\begin{table}[ht]
\caption{PID controller gains for experiments} 
\centering 
\begin{tabular}{c c} 

\hline\hline 
			
				 Gains for controller& Magnitude \\[0.5ex] 
				 \hline 
			$k_p$	  & $0.03$
				 \\
			
				$k_I$  & $1.2$
				 \\
			
			$k_{d1}$	  & $0.004$ \\
				
			$k_{d2}$	  & 0.0005
				 \\[1ex]
				\hline 
			\end{tabular} 
		\label{tab.4.2}
	
\end{table}

\subsection{Interaction With The Environment}
\begin{figure}[H]
	\centering
\captionsetup{justification=centering}		\includegraphics[width=153mm,height=72mm]{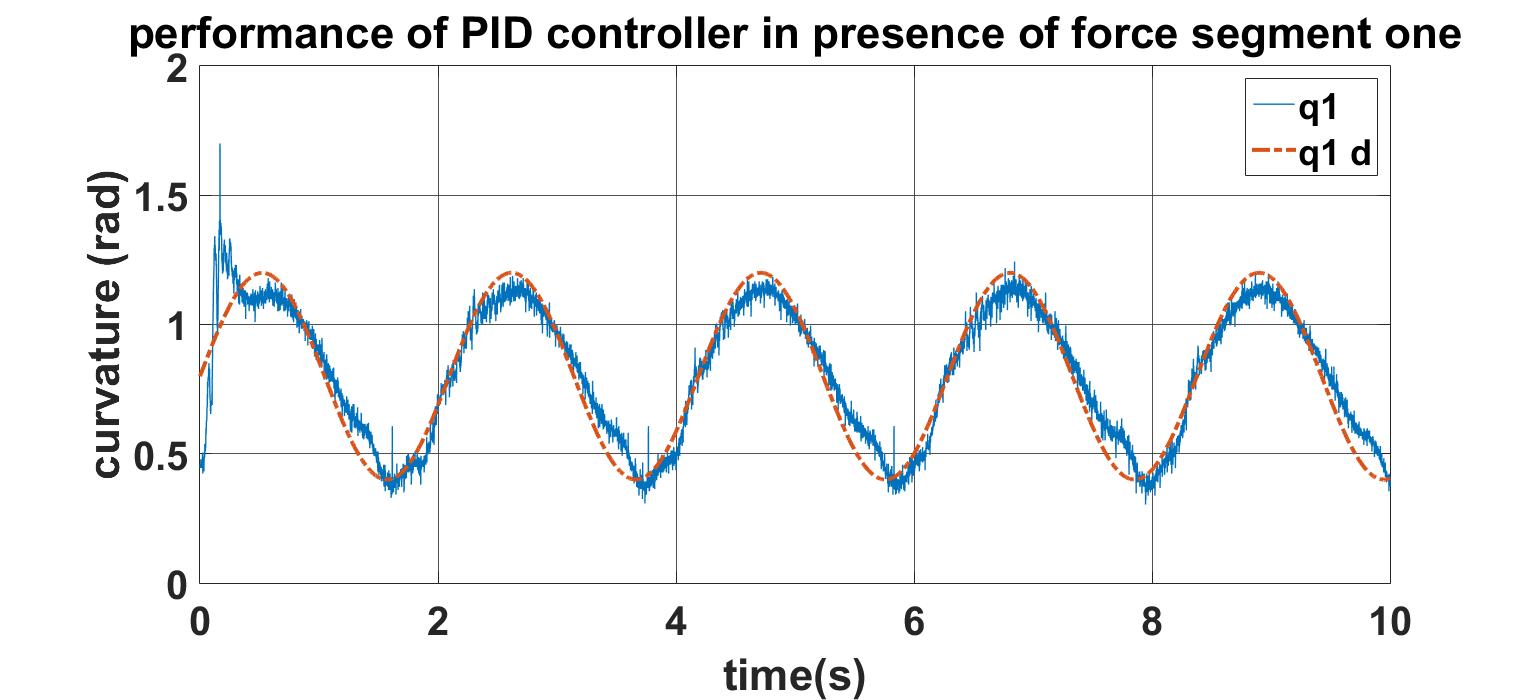}
\caption{PID controller tracking for upper actuator in interaction with the environment}
\label{fig:4-14}
\end{figure}

\begin{figure}[H]
	\centering
\captionsetup{justification=centering}		\includegraphics[width=163mm,height=80mm]{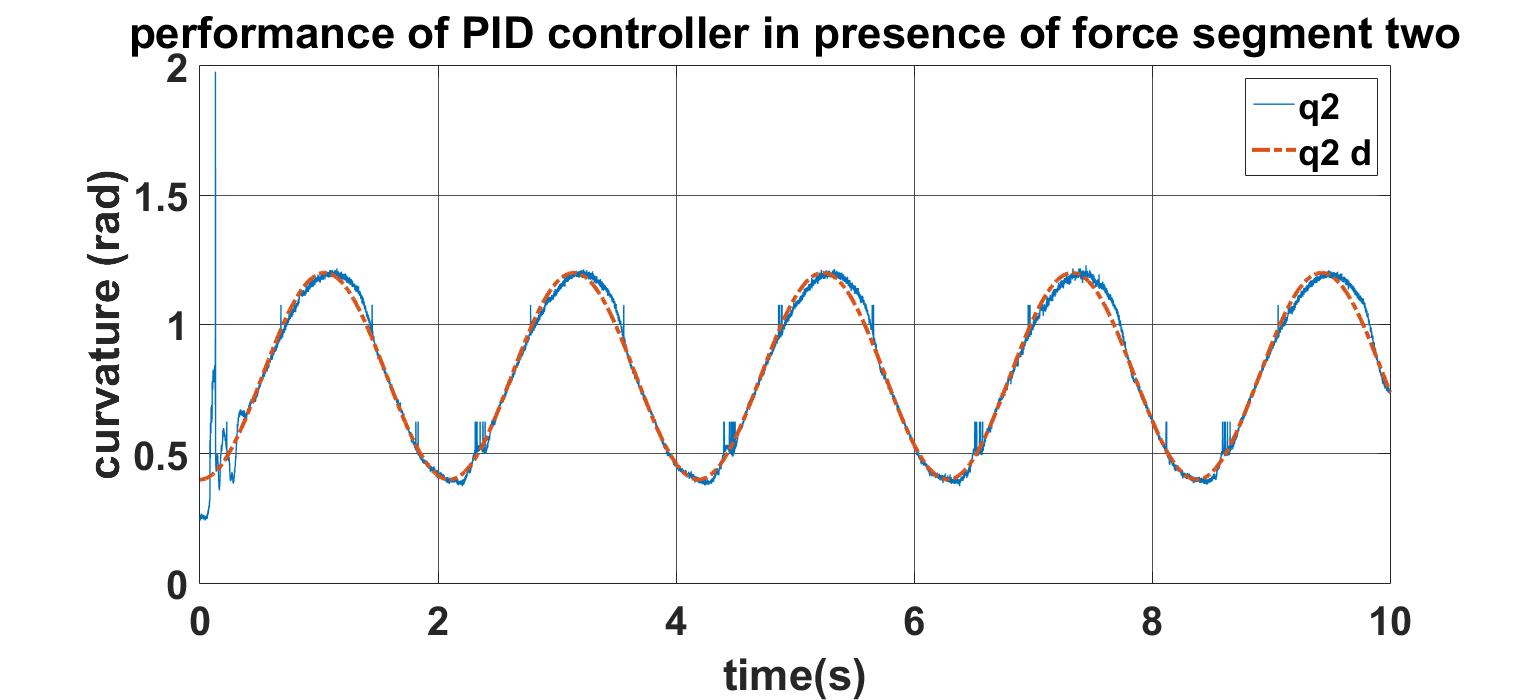}
\caption{PID controller tracking for lower actuator in interaction with the environment}
\label{fig:4-15}
\end{figure}

\begin{figure}[H]
	\centering
\captionsetup{justification=centering}		\includegraphics[width=163mm,height=80mm]{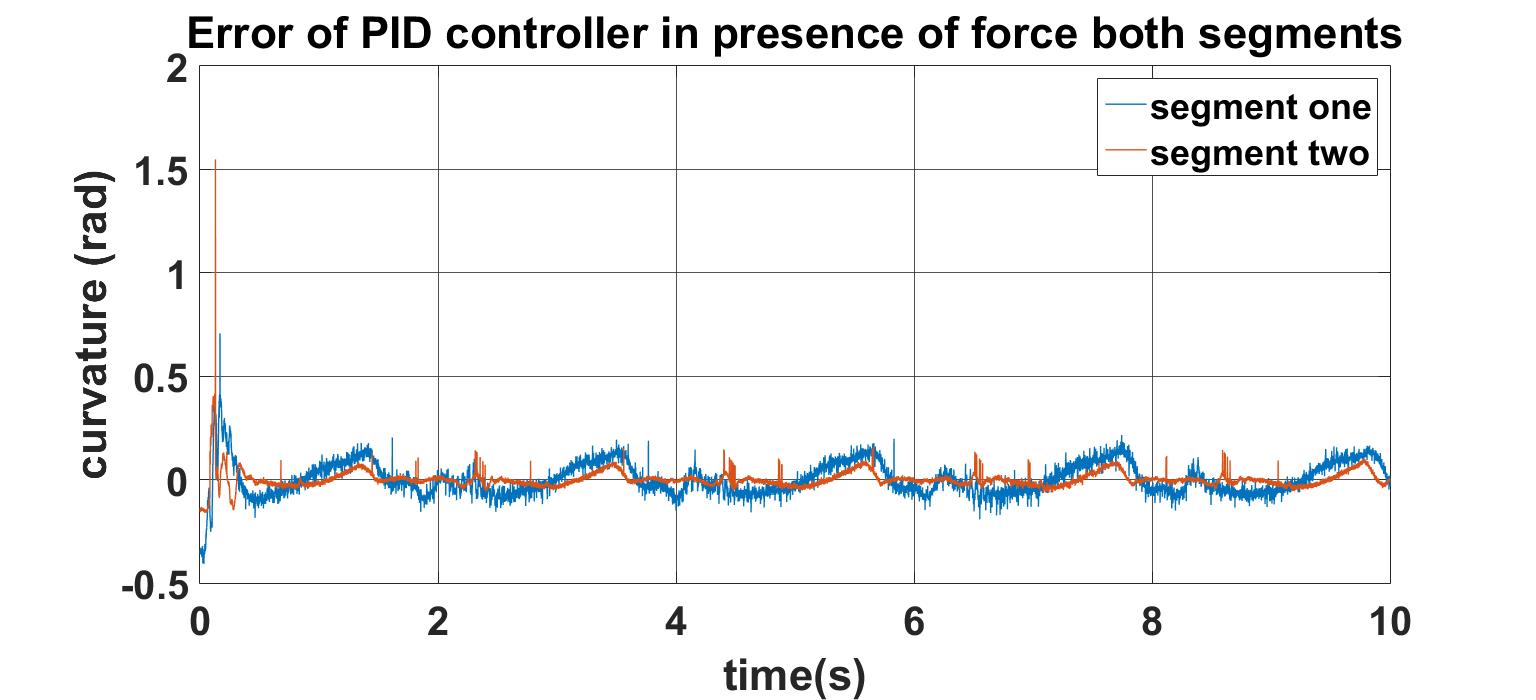}
\caption{Error of tracking the reference trajectory through PID controller for both actuators in interaction with the environment}
\label{fig:4-17}
\end{figure}

\begin{figure}[H]
	\centering
\captionsetup{justification=centering}		\includegraphics[width=153mm,height=75mm]{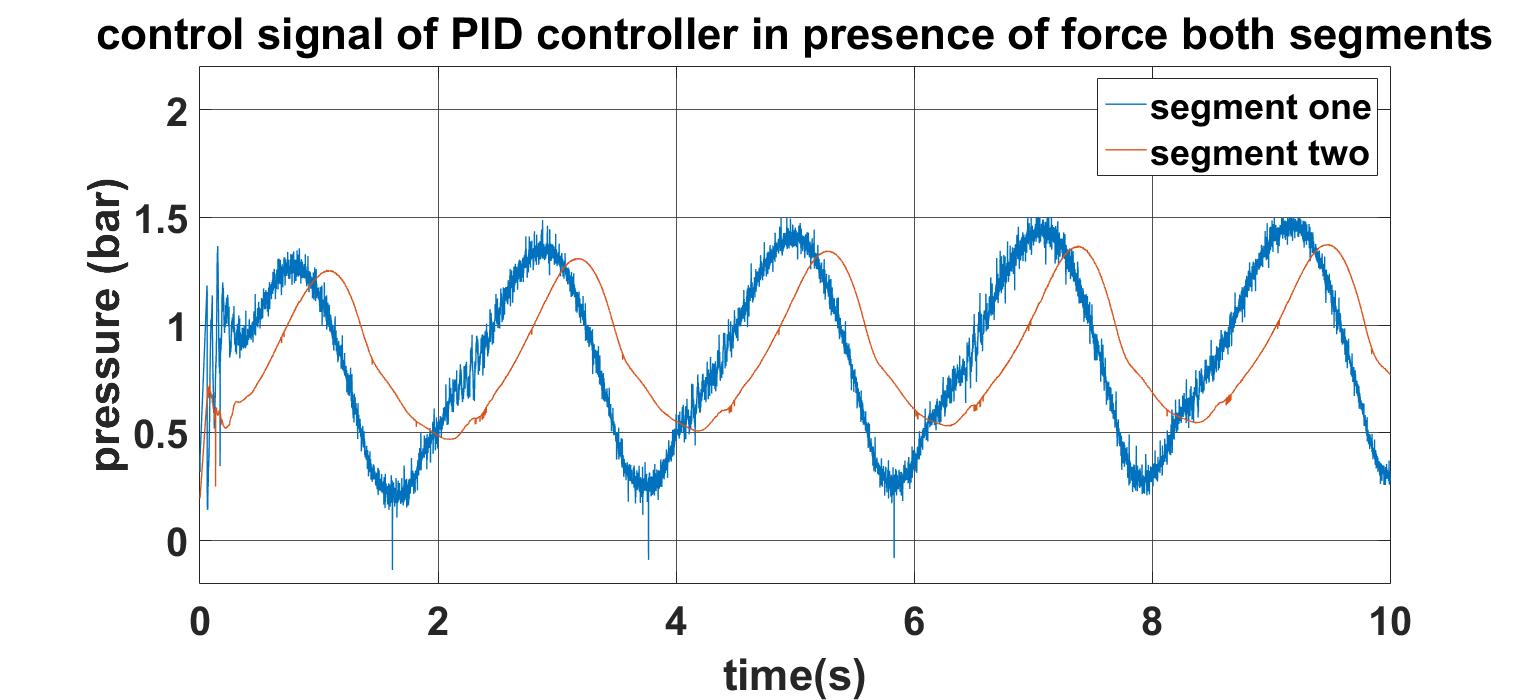}
\caption{Control effort for both actuators in tracking the desired trajectory by PID controller in interaction with the environment }
\label{fig:4-16}
\end{figure}

Tracking the same reference inputs as in free motion, PID controller exhibits the responses of Figures \ref{fig:4-14} and \ref{fig:4-15}, and its curvature error along with control efforts of each actuator are depicted in figures \ref{fig:4-17} and \ref{fig:4-16} respectively.

\begin{table}[h]
\caption{ $RMSE$ for both controllers in each of the operating modes} 
\centering 
\begin{tabular}{c c c c} 
\hline\hline 
Controller & Mode & $RMSE$ upper actuator $rad$ & $RMSE$ lower actuator $rad$
\\ [0.5ex]
\hline 
& Free Motion & 0.0323 & 0.0303  \\[-1ex]
\raisebox{1.5ex}{Adaptive-sliding} & Constrained Motion& {$0.0328$}
& 0.0226  \\[1ex]
& Free Motion & 0.0342 & 0.0181  \\[-1ex]
\raisebox{1.5ex}{PID} & Constrained Motion& {$0.0747$}
& 0.0303  \\[1ex]

\hline 
\end{tabular}
\label{tab.4.3}
\end{table}

\begin{figure}[H]
	\centering
\captionsetup{justification=centering}		\includegraphics[width=143mm,height=70mm]{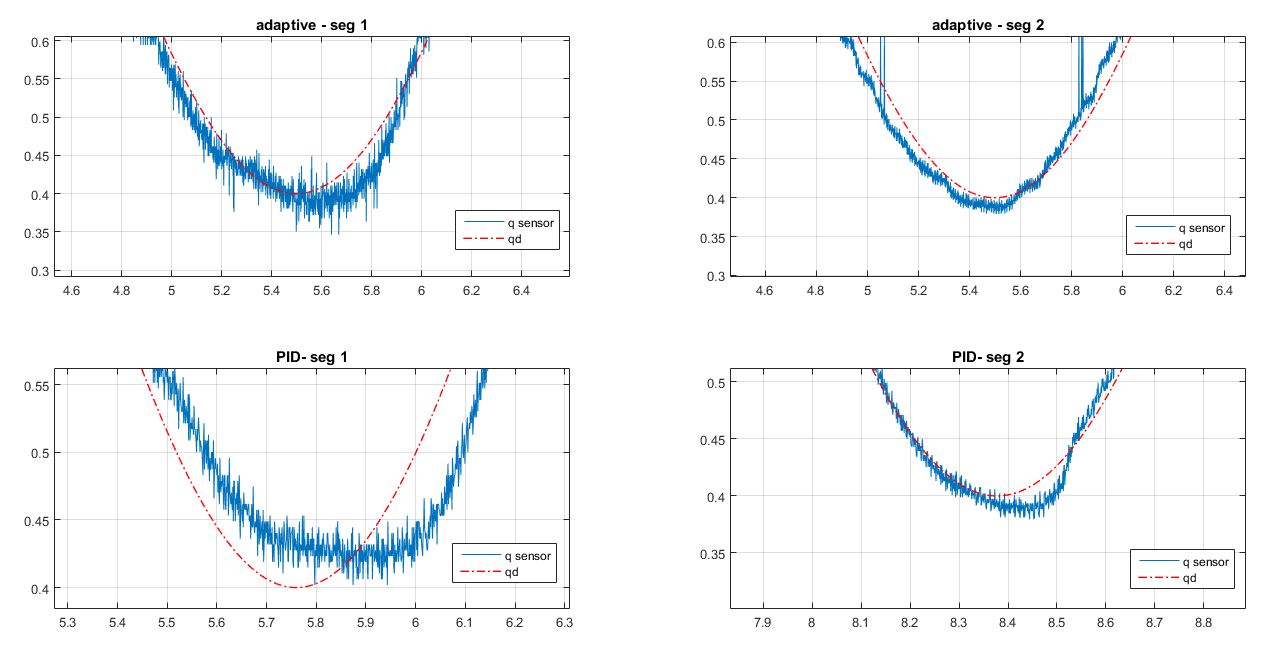}
\caption{Comparing the performance of PID and Adaptive controllers in free motion for both actuators}
\label{fig:4-20}
\end{figure}

\begin{figure}[H]
	\centering
\captionsetup{justification=centering}		\includegraphics[width=143mm,height=70mm]{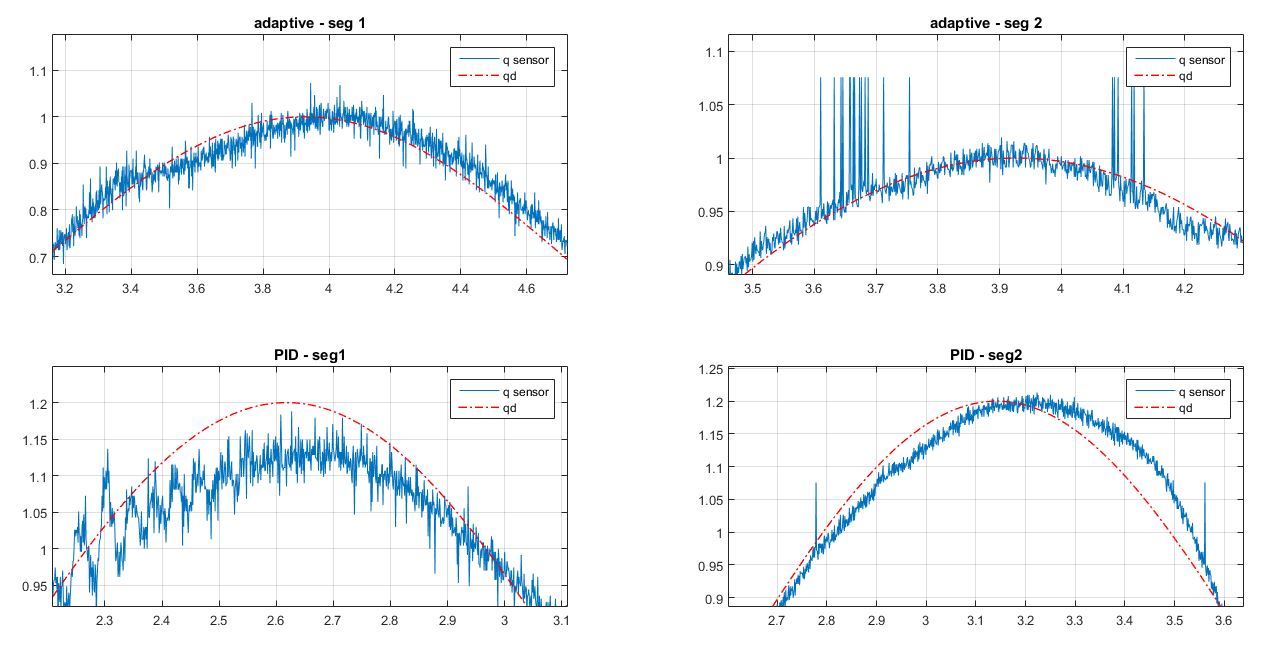}
\caption{Comparing the performance of PID and Adaptive controllers in interaction with the environment for both actuators}
\label{fig:4-21}
\end{figure}

Comparing the results of both controllers with each other, one might at first sight interpret that PID controller is more capable of tracking the desired trajectory; however, carrying out a more thorough inspection proves otherwise. The distinguishing characteristic of adaptive-sliding controller in free and constrained motion tracking with comparison to PID are illuminated in Figures \ref{fig:4-20}  and \ref{fig:4-21} respectively. Moreover, Table \ref{tab.4.3} represents the $RMSE$ of examined controllers without taking the transient responses into account.   

\clearpage
\chapter{Interaction With The Environment And Cartesian Control} \label{sec:nnmf}
\clearpage
%
Having observed the performance of adaptive controller interacting with the environment in the preceding chapter, we were encouraged to investigate the effects of external force on the behavior of soft finger more profoundly. Therefore, we conducted a series of experiments so as to measure the position of the finger's tip in Cartesian space while the tip force was being estimated.\\

\section{ Effect of External Force on Tip Position }
\begin{figure}[H]
	\centering
	\captionsetup{justification=centering}		\includegraphics[width=140mm,height=70mm]{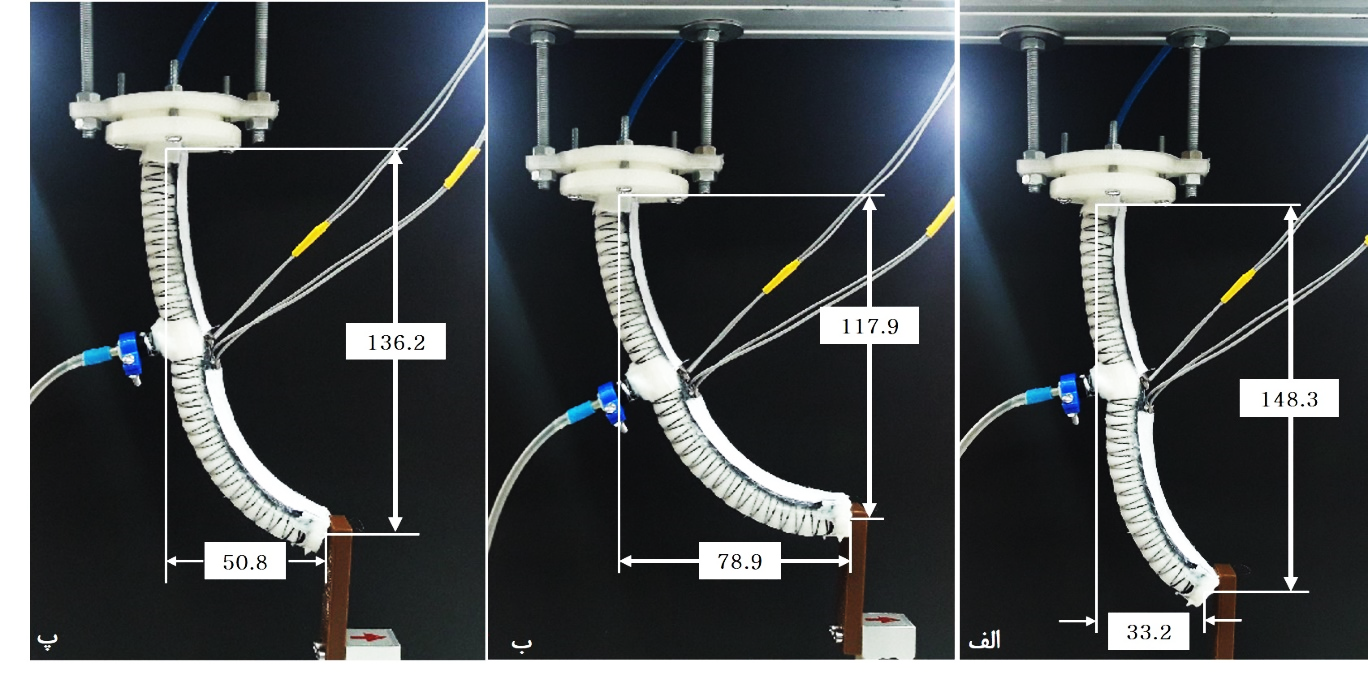}
\caption{Obtaining the tip position of soft finger while measuring the force by load cell}
\label{fig:3-14}
\end{figure}

In order to assess merely the impact of external force, the amount of operating pressure were fixed, and, by means of a mobile load cell, we intercepted the trajectory of the tip while recording the exerted force. Equipped with a slider mounted on it, load cell were able to move horizontally and vertically; so we repeated the experiment by changing the location of load cell and with different pressures. In other words, as shown in Figure \ref{fig:3-14}, we recorded the position of the tip through image processing and applied force by load cell.

\begin{figure}[H]
	\centering
	\captionsetup{justification=centering}		\includegraphics[width=115mm,height=107mm]{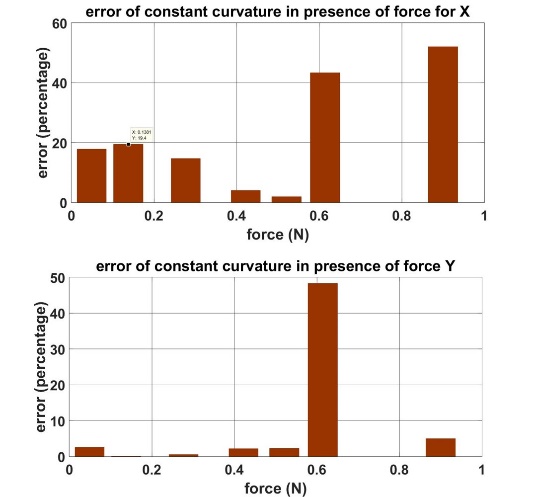}
\caption{Position error in contact with the environment for vertical and Horizontal directions}
\label{fig:3-15}
\end{figure}

Hence, having knowledge about the tip position, applied force and operating pressure, we can detect the error of constant curvature model subjected to external force. Inputting force and pressure data acquired from experiments into our proposed model, we could attain bending curvatures. Subsequently, inserting these curvatures into forward kinematics generates the position of the tip based on constant curvature hypothesis, which can be contrasted with the position gained by image processing. Consequently, this deviations between predictions and experimental data of tip position for different forces are plotted in Figure \ref{fig:3-15}. It is observed that the increase in force leads to the rise in error especially the horizontal direction.

This difference between experiments and simulations, some scholars have argued, might stem from the disruption in constant curvature assumption. As a matter of fact, in presence of force at the tip, the bending angles of actuators could not be defined through a single curve with a constant radius; instead, they follow multiple curves with diverse radiuses. Although this theory could be applied to actuators with low ratio of thickness to length, we should trace another origin for our case since our finger's thickness is high enough to maintain a single curve.\\

\begin{figure}[H]
	\centering
\captionsetup{justification=centering}		\includegraphics[width=0.54\linewidth,height=0.315\textheight]{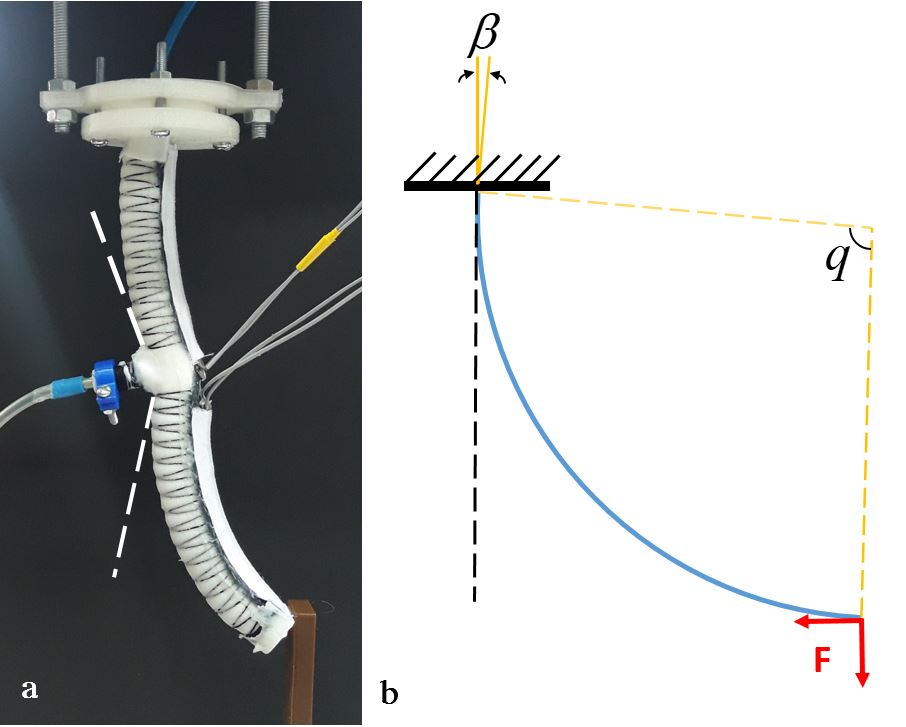}
\caption{Effect of external force a) on two-segmented soft finger. b) on a single bending actuator \cite{68} }
\label{fig:5-1}
\end{figure}

According to \cite{68} and Figure \ref{fig:5-1}, an applied force of $F$ can yield a moment at the base or the junction of our soft finger, which culminates in its bending at these spots and is indicated by $\beta$. This phenomenon can be attributed to the fact that the junction and the base of the finger is devoid of pneumatic pressure, so the finger is more conducive to bend at these regions. As depicted in Figure \ref{fig:5-1}, Tangent lines at the junction do not coincide, which elucidates the aforementioned deviation.\\
The original study \cite{68} addresses the issue by identifying the deflection of $\beta$ based on the robot's material and the exerted force. Thereafter, the article utilizes a rotational matrix, which is function of $\beta$, by which pre multiplies the forward kinematics of the soft robot. Thus, a new forward kinematics and dynamic equation would be derived for the soft finger, but the process of identification must be repeated for different forces, which seems inefficient. 

\section{A Controller With Compensation of Kinematic Uncertainty }

\begin{figure}[H]
	\centering
\captionsetup{justification=centering}		\includegraphics[width=123mm,height=61mm]{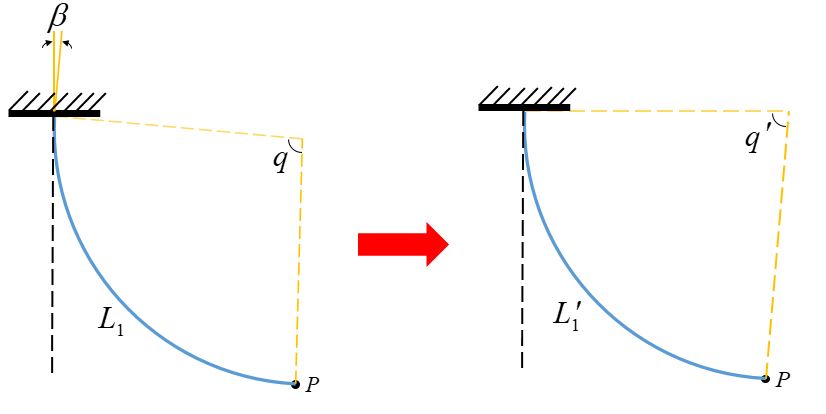}
\caption{Reaching to the same position without deflection by appropriate estimation of length and curvature}
\label{fig:5-2}
\end{figure}
In our project, we planned to compensate this problematic deflection at the junction of the soft finger through the implementation of an appropriate controller. To be more specific, according to Figure \ref{fig:5-2}, the tip of the actuator with length of $L_1$ and under an arbitrary force will be located at point $P$ through the bending curve of $q$ and the deflection of $\beta$. It has been generally accepted that for any given point in the plane, there will be a set of length $L_1'$ and bending curvature $q'$ by which the actuator can reach $P$ without the deflection at the base. Thereby, the objective of controller is to estimate $L_1'$ based on the feedback from the position of $P$. Hence, having the length estimated, the controller generates the necessary control effort to produce corresponding bending curvature of $q'$. \\

Since according to \cite{68}, the aforementioned deflection is considered in kinematics of the robot, we propose a Jacobian-based adaptive-sliding controller \cite{67} that not only compensates the dynamic uncertainties, but also remains robust under the kinematic uncertainties. The control signal, dynamic parameters adaptive law and kinematic parameters adaptive law are designed as in equations \eqref{eq:5-15} to \eqref{eq:5-17}.

\begin{equation} \label{eq:5-15}
u=-{{\hat{K}}^{-1}}{{\hat{J}}^{T}}(q,{{\hat{\theta }}_{k}})({{K}_{v}}\Delta \dot{x}+{{K}_{p}}\Delta x)+{{\hat{K}}^{-1}}{{Y}_{d}}(q,\dot{q},{{\dot{q}}_{r}},{{\ddot{q}}_{r}}){{\hat{\theta }}_{d}}+{{\hat{K}}^{-1}}{{Y}_{a}}({{\tau }_{o}}){{\hat{\theta }}_{a}}
\end{equation}
\begin{equation} \label{eq:5-16}
{{\dot{\hat{\theta }}}_{d}}=-{{L}_{d}}Y_{d}^{T}(q,\dot{q},{{\dot{q}}_{r}},{{\ddot{q}}_{r}})s
\end{equation}
\begin{equation} \label{eq:5-17}
{{\dot{\hat{\theta }}}_{k}}={{L}_{k}}Y_{k}^{T}(q,\dot{q})({{K}_{v}}\Delta \dot{x}+{{K}_{p}}\Delta x)
\end{equation}

In these equations, $Y_d$ is dynamic regressor and $Y_k$ is kinematic regressor which is obtained by equation \eqref{eq:5-2}.
\begin{equation} \label{eq:5-2}
\dot{x}=J(q)\dot{q}={{Y}_{k}}(q,\dot{q}){{\theta }_{k}}
\end{equation}

\section{Simulation Results}
Taking the length of bending actuators as kinematic parameters, we opted the coefficients of the proposed controller in accordance with Table \ref{tab.5.1}. The objective of the controller is to track the trajectory of ($x_d =62+33sin(t) mm$ and $yd=113+26.5sin(t+1.57) mm$ in Cartesian space. The initial estimation of dynamic and kinematic parameters are assigned to be 70 percent of their actual quantities. Moreover, in order to observe the effect of external force, a constant force of 0.2 N is applied to the soft finger. In addition, the evaluation of the proposed controller is carried out by comparing its responses to those of feedback linearization controller.

\begin{table}[ht]
\caption{Adaptive controller gains for Cartesian space simulations} 
\centering 
\begin{tabular}{c c} 

\hline\hline 
			
				 Gains for controller& Magnitude \\[0.5ex] 
				 \hline 
			$K_p$	  & $50$
				 \\
			
				$K_v$  & $10$
				 \\
			
			$L_d$	  & $0.01$ \\
			
			$L_k$	  & $1.6$ \\
				
			$\alpha$	  & 3
				 \\[1ex]
				\hline 
			\end{tabular} 
		\label{tab.5.1}
	
\end{table}

\begin{figure}[H]
	\centering
\captionsetup{justification=centering}		\includegraphics[width=165mm,height=155mm]{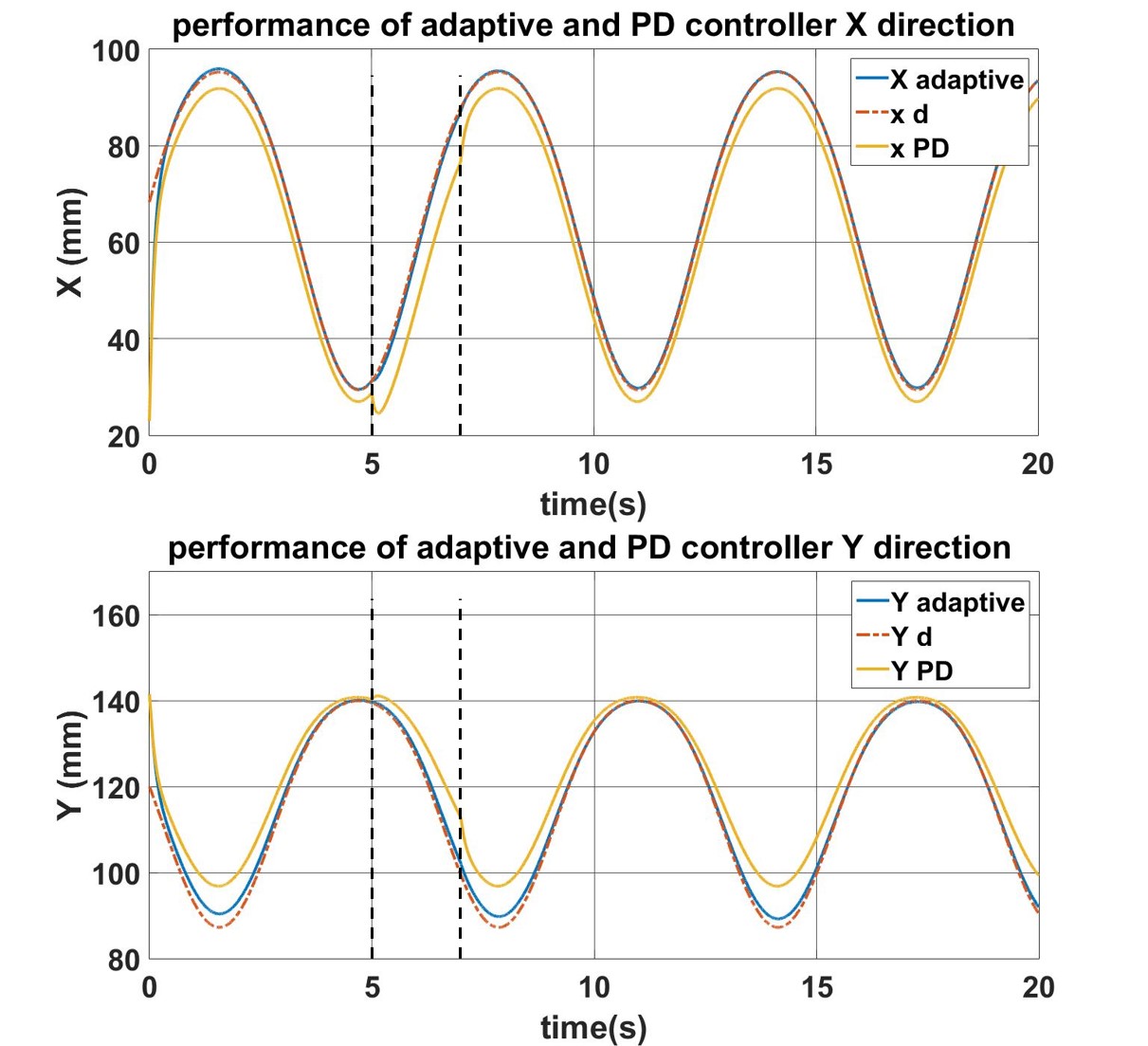}
\caption{Performance of adaptive controller in response to kinematic and dynamic uncertainties in comparison to performance of feed back linearization controller } 

\label{fig:5-14}
\end{figure}

The results of both controllers tracking the reference trajectory are depicted in Figure \ref{fig:5-14} with indication of starting and ending point of force exertion through dash lines. Furthermore, the $RMSE$ of controllers were calculated as in Table \ref{tab.5.2}.


\clearpage
\chapter{In-hand Object Manipulation With Two Soft Fingers} \label{sec:nnmf}
\clearpage
%
\section{ Shaped And Locked system }
\begin{figure}[H]
	\centering
\captionsetup{justification=centering}		\includegraphics[width=97mm,height=53.4mm]{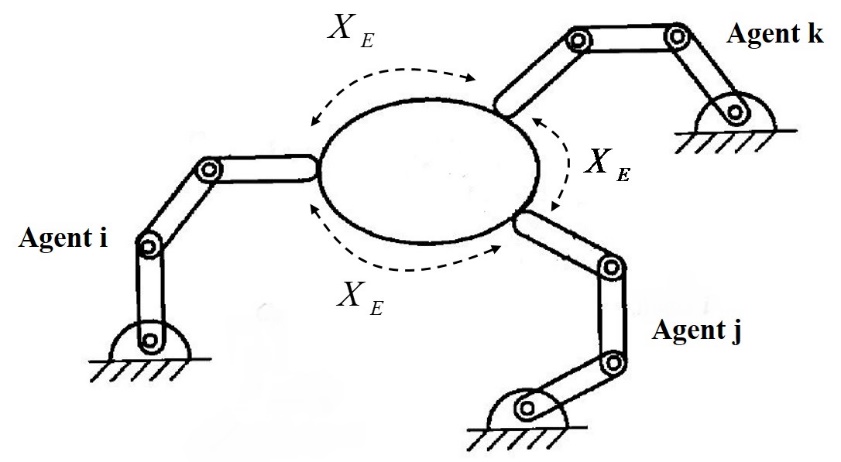}
\caption{Co-operative system composed of several agents \cite{21}}
\label{fig:6-1}
\end{figure}

Having found the dynamic behavior of each soft finger and investigated their performances under the external forces, we could propose a control system that is capable of controlling the states of soft fingers in a way that the overall system ends up doing a co-operative task such as in-hand object manipulation. Thus, we introduce shaped and locked system which decomposes a system composed of several agents (in our case, several soft fingers) into two separately controllable subsystems \cite{69}. These two systems are responsible for defining the formation and maneuver behavior of the overall system. The former is called shaped system which determines the formation of agents relative to their adjacent ones, and the latter is referred to as locked system which states the overall maneuver of the system. For instance, Figure \ref{fig:6-1} shows a system of three agents which are supposed to move the mass center of a grasped object. For this purpose, agents should always maintain specific distance with each other so as to prevent the object form collapsing. Consequently, the control signal of each agent incorporates the information about the states of other agents in the co-operative system..\\

\begin{figure}[H]
	\centering
\captionsetup{justification=centering}		\includegraphics[width=155mm,height=57mm]{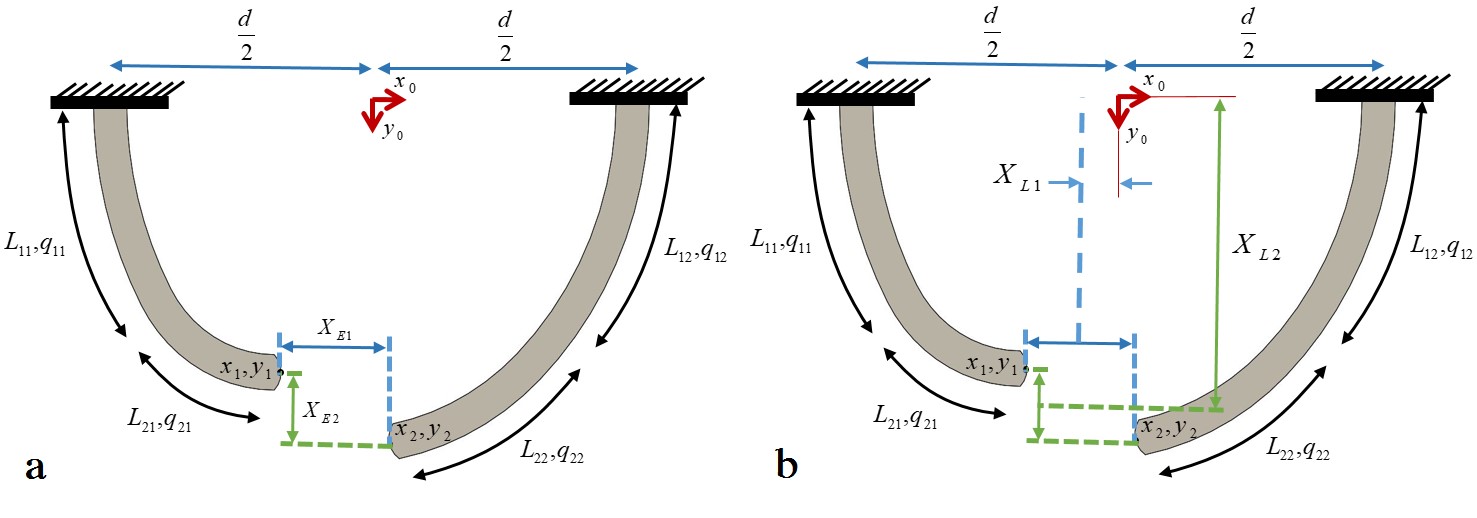}
\caption{Defining shaped and locked variables for two-fingered soft gripper}
\label{fig:6-2}
\end{figure}

It should be stated that the key aspect of developing shaped and locked systems is how their variables are defined based on the states of every individual agent. This issue is addressed by taking into account the objective of the overall system, the number of agents and distinguishing the influential states of the agents on which the control signals are based. For the purpose of the illustration, as in our case, performing the in-hand object manipulation in Cartesian space with two soft fingers requires the positions of the tips of soft fingers in the vertical plane since we do not possess any information about the states of the object itself (blind grasp strategy). Thereby, inspired by human fingers, we try to control the orientation and position of an object merely with the knowledge about the states of our soft fingers. However, it should be noted that the prerequisite for performing such task is to have no slippage at the contact surfaces between the object and fingers, which seems logical since the soft contact boasts a great Coulomb friction coefficient. Hence, having the rolling contact taken for granted, we defined the shaped and locked variables as follows:
\begin{itemize}
\item{Shaped variables ($X_E$): the vertical and horizontal distances between the tip positions of fingers so as to guarantee the stable grasping of the object. Moreover, by controlling the angle of the line connecting the tips we are capable of controlling the orientation of the object (Figure \ref{fig:6-2}.a).}
\item{Locked variables ($X_L$): the mean of vertical and horizontal tip position of soft fingers with the intention of controlling the mass center of the object since we tend to manipulate a rigid square-shaped object (Figure \ref{fig:6-2}.a).}
\end{itemize}

Bringing the shaped and locked variables into matrix form, we can define matrix $S$ as in equation \eqref{eq:6-4}. Furthermore, Jacobian matrix for overall system is calculated based on equation \eqref{eq:6-3}. The aforementioned matrices are critical in obtaining the dynamic matrices of shaped and locked system for equations \eqref{eq:6-2} and \eqref{eq:6-1} respectively. In fact, the inertia and Coriolis matrices could be calculated through the course of operations involving the dynamic matrices of each soft finger and matrices of Jacobian and $S$ as in \cite{70}.

\begin{equation} \label{eq:6-4}
\left[ \begin{matrix}
   {{X}_{L}}  \\
   {{X}_{E}}  
\end{matrix} \right]=\underbrace{\left[ \begin{matrix}
   \frac{1}{2} & 0 & \frac{1}{2} & 0  \\
   0 & \frac{1}{2} & 0 & \frac{1}{2}  \\
   1 & 0 & -1 & 0  \\
   0 & 1 & 0 & -1  
\end{matrix} \right]}_{S}\left[ \begin{matrix}
   {{x}_{1}}  \\
   {{y}_{1}}  \\
   {{x}_{2}}  \\
   {{y}_{2}}  
\end{matrix} \right]=S\left[ \begin{matrix}
   {{x}_{1}}  \\
   {{y}_{1}}  \\
   {{x}_{2}}  \\
   {{y}_{2}}  
\end{matrix} \right]
\end{equation}

\begin{equation} \label{eq:6-3}
J=\left[ \begin{matrix}
   {{J}_{1}} & 0  \\
   0 & {{J}_{2}}  
\end{matrix} \right]
\end{equation}

\begin{equation} \label{eq:6-2}
{{M}_{E}}(q){{\ddot{X}}_{E}}+{{C}_{E}}(q,\dot{q}){{\dot{X}}_{E}}+{{C}_{EL}}(q,\dot{q}){{\dot{X}}_{L}}={{T}_{E}}+{{F}_{EG}}
\end{equation}

\begin{equation} \label{eq:6-1}
{{M}_{L}}(q){{\ddot{X}}_{L}}+{{C}_{L}}(q,\dot{q}){{\dot{X}}_{L}}+{{C}_{LE}}(q,\dot{q}){{\dot{X}}_{E}}={{T}_{L}}+{{F}_{LG}}
\end{equation}

As discussed earlier in preceding chapters, our soft fingers are vulnerable to external forces in a way that their real tip positions mismatch with predicted ones. Since performing in-hand manipulation exposes the fingers to external forces, it is recommended that the same adaptive-sliding controller proposed in previous chapter is used for our goal so as to control the pose of the object as accurately as possible.\\

\section{Controller of Locked System}

Considering the presence of kinematic and dynamic uncertainties, we propose the control signal of equation \eqref{eq:6-24} with adaptation laws of \eqref{eq:6-25}, \eqref{eq:6-30} and \eqref{eq:6-31} \cite{70}.

\begin{equation} \label{eq:6-24}
{{{T}'}_{L}}={{Y}_{Lr}}{{\hat{\theta }}_{Ld}}-k_{v}^{L}{{\hat{\dot{e}}}_{L}}-k_{p}^{L}{{\hat{e}}_{L}}-{{\hat{F}}_{L}}
\end{equation}

\begin{equation} \label{eq:6-25}
{{\hat{\dot{\theta }}}_{Lk}}={{\Gamma }_{Lk}}Y_{Lk}^{T}\left\{ k_{L}^{v}{{{\hat{\dot{e}}}}_{L}}+k_{L}^{p}{{{\hat{e}}}_{L}} \right\}
\end{equation}
\begin{equation} \label{eq:6-30}
{{\hat{\dot{\theta }}}_{Ld}}=-\Gamma _{Ld}^{T}{{({{Y}_{Lr}})}^{T}}{{\hat{s}}_{L}}
\end{equation}
\begin{equation} \label{eq:6-31}
{{\hat{\dot{F}}}_{L}}=-{{P}_{L}}^{-T}{{\hat{s}}_{L}}
\end{equation}

It should be mentioned that $Y_Lr$ and $Y_Lk$ are dynamic and kinematic regressors of locked system and $s_L$ is sliding vector for locked system that are determined in \cite{70}. Moreover, the position error of locked system is defined as $e_L=X_L-X_{ld}$, and the stability analysis of proposed control signal of \eqref{eq:6-24} is available in \cite{70}. 

\section{Controller of Shaped System}
For the sake of controlling the orientation of the grasped object, we propose, based on \cite{70}, the control signal of \eqref{eq:6-35} for shaped system without considering kinematic uncertainties since they are compensated through the implementation of locked control system. The adaptation laws are calculated as in \eqref{eq:6-36} and \eqref{eq:6-37}.
\begin{equation} \label{eq:6-35}
{{{T}'}_{E}}={{Y}_{Er}}{{\hat{\theta }}_{Ed}}-{{k}_{d}}{{\hat{s}}_{E}}-k{{\hat{e}}_{E}}-{{\hat{F}}_{E}}
\end{equation}
\begin{equation} \label{eq:6-36}
{{\hat{\dot{\theta }}}_{Ed}}=-\Gamma _{Ed}^{T}{{({{Y}_{Er}})}^{T}}{{\hat{s}}_{E}}
\end{equation}
\begin{equation} \label{eq:6-37}
{{\hat{\dot{F}}}_{E}}=-{{P}_{E}}^{-T}{{\hat{s}}_{E}}
\end{equation}

In these equations $Y_{Er}$ is dynamic regressor of locked system and $s_{E}$ is sliding vector for shaped system that are determined in [70]. In addition, the position error of shaped system is defined as $e_E=X_E-X_{Ed}$, and the stability analysis of proposed control signal was carried out in \cite{70} as well.

\section{Gripping Force}
\begin{figure}[H]
	\centering
\captionsetup{justification=centering}		\includegraphics[width=72mm,height=48mm]{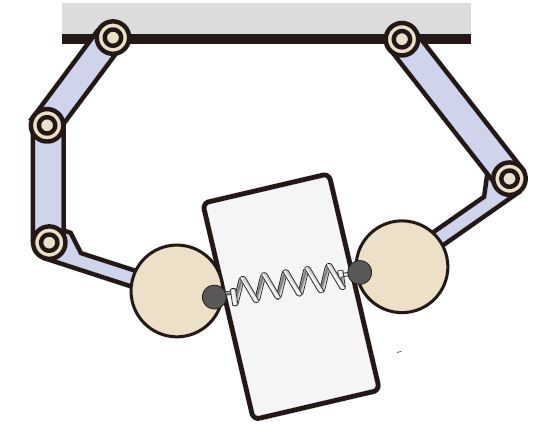}
\caption{ Assuming an imaginary spring for generating gripping force \cite{72}}
\label{fig:6-4}
\end{figure}

Since the soft fingers are supposed to perform blind grasp, it is likely that they are obliged to grasp a slightly heavy object whose weight might excess the friction forces at the contact surfaces. Hence, the control system has to yield adequate normal force at the contact surfaces to guarantee the grasp of the object. Influenced by \cite{72}, we considered an imaginary spring between the soft fingers' tips as illustrated in figure \ref{fig:6-4}. Accordingly, there would be always a pair of normal forces, as in equation \eqref{eq:6-63}, that hold the object at contact with tips of the fingers. Finally, the stiffness of the imaginary spring should be determined with respect to the maximum weight of the object which might be involved in manipulation. Subsequently, the gripping force will be incorporated in control signal of the system.
\begin{equation} \label{eq:6-63}
{{f}_{i}}={{(-1)}^{i+1}}\,{{K}_{s}}\,\,\left[ \begin{matrix}
   {{x}_{2}}-{{x}_{1}}  \\
   {{y}_{2}}-{{y}_{1}}  \\
\end{matrix} \right]\,\,,\,\,i=1,2
\end{equation}

\section{Simulations}
In order to evaluate the performance of the proposed shaped and locked system along with adaptive-sliding controller, we conducted a Simulink simulation in which the system of two soft fingers were supposed to grasp a rigid square-shaped object with 20 mm in length and 20 g in mass. The mass distribution of the object was assumed to be uniform. It should be noted that the distance between the bases of the fingers is 20 cm.\\
Therefore, the objective of the controller is to track the reference inputs of \eqref{eq:6-64} and \eqref{eq:6-65} while a constant force of 0.3 N was being exerted on both fingers after the first second of simulation.

\begin{equation} \label{eq:6-64}
{{X}_{Ed}}=\left[ \begin{matrix}
   20  \\
   0  \\
\end{matrix} \right]
\end{equation}
\begin{equation} \label{eq:6-65}
{{X}_{Ld}}=\left[ \begin{matrix}
   10+10\sin (3t)  \\
   81-17\sin (3t)  \\
\end{matrix} \right]
\end{equation}

\begin{figure}[H]
	\centering
\captionsetup{justification=centering}		\includegraphics[width=163mm,height=80mm]{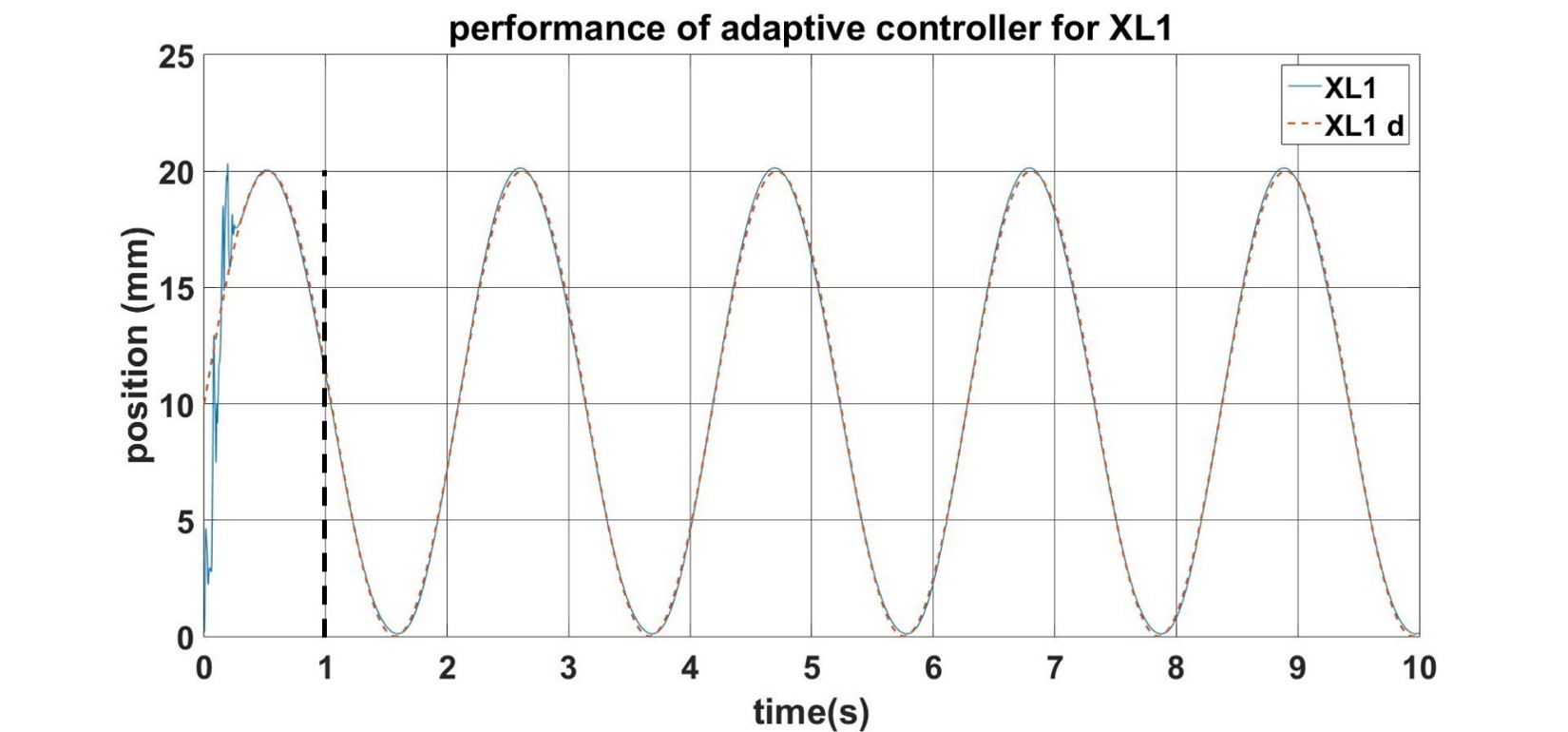}
\caption{Performance of locked control system in horizontal direction}
\label{fig:6-6}
\end{figure}

\begin{figure}[H]
	\centering
\captionsetup{justification=centering}		\includegraphics[width=163mm,height=80mm]{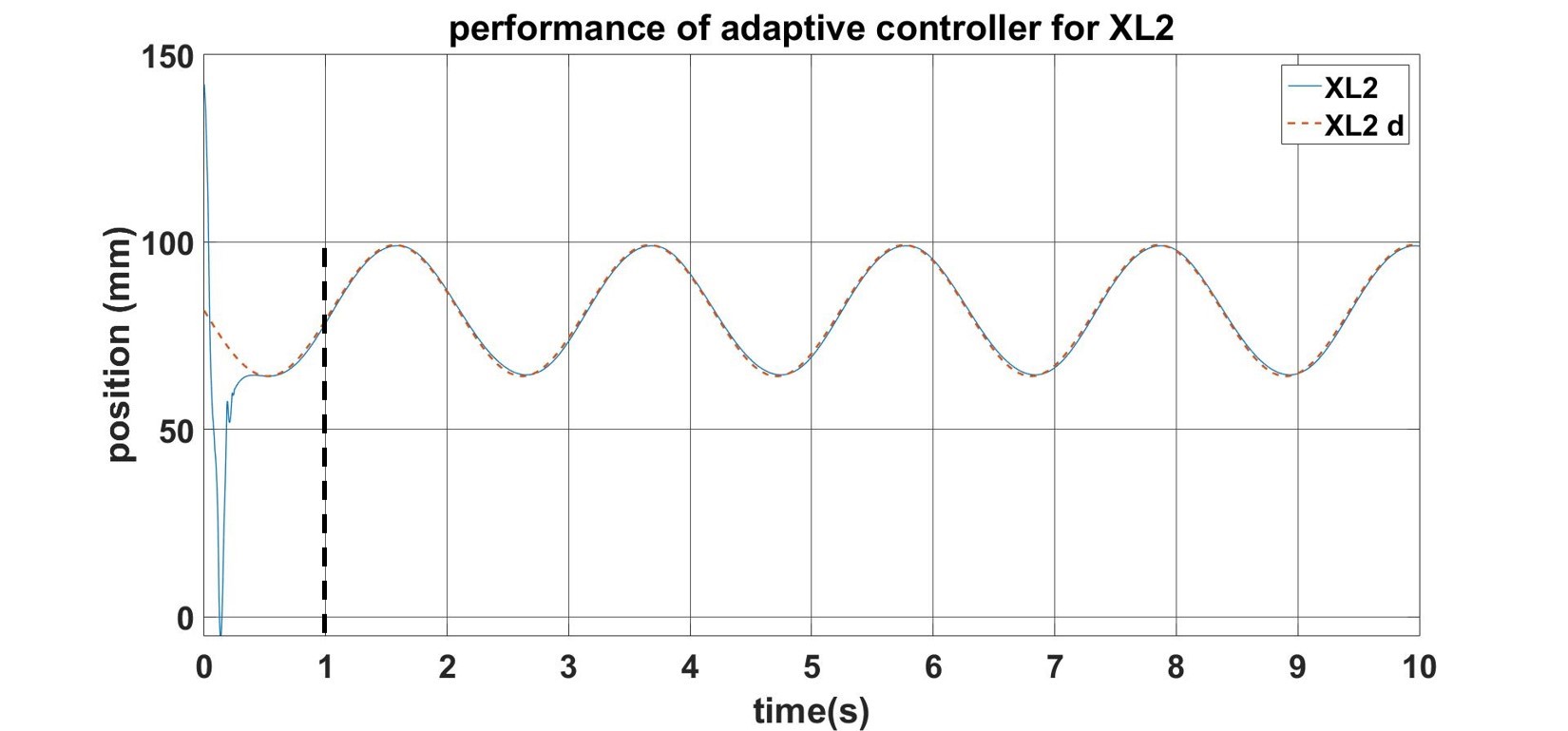}
\caption{Performance of locked control system in vertical direction}
\label{fig:6-7}
\end{figure}

\begin{figure}[H]
	\centering
\captionsetup{justification=centering}		\includegraphics[width=163mm,height=80mm]{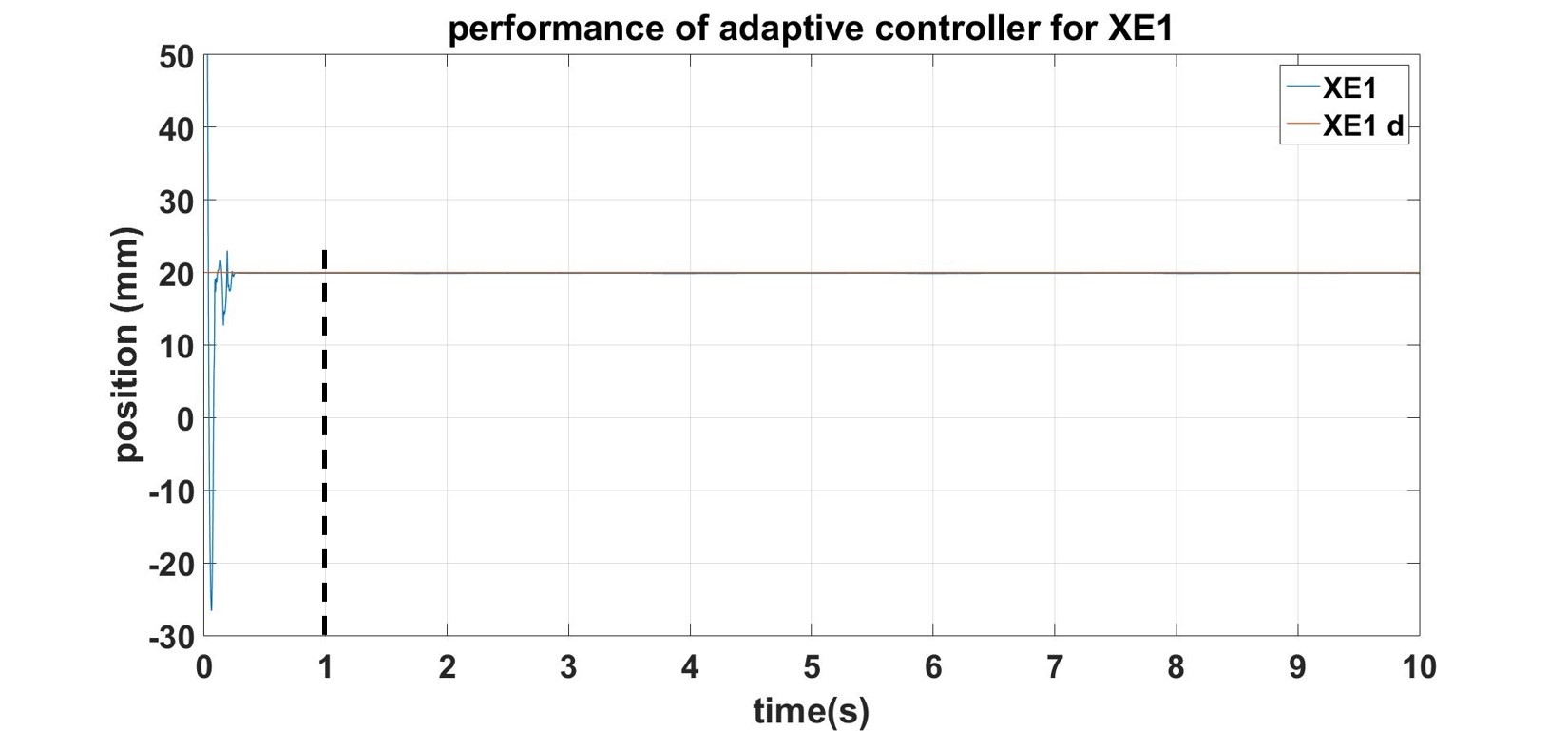}
\caption{Performance of shaped control system in horizontal direction}
\label{fig:6-9}
\end{figure}

\begin{figure}[H]
	\centering
\captionsetup{justification=centering}		\includegraphics[width=163mm,height=80mm]{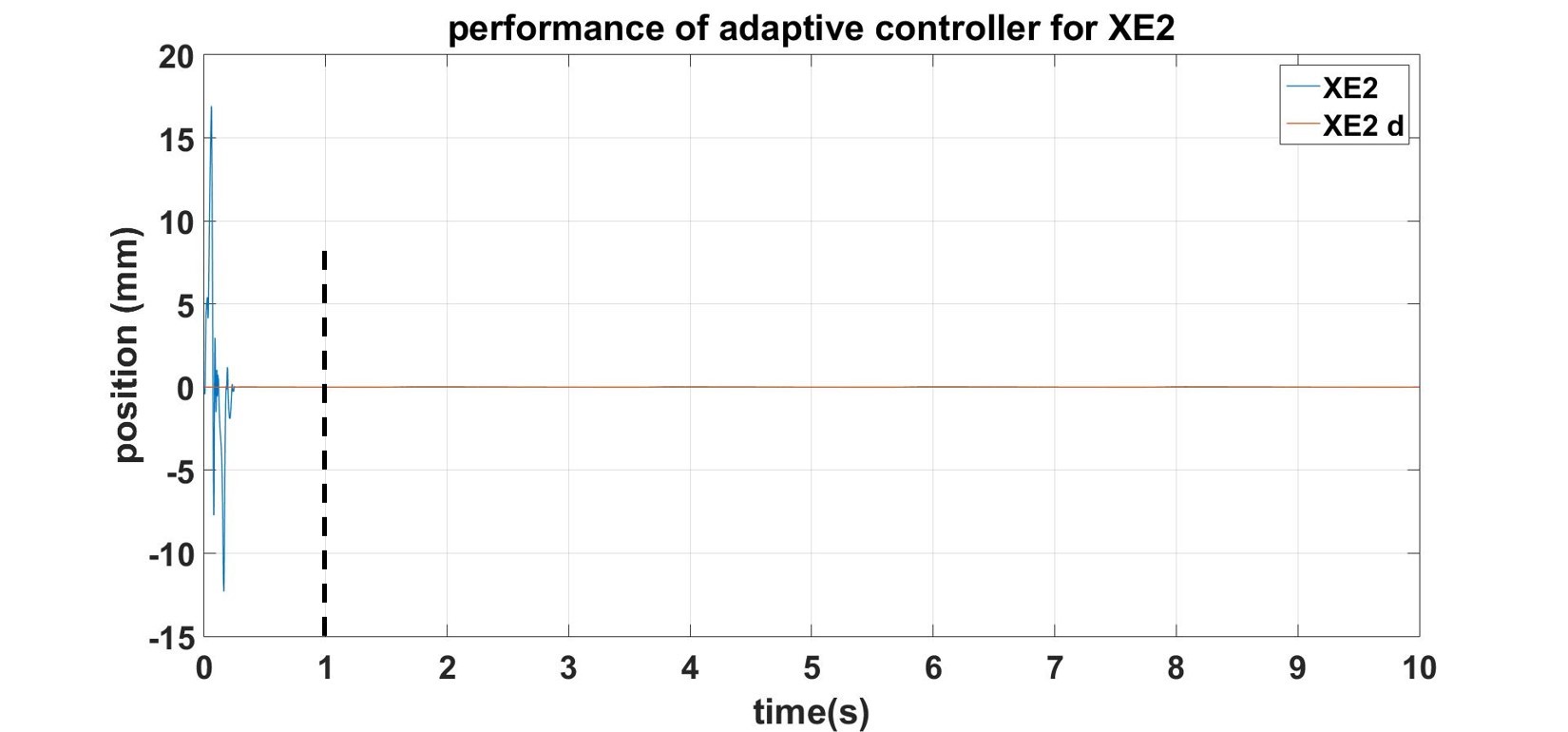}
\caption{Performance of shaped control system in vertical direction}
\label{fig:6-10}
\end{figure}

The results of the simulation, based on the selected control parameters of Tables \ref{tab.6.1} and \ref{tab.6.2}, are shown in Figures \ref{fig:6-6} to \ref{fig:6-10}.

\begin{table}[ht]
\caption{Adaptive controller gains for locked system} 
\centering 
\begin{tabular}{c c} 

\hline\hline 
			
				 Gains for controller& Magnitude \\[0.5ex] 
				 \hline 
			$k_P^L$	  & $200$
				 \\
			
				$k_v^L$  & $3$
				 \\
			
			$\Gamma_{Lk}$	  & $0.0325$ \\
			
			$\Gamma_{Ld}$	  & $0.4$ \\
			
			$P_L$	  & $1$ \\
				
			$\lambda_1$	  & 20
				 \\[1ex]
				\hline 
			\end{tabular} 
		\label{tab.6.1}
	
\end{table}

\begin{table}[ht]
\caption{Adaptive controller gains for shaped system} 
\centering 
\begin{tabular}{c c} 

\hline\hline 
			
				 Gains for controller& Magnitude \\[0.5ex] 
				 \hline 
			$k$	  & $100$
				 \\
			
				$k_d$  & $5$
				 \\
			
			$\Gamma_{Lk}^T$	  & $0.4$ \\
			
			$P_E$	  & $1$ \\
				
			$\lambda_2$	  & 20
				 \\[1ex]
				\hline 
			\end{tabular} 
		\label{tab.6.2}
	
\end{table}

\begin{figure}[H]
	\centering
\captionsetup{justification=centering}		\includegraphics[width=163mm,height=80mm]{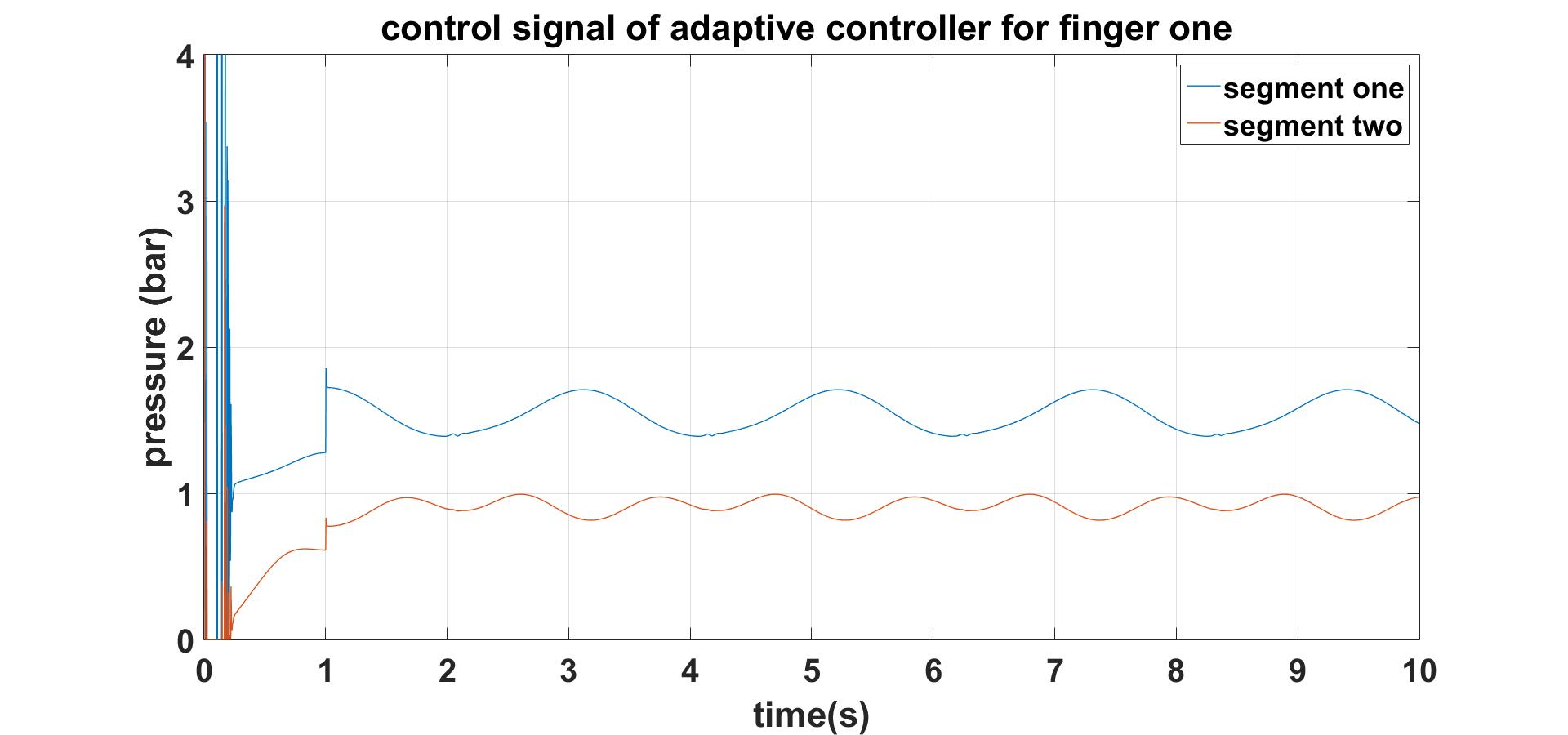}
\caption{Control pressure for soft finger one in shaped and locked system}
\label{fig:6-12}
\end{figure}

\begin{figure}[H]
	\centering
\captionsetup{justification=centering}		\includegraphics[width=163mm,height=80mm]{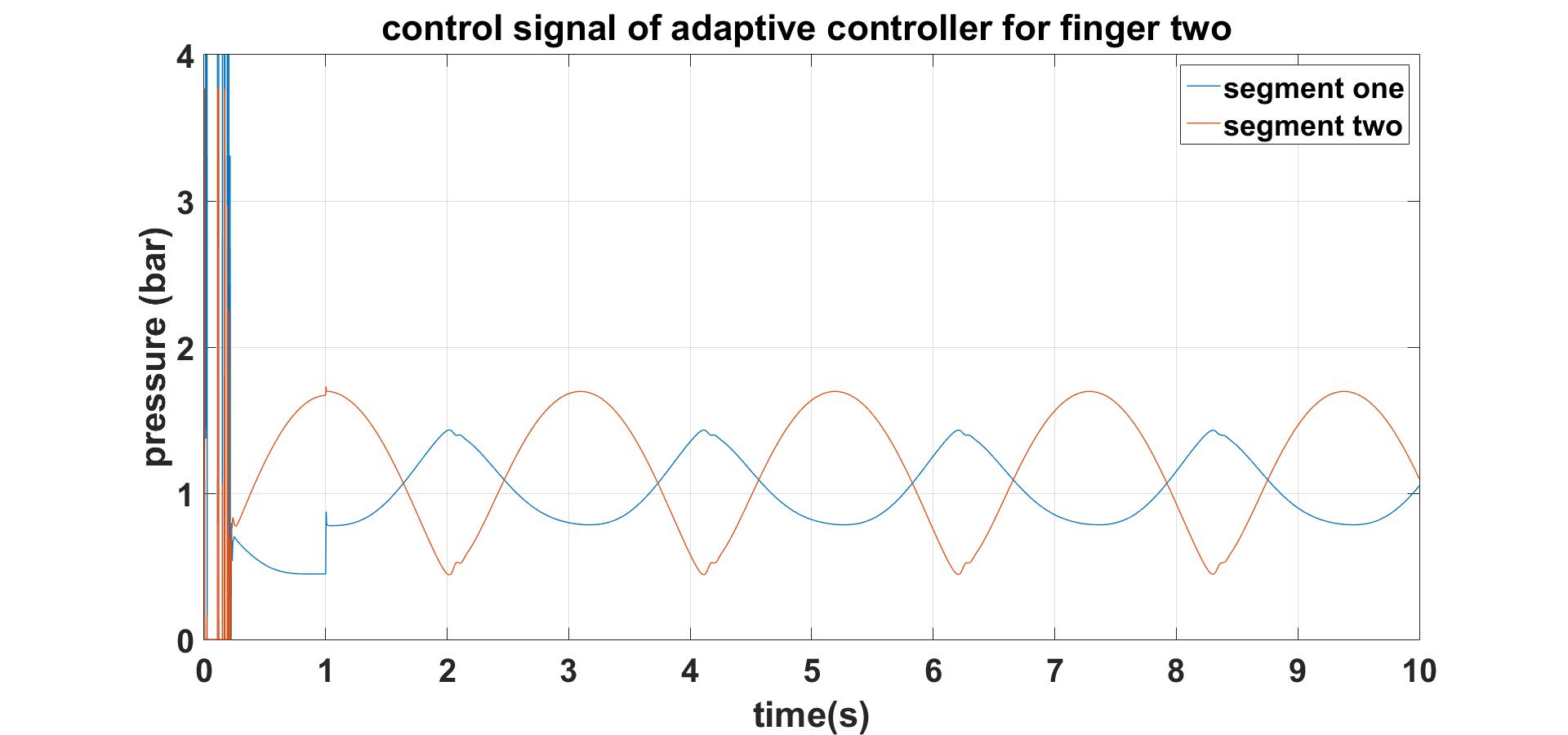}
\caption{Control pressure for soft finger two in shaped and locked system}
\label{fig:6-13}
\end{figure}

Moreover, the pressure input for each finger is depicted in \ref{fig:6-12} and \ref{fig:6-13}.

Finally, $RMSE$ calculated for the proposed controller in vertical and horizontal directions are listed in Table \ref{tab.6.4}.

\begin{table}[h]
\caption{ $RMSE$ for locked and shaped controllers in both directions} 
\centering 
\begin{tabular}{c c c c} 
\hline\hline 
Controller & System & $RMSE$ X $mm$ & $RMSE$ Y $mm$
\\ [0.5ex]
\hline 
& Locked & 0.56 & 4.7  \\[-1ex]
\raisebox{1.5ex}{Adaptive-sliding} & Shaped& {$3.9$}
& 0.66  \\[1ex]

\hline 
\end{tabular}
\label{tab.6.4}
\end{table}



\appendix
\renewcommand*{\thesection}{\Alph{section}}\textbf{}




\bibliographystyle{ieeetr}
\bibliography{references}

\begin{thebibliography}{10}

\bibitem{13}
D.~Trivedi, C.~D. Rahn, W.~M. Kier, and I.~D. Walker, ``{Soft robotics:
  Biological inspiration, state of the art, and future research},'' {\em
  Applied Bionics and Biomechanics}, vol.~5, no.~3, pp.~99--117, 2008.

\bibitem{59}
R.~Ozawa and K.~Tahara, ``{Grasp and dexterous manipulation of multi-fingered
  robotic hands: a review from a control view point},'' {\em Advanced
  Robotics}, vol.~31, no.~19-20, pp.~1030--1050, 2017.

\bibitem{20}
R.~K. Katzschmann, ``{Building and controlling fluidically actuated soft robots
  : from open loop to model-based control},'' no.~2013, 2018.

\bibitem{40}
C.~{Della Santina}, R.~K. Katzschmann, A.~Bicchi, and D.~Rus, ``{Dynamic
  control of soft robots interacting with the environment},'' {\em 2018 IEEE
  International Conference on Soft Robotics, RoboSoft 2018}, pp.~46--53, 2018.

\bibitem{68}
Z.~Gong, J.~Cheng, X.~Chen, W.~Sun, X.~Fang, K.~Hu, Z.~Xie, T.~Wang, and
  L.~Wen, ``{A Bio-inspired Soft Robotic Arm: Kinematic Modeling and
  Hydrodynamic Experiments},'' {\em Journal of Bionic Engineering}, vol.~15,
  no.~2, pp.~204--219, 2018.

\bibitem{21}
Z.~Li, P.~Hsu, and S.~Sastry, ``{Grasping and Coordinated Manipulation by a
  Multifingered Robot Hand},'' {\em The International Journal of Robotics
  Research}, vol.~8, no.~4, pp.~33--50, 1989.

\bibitem{72}
R.~Ozawa and J.~H. Bae, {\em {Dynamic manipulation based on thumb opposability:
  Passivity-based blind grasping and manipulation}}.
\newblock Elsevier Inc., 2018.

\bibitem{23}
D.~Rus and M.~T. Tolley, ``{Design, fabrication and control of soft robots},''
  {\em Nature}, vol.~521, no.~7553, pp.~467--475, 2015.

\bibitem{24}
C.~Majidi, ``{Soft Robotics: A Perspective - Current Trends and Prospects for
  the Future},'' {\em Soft Robotics}, vol.~1, no.~1, pp.~5--11, 2014.

\bibitem{27}
J.~Shintake, V.~Cacucciolo, D.~Floreano, and H.~Shea, ``{Soft Robotic
  Grippers},'' {\em Advanced Materials}, vol.~30, no.~29, 2018.

\bibitem{19}
R.~R. Ma and A.~M. Dollar, ``{On dexterity and dexterous manipulation},'' {\em
  IEEE 15th International Conference on Advanced Robotics: New Boundaries for
  Robotics, ICAR 2011}, pp.~1--7, 2011.

\bibitem{60}
R.~M. Murray and S.~S. Sastry, ``{Control experiments in planar manipulation
  and grasping},'' no.~6, pp.~624--629, 1989.

\bibitem{1}
R.~Ozawa, S.~Arimoto, S.~Nakamura, and J.~H. Bae, ``{Control of an object with
  parallel surfaces by a pair of finger robots without object sensing},'' {\em
  IEEE Transactions on Robotics}, vol.~21, no.~5, pp.~965--976, 2005.

\bibitem{61}
S.~Arimoto, M.~Yoshida, and J.~H. Bae, ``{Stable "blind grasping" of a 3-D
  object under non-holonomic constraints},'' {\em Proceedings - IEEE
  International Conference on Robotics and Automation}, vol.~2006, no.~May,
  pp.~2124--2130, 2006.

\bibitem{30}
H.~Al-Fahaam, S.~Davis, and S.~Nefti-Meziani, ``{The design and mathematical
  modelling of novel extensor bending pneumatic artificial muscles (EBPAMs) for
  soft exoskeletons},'' {\em Robotics and Autonomous Systems}, vol.~99,
  pp.~63--74, 2018.

\bibitem{34}
Z.~Wang, P.~Polygerinos, J.~T. Overvelde, K.~C. Galloway, K.~Bertoldi, and
  C.~J. Walsh, ``{Interaction Forces of Soft Fiber Reinforced Bending
  Actuators},'' {\em IEEE/ASME Transactions on Mechatronics}, vol.~22, no.~2,
  pp.~717--727, 2017.

\bibitem{39}
X.~Zhou, C.~Majidi, and O.~M. O'Reilly, ``{Soft hands: An analysis of some
  gripping mechanisms in soft robot design},'' {\em International Journal of
  Solids and Structures}, vol.~64, pp.~155--165, 2015.

\bibitem{32}
K.~Elgeneidy, N.~Lohse, and M.~Jackson, ``{Bending angle prediction and control
  of soft pneumatic actuators with embedded flex sensors – A data-driven
  approach},'' {\em Mechatronics}, vol.~50, no.~October, pp.~234--247, 2018.

\bibitem{53}
G.~Zheng, Y.~Zhou, and M.~Ju, ``{Robust control of a silicone soft robot using
  neural networks},'' {\em ISA Transactions}, vol.~100, no.~December,
  pp.~38--45, 2020.

\bibitem{58}
N.~Tan, P.~Yu, X.~Zhang, and T.~Wang, ``{Model-free motion control of continuum
  robots based on a zeroing neurodynamic approach},'' {\em Neural Networks},
  vol.~133, pp.~21--31, 2021.

\bibitem{36}
Z.~Wang and S.~Hirai, ``{Soft Gripper Dynamics Using a Line-Segment Model With
  an Optimization-Based Parameter Identification Method},'' {\em IEEE Robotics
  and Automation Letters}, vol.~2, no.~2, pp.~624--631, 2017.

\bibitem{42}
V.~Falkenhahn, A.~Hildebrandt, R.~Neumann, and O.~Sawodny, ``{Dynamic Control
  of the Bionic Handling Assistant},'' {\em IEEE/ASME Transactions on
  Mechatronics}, vol.~22, no.~1, pp.~6--17, 2017.

\bibitem{37}
T.~Wang, Y.~Zhang, Y.~Zhu, and S.~Zhu, ``{A computationally efficient dynamical
  model of fluidic soft actuators and its experimental verification},'' {\em
  Mechatronics}, vol.~58, no.~January, pp.~1--8, 2019.

\bibitem{38}
T.~Wang, Y.~Zhang, Z.~Chen, and S.~Zhu, ``{Parameter Identification and
  Model-Based Nonlinear Robust Control of Fluidic Soft Bending Actuators},''
  {\em IEEE/ASME Transactions on Mechatronics}, vol.~24, no.~3, pp.~1346--1355,
  2019.

\bibitem{49}
F.~Renda, F.~Boyer, J.~Dias, and L.~Seneviratne, ``{Discrete Cosserat Approach
  for Multisection Soft Manipulator Dynamics},'' {\em IEEE Transactions on
  Robotics}, vol.~34, no.~6, pp.~1518--1533, 2018.

\bibitem{45}
H.~Wang, W.~Chen, X.~Yu, T.~Deng, X.~Wang, and R.~Pfeifer, ``{Visual servo
  control of cable-driven soft robotic manipulator},'' {\em IEEE International
  Conference on Intelligent Robots and Systems}, pp.~57--62, 2013.

\bibitem{50}
Y.~Shapiro, K.~Gabor, and A.~Wolf, ``{Modeling a hyperflexible planar bending
  actuator as an inextensible euler-bernoulli beam for use in flexible
  robots},'' {\em Soft Robotics}, vol.~2, no.~2, pp.~71--79, 2015.

\bibitem{52}
L.~Lindenroth, J.~Back, A.~Schoisengeier, Y.~Noh, H.~W{\"{u}}rdemann,
  K.~Althoefer, and H.~Liu, ``{Stiffness-based modelling of a
  hydraulically-actuated soft robotics manipulator},'' {\em IEEE International
  Conference on Intelligent Robots and Systems}, vol.~2016-Novem,
  pp.~2458--2463, 2016.

\bibitem{57}
Y.~Chen, W.~Li, and Y.~Gong, ``{Static modeling and analysis of soft
  manipulator considering environment contact based on segmented constant
  curvature method},'' {\em Industrial Robot}, no.~September, 2020.

\bibitem{galloway2013}
K.~C. Galloway, P.~Polygerinos, C.~J. Walsh, and R.~J. Wood, ``{Galloway et al.
  - 2013 - Mechanically programmable bend radius for fiber-rein.pdf},'' 2013.

\bibitem{santina2020}
C.~{Della Santina}, R.~K. Katzschmann, A.~Bicchi, and D.~Rus, ``{Model-based
  dynamic feedback control of a planar soft robot: trajectory tracking and
  interaction with the environment},'' {\em International Journal of Robotics
  Research}, vol.~39, no.~4, pp.~490--513, 2020.

\bibitem{66}
J.~J.~E. Slotine and W.~Li, ``{on the Adaptive Control of Robot
  Manipulators.},'' {\em International Journal of Robotics Research}, vol.~6,
  no.~3, pp.~49--59, 1987.

\bibitem{67}
C.~C. Cheah, C.~Liu, and J.~J. Slotine, ``{Adaptive Jacobian tracking control
  of robots with uncertainties in kinematic, dynamic and actuator models},''
  {\em IEEE Transactions on Automatic Control}, vol.~51, no.~6, pp.~1024--1029,
  2006.

\bibitem{69}
D.~Lee and P.~Y. Li, ``{Passive decomposition approach to formation and
  maneuver control of multiple rigid bodies},'' {\em Journal of Dynamic
  Systems, Measurement and Control, Transactions of the ASME}, vol.~129, no.~5,
  pp.~662--677, 2007.

\bibitem{70}
R.~Monfaredi, S.~M. Rezaei, and A.~Talebi, ``{A new observer-based adaptive
  controller for cooperative handling of an unknown object},'' {\em Robotica},
  vol.~34, no.~7, pp.~1437--1463, 2016.

\end{thebibliography}
\addcontentsline{toc}{chapter}{Bibliography}

\end{document}